\documentclass{ieeeaccess}
\usepackage{cite}
\usepackage{amsmath,amssymb,amsfonts}
\usepackage{algorithmic}
\usepackage{graphicx}
\usepackage{textcomp}
\usepackage{subcaption}
\usepackage{soul}
\usepackage{booktabs}
\usepackage{cuted}

\usepackage{soul}

\usepackage{multirow}
\newcommand{\NA}{---}

\def\BibTeX{{\rm B\kern-.05em{\sc i\kern-.025em b}\kern-.08em
    T\kern-.1667em\lower.7ex\hbox{E}\kern-.125emX}}

\begin{document}

\history{Date of publication xxxx 00, 0000, date of current version xxxx 00, 0000.}
\doi{10.1109/ACCESS.2024.3349970}

\title{Sentiment Analysis in Finance: From Transformers Back to eXplainable Lexicons (XLex)}

\author{
    \uppercase{Maryan Rizinski}\authorrefmark{1,2},
    \uppercase{Hristijan Peshov}\authorrefmark{2},
    \uppercase{Kostadin Mishev}\authorrefmark{2},
    \uppercase{Milos Jovanovik}\authorrefmark{2},
    \uppercase{and Dimitar Trajanov}\authorrefmark{2,1}\IEEEmembership{Member, IEEE}
}

\address[1]{Department of Computer Science, Metropolitan College, Boston University, Boston, MA 02215, USA}
\address[2]{Faculty of Computer Science and Engineering, Ss. Cyril and Methodius University, 1000 Skopje, North Macedonia}


\markboth
{Author \headeretal: Preparation of Papers for IEEE TRANSACTIONS and JOURNALS}
{Author \headeretal: Preparation of Papers for IEEE TRANSACTIONS and JOURNALS}

\corresp{Corresponding author: Maryan Rizinski (e-mail: rizinski@bu.edu).}

\begin{abstract}
Lexicon-based sentiment analysis in finance leverages specialized, manually annotated lexicons created by human experts to extract sentiment from financial texts effectively. Although lexicon-based methods are simple to implement and fast to operate on textual data, they require considerable manual annotation efforts to create, maintain, and update the lexicons. These methods are also considered inferior to the deep learning-based approaches, such as transformer models, which have become dominant in various natural language processing (NLP) tasks due to their remarkable performance. However, their efficacy comes at a cost: these models require extensive data and computational resources for both training and testing. Additionally, they involve significant prediction times, making them unsuitable for real-time production environments or systems with limited processing capabilities. In this paper, we introduce a novel methodology named eXplainable Lexicons (XLex) that combines the advantages of both lexicon-based methods and transformer models. We propose an approach that utilizes transformers and SHapley Additive exPlanations (SHAP) for explainability to automatically learn financial lexicons. Our study presents four main contributions. 
Firstly, we demonstrate that transformer-aided explainable lexicons can enhance the vocabulary coverage of the benchmark Loughran-McDonald (LM) lexicon. This enhancement leads to a significant reduction in the need for human involvement in the process of annotating, maintaining, and updating the lexicons. Secondly, we show that the resulting lexicon outperforms the standard LM lexicon in sentiment analysis of financial datasets.
Our experiments show that XLex outperforms LM when applied to general financial texts, resulting in enhanced word coverage and an overall increase in classification accuracy by 0.431. Furthermore, by employing XLex to extend LM, we create a combined dictionary, XLex+LM, which achieves an even higher accuracy improvement of 0.450. Thirdly, we illustrate that the lexicon-based approach is significantly more efficient in terms of model speed and size compared to transformers.
Lastly, the proposed XLex approach is inherently more interpretable than transformer models. This interpretability is advantageous as lexicon models rely on predefined rules, unlike transformers, which have complex inner workings. The interpretability of the models allows for better understanding and insights into the results of sentiment analysis, making the XLex approach a valuable tool for financial decision-making.
\end{abstract}


\titlepgskip=-15pt

\maketitle

\section{Introduction}
\label{sec:1}

The financial industry generates massive amounts of data, from transactional data to news articles and social media posts \cite{hasan2020current, goldstein2021big}. This big data poses significant challenges and opportunities for financial institutions as they struggle to extract insights and make sense of the vast amounts of information generated every day. Extracting meaningful trends and actionable knowledge from such an immense quantity of data is so complex and time-consuming that it makes it impossible to perform by any individual actor or stakeholder in the financial market. Thus, automatic approaches for big data analytics are becoming essential in addressing the underlying challenges in finance \cite{mohamed2014real, ravi2017big, cao2015big}.

Sentiment analysis can play a crucial role in analyzing, interpreting, and extracting insights from big financial data. Sentiment analysis has become increasingly important in the field of finance and fintech, where it has gained popularity in a wide range of applications. One of the main use cases of sentiment analysis in finance is to predict stock market trends \cite{smailovic2013predictive, derakhshan2019sentiment, ren2018forecasting, yang2020big}. By analyzing news articles, social media posts, balance sheets, cash flow statements, and other sources of financial information, sentiment analysis can be used to capture market sentiment, which can help investors in making more informed decisions. For example, if sentiment analysis indicates that the overall market sentiment is negative, investors may choose to sell their stocks to avoid potential losses. Additionally, sentiment analysis can help financial institutions and regulators monitor financial markets and investors' behavior to detect potential manipulations, speculations, or fraudulent activities.

Another application of sentiment analysis in finance is to assess the creditworthiness of individuals and companies \cite{tsai2010effects, gul2018multiple, lu2012credit, zhang2015can, yoon2020detecting}. By analyzing social media activity, customer reviews, and other sources of data, sentiment analysis can provide insights into the financial behavior and reputation of borrowers. This can help lenders make more informed decisions about lending and pricing, ultimately reducing the risk of default and improving profitability.

In fintech, sentiment analysis can be used to improve customer experience and engagement \cite{mccoll2019gaining, ziora2016sentiment, mili2016context, tian2021data, chen2021fintech}. By analyzing customer feedback, fintech companies can better identify and address customer needs, preferences, and problems. This information can be used to develop personalized products and services that are tailored to customer expectations, thereby resulting in increased customer satisfaction and loyalty. Additionally, sentiment analysis can help fintech companies monitor their brand reputation and detect potential issues before they become widespread, improving overall brand image and customer trust \cite{liu2020aspect, benedetto2016big, alessia2015approaches}.

Lexicon-based sentiment analysis is a commonly used approach that relies on pre-defined sets of words known as lexicons \cite{taboada2011lexicon, taj2019sentiment, wankhade2022survey}. Lexicons are manually annotated by experts in the field, and sentiment scores are assigned to individual words (positive, negative, or neutral). While knowledge extraction using lexicons exhibits a simplistic implementation and fast operation on textual data, considerable manual annotation efforts are required to create, maintain, and update such lexicons. However, even after such laborious annotation, some relevant words may still not be included in the lexicon, potentially leading to reduced sentiment classification accuracy. Furthermore, lexicons tailored for one domain, such as finance, cannot be easily reused in other domains. As indicated in the seminal study by Loughran and McDonald \cite{loughran2011liability}, dictionaries developed for other disciplines may misclassify common words in financial texts, highlighting the importance of domain-specific lexicons. Generic lexicons are also used for general-purpose sentiment analysis. However, they are known to be imprecise in various domains, introducing inaccuracies and biases \cite{loughran2011liability}.

Another approach to sentiment analysis is by using machine learning (ML) \cite{boiy2009machine, sharma2012comparative, malviya2020machine, neethu2013sentiment} and deep learning (DL) techniques \cite{zhang2018deep, yadav2020sentiment, dang2020sentiment, tang2015deep, wankhade2022survey}. ML/DL techniques are based on sophisticated algorithms that can capture complex linguistic patterns. For example, DL approaches, such as the state-of-the-art (SOTA) transformer models \cite{vaswani2017attention, devlin2018bert}, can learn contextual and semantic information as well as capture long-term dependencies in text, making them effective in capturing the nuances of sentiment in text \cite{mishev2020evaluation}. However, transformer models typically require massive amounts of text data, which can be computationally expensive to train and implement \cite{huang2020improving}.

Sentiment extraction from financial texts requires the use of domain-specific language. The traditional approach for sentiment analysis in finance is to use manually annotated lexicons, such as the Loughran-McDonald (LM) lexicon. To create the LM lexicon, its authors employed Release 4.0 of the \textit{2of12inf} dictionary as a basis and extended it using 10-X fillings\footnote{Detailed documentation on the LM development can be found in https://sraf.nd.edu/loughranmcdonald-master-dictionary/}. The LM authors do not extract all words from the fillings; they rather use only those words that appear frequently (with a frequency count of 50 or more)\footnote{It is worth mentioning that the XLex methodology that we develop in this paper is not bound to the frequency count.}. This means that LM will not have a recall even for a large number of the sentences present in the 10-X fillings. The approach of using lexicons annotated by experts has its limitations as manual editing efforts are required to maintain and update such lexicons. While transformers have shown superior performance in sentiment classification tasks, little work has been done to investigate how these approaches can be combined to create improved lexicons automatically.

In this paper, we explore the potential of transformers and ML explainability tools such as SHapley Additive exPlanations (SHAP) \cite{lundberg2017unified} for automating the creation of lexicons, reducing their maintenance efforts, and expanding their vocabulary coverage. We propose a new methodology for building eXplainable Lexicons (XLex) using pre-trained transformer models and explainable ML tools. The results demonstrate that the proposed methodology leads to the creation of new lexicons that outperform the current state-of-the-art sentiment lexicons in finance. 

Our research focuses on sentiment analysis in the field of finance, driven by the recognition of the Loughran-McDonald lexicon as a standard baseline for sentiment analysis in this domain. As indicated in a 2016 paper by the LM authors \cite{loughran2016textual}, the LM dictionary has been used in various studies to measure the sentiment in newspaper articles and columns such as \cite{tetlock2007giving, dougal2012journalists, gurun2012don}, among others. This well-established lexicon provides a solid foundation for our study, allowing us to pose a well-defined research question. Furthermore, this work is part of a broader research project conducted by our team, which explores the application of advanced NLP techniques in finance-related contexts. Consequently, our primary emphasis lies on sentiment analysis within the field of finance.

We compare the newly created explainable lexicon with the  LM lexicon (known to outperform general-purpose lexicons in financial contexts) on financial datasets to assess the overall potential and performance of the methodology. Our study demonstrates that generated lexicons can improve the accuracy and coverage of lexicons annotated by domain experts, potentially leading to faster and more automated data processing pipelines tailored to productive NLP applications while reducing the manual work needed by domain experts. Additionally, we show that our methodology has a generic architecture and can be applied in other areas beyond financial applications.

Dictionary-based sentiment models have their own advantages and disadvantages. To use a dictionary-based sentiment model, the text to be analyzed is first preprocessed to remove stop words, punctuation, and other non-alphanumeric characters. Then, each word in the preprocessed text is matched against the words in the sentiment dictionary and assigned a sentiment score based on its associated sentiment value. The sentiment scores for each word in the text are then aggregated to obtain an overall sentiment score for the text. This approach is relatively simple and straightforward, as it does not require any training or complex modeling. The sentiment dictionary is fixed and does not change during analysis, making it easy to use and implement. Another advantage of dictionary-based sentiment models is their interpretability. Since the sentiment scores assigned to each word in the dictionary are pre-defined, it is easy to understand why a particular text was classified as positive, negative, or neutral. This can be useful for analyzing the sentiment of text in various applications, such as customer feedback analysis, social media monitoring, and market research. With its inherent interpretability, utilizing lexicons for sentiment analysis can also aid in examining the relationship between the polarity of news articles and the movements of stock prices \cite{dodevska2019predicting}. In addition, dictionary-based sentiment models have low computational requirements, making them suitable for the real-time analysis of high-volume text sources like social media streams. They can be implemented on low-powered devices, such as mobile phones, which is useful for applications that require quick sentiment analysis results. 
Despite their benefits, dictionary-based models also exhibit limitations. They may fail to capture the nuances and complexities of natural languages, such as sarcasm and irony, and may exhibit biases towards certain words or sentiment values. Additionally, these models might be ineffective for analyzing text in multiple languages or domains with specialized terminology.

The paper is organized as follows. In Section \ref{sec:2}, we make a review of the relevant literature. Section \ref{sec:3} describes the methodology and data processing pipeline for extracting words and generating explainable lexicons using transformers and SHAP explainability. In Section \ref{sec:4}, we explain in detail the constituent phases of the pipeline to create an explainable lexicon based on SHAP that is used to expand the standard LM lexicon. We use this explainable lexicon in Section \ref{sec:5} to create a new model for sentiment classification. We demonstrate the effectiveness of our approach in Section \ref{sec:6}, where we show it outperforms the LM lexicon. Specifically, we provide a discussion assessing the performance of the model in sentiment classification tasks on financial datasets. We use the last Section \ref{sec:7} to give concluding remarks and suggest directions for future research.

\section{Related Work}
\label{sec:2}

The process of lexicon-based sentiment analysis has traditionally focused on creating lexicons by manually labeling the sentiment of the words included in the lexicons. While such lexicons are of high quality, they require laborious curation and domain expertise \cite{taboada2011lexicon}. Thus, lexicons created for one domain use specialized vocabulary and may not be suitable or directly applicable to other domains. As the polarity of words may vary across disciplines, domain dependence in sentiment analysis has been emphasized by researchers in the field \cite{loughran2014measuring, loughran2016textual, krishnamoorthy2018sentiment}. \cite{loughran2011liability} showed that word lists curated for other domains misclassify common words in financial texts. For example, the word ``liability'' is considered neutral in finance, but it usually conveys a negative polarity in general-purpose applications, making the reuse difficult in specialized lexicons. In their seminal study \cite{loughran2011liability}, the authors created an expertly annotated lexicon, called the Loughran-McDonald (LM) lexicon, to more accurately capture sentiments in financial texts. Other dictionaries used in finance include General Inquirer (GI) \cite{stone1966general}, Harvard IV-4 (HIV4), and Diction, but their performance is known to be inferior compared to the LM lexicon in sentiment classification tasks in finance.

Given these drawbacks, statistical methods have been proposed for automatic lexicon learning. For example, \cite{mohammad2013nrc} showed that emoticons or hashtags in tweet messages can be used to avoid manual lexicon annotation and to significantly improve lexicon coverage while effectively leveraging the abundance of training data. While \cite{mohammad2013nrc} relied on calculating pointwise mutual information (PMI) between words and emoticons, \cite{vo2016don} uses a simple neural network to train lexicons that improve the accuracy of predicting emoticons in tweets.

The study in \cite{viegas2020exploiting} takes a different approach that proves to be beneficial; it recognizes that supervised solutions can be expensive due to the need to perform burdensome labeling of data. The data labeling process is not only challenging and costly but also suffers from the drawback of producing limited lexicon coverage. Therefore, as its main contribution, \cite{viegas2020exploiting} proved that semantic relationships between words can be effectively used for lexicon expansion, contrary to what has been widely assumed in the semantic analysis literature. Their method uses word embeddings to expand lexicons in the following way: it adds new words whose sentiment values are inferred from ``close'' word vectors that are already present in the lexicon. Surprisingly, the experimental analysis in \cite{viegas2020exploiting} showed that the unsupervised method proposed by the authors is as competitive as state-of-the-art supervised solutions such as transformers (BERT) without having to rely on any training (labeled) data.

Automatic lexicon building has been studied in several papers in the literature. For instance, certain approaches have shown that taking negation into account improves the performance of financial sentiment lexicons on various sentiment classification tasks \cite{bos2021automatically}. Adapting lexicons that depend on word context is studied in \cite{saif2014adapting}; this work captures the context of words appearing in tweet messages and uses it to update their prior sentiment accordingly. The methodology in \cite{saif2014adapting} showed improvement in lexicon performance due to the sentiment adaptation to the underlying context. Earlier works explored various directions such as lexicon generation from a massive collection of web resources \cite{kaji2007building}, automatic lexicon expansion for domain-oriented sentiment analysis \cite{kanayama2006fully}, construction of polarity-tagged corpus from HTML documents \cite{kaji2006automatic}, etc.

Inducing domain-specific sentiment lexicons from small seed words and domain-specific corpora is studied in \cite{hamilton2016inducing}, where it is shown that this approach outperforms methods that rely on hand-curated resources. The approach is validated by showing that it accurately captures the sentiment mood of important economic topics of interest, such as data from the Beige Book of the U.S. Federal Reserve Board (FED) and data from the Economic Bulletin of the European Central Bank (ECB). Combining word embeddings with semantic similarity metrics between words and lexicon vocabulary is shown to better extract subjective sentiment information from lexicons \cite{araque2019semantic}. This paper emphasizes that the capability to infer embedding models automatically leads to higher vocabulary coverage. The experiments in \cite{araque2019semantic} also demonstrate that lexicon words largely determine the performance of the resulting sentiment analysis, meaning that similar lexicons (i.e., with similar vocabulary) result in similar performance.

The comparable performance among lexicons containing similar vocabulary is one of our main reasons to explore the potential of transformers to automatically learn and expand known lexicons in an explainable way. The power of NLP transformers to accurately extract sentiment from financial texts is presented in \cite{mishev2020evaluation}, where the authors perform a comprehensive analysis with more than one hundred experiments to prove the capabilities of transformers, and, in particular, how their word embeddings outperform lexicon-based knowledge extraction approaches or statistical methods.

Due to the complexity of machine learning (ML) techniques, especially deep learning models, the outputs of the models are hard to visualize, explain, and interpret. In recent years, this has given rise to a vast amount of research on the explainability of ML models. A state-of-the-art technique for explainability is considered SHAP (\underline{SH}apley \underline{A}dditive ex\underline{P}lanations), which uses Shapley values from game theory to explain the output of ML models \cite{lundberg2017unified}.

The potential of SHAP is explored in different use cases. SHAP has been recently proven beneficial for diagnosing the explainability of text classification models based on Convolutional Neural Networks (CNNs) \cite{zhao2020shap}. When combined with CNNs, SHAP effectively explains the importance of local features while also taking advantage of CNN's potential to reduce the high feature dimensionality of NLP tasks. CNNs are known to outperform other ML algorithms for text classification, which implies that the SHAP-based analysis of CNNs in \cite{zhao2020shap} can be potentially carried out to explain any text classification tasks. The increased interest in SHAP has also been extended to the financial domain, where SHAP values are used for topics such as interpreting financial time series \cite{mokhtari2019interpreting} and financial data of bankrupt companies \cite{xiaomao2019comparison}. A comprehensive study has been performed in \cite{rizinski2022ethically} to evaluate SHAP in the context of ethically responsible ML in finance. The SHAP method has been adapted for explaining SOTA transformer language models such as BERT to improve the visualizations of the generated explanations \cite{kokalj2021bert}.

Extracting sentiment from news text, social media, and blogs has gained increasing interest in economics and finance. The study in \cite{consoli2022fine} proposes a fine-grained aspect-based sentiment analysis to identify sentiment associated with specific topics of interest in each sentence of a document. Business news texts are used to compile a comprehensive domain-specific lexicon in \cite{moreno2020design}. A hybrid lexicon that combines corpus-based and dictionary-based methods with statistical and semantic measures is proposed in \cite{yekrangi2021financial}, showing that sentiments extracted from a large dataset of financial tweets exhibit a correlation with market trends.

Sentiment analysis of news articles using lexicons has been performed on the BBC news dataset in \cite{taj2019sentiment}. The work outlines the two main lexicon approaches to sentiment analysis, namely dictionary-based and corpus-based methods, but it does not involve machine learning techniques. The study in \cite{fang2011incorporating} recognized that focusing entirely on machine learning by ignoring the knowledge encoded in sentiment lexicons may not be optimal. Thus, the authors presented a method that incorporates domain-specific lexicons as prior knowledge into algorithms such as Support Vector Machine (SVM) and showed that it could improve the accuracy of sentiment analysis tasks.

While acknowledging the advantages of deep learning methods, the results in \cite{catelli2022lexicon} showed that lexicon-based methods are preferred for use cases with low-resource languages or limited computational resources at the expense of slightly lower performance. The authors performed a comparative study between the BERT Base Italian XXL language model and the NooJ-based lexical system with Sentix and SentIta lexicons, thereby validating the idea of using lexicons in use cases with scarce datasets. The paper used SHAP to perform qualitative analysis between the two approaches, but SHAP was not used to improve the coverage of existing lexicons. To the best of our knowledge, SHAP has still not been explored for the purpose of automatic lexicon generation.

\section{The XL\MakeLowercase{ex} Word Extraction Methodology}
\label{sec:3}

The construction of a lexicon for sentiment analysis comprises several consecutive stages, each involving suitable text processing\footnote{The source code and datasets related to the XLex methodology, including all conducted experiments, are accessible on GitHub at the following link: https://github.com/hristijanpeshov/SHAP-Explainable-Lexicon-Model}. To facilitate sentiment analysis, the lexicon must incorporate words from both positive and negative polarities. In this section, we explain the steps involved in generating the positive and negative sentiment sets, which will be merged to form an explainable lexicon.

\begin{figure*}[h]
    \centering
    \includegraphics[width=\textwidth]{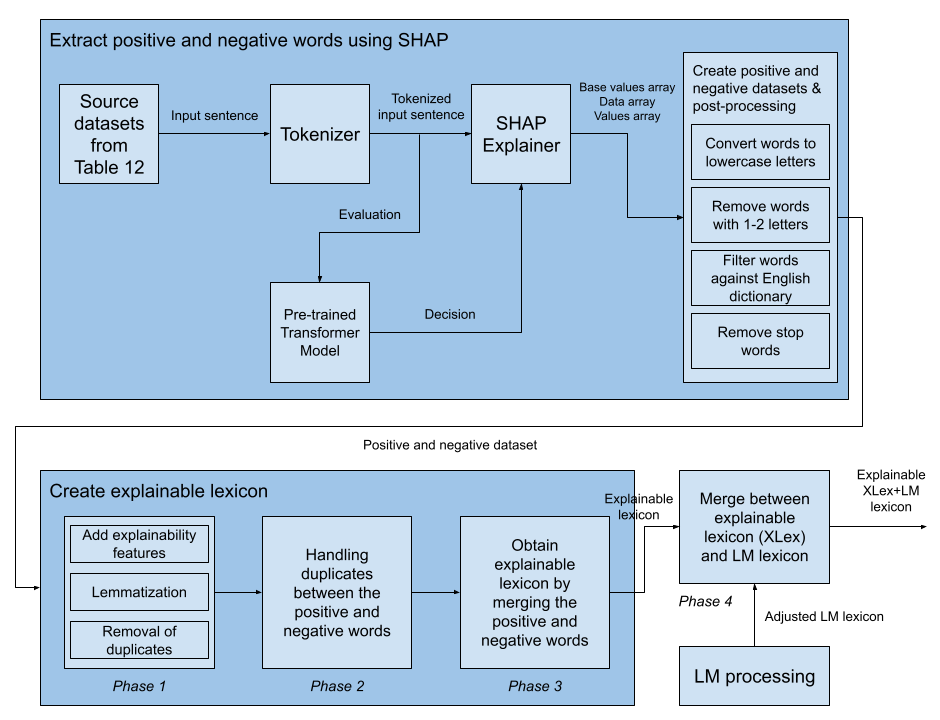}
    \caption{Architecture of the data processing pipeline for generating the explainable lexicon (XLex). The upper section of the figure, labeled as ``Extract positive and negative words using SHAP'', illustrates the word extraction process using SHAP, followed by post-processing steps to generate separate positive and negative word datasets from the chosen source datasets. The lower section of the figure, referred to as ``Create explainable lexicon'', encompasses adding explainability features, handling duplicates, and merging the positive and negative datasets to form the comprehensive explainable lexicon XLex. The pipeline concludes by merging XLex with the Loughran-McDonald (LM) lexicon, resulting in the combined XLex+LM lexicon.}
    \label{fig:data_processing_pipeline}
\end{figure*}

The architecture of the data processing pipeline is depicted in Figure \ref{fig:data_processing_pipeline}. The individual components of the pipeline are elaborated in detail in the following subsections of the paper.

\begin{figure*}[h]
    \centering
    \includegraphics[width=0.8\textwidth]{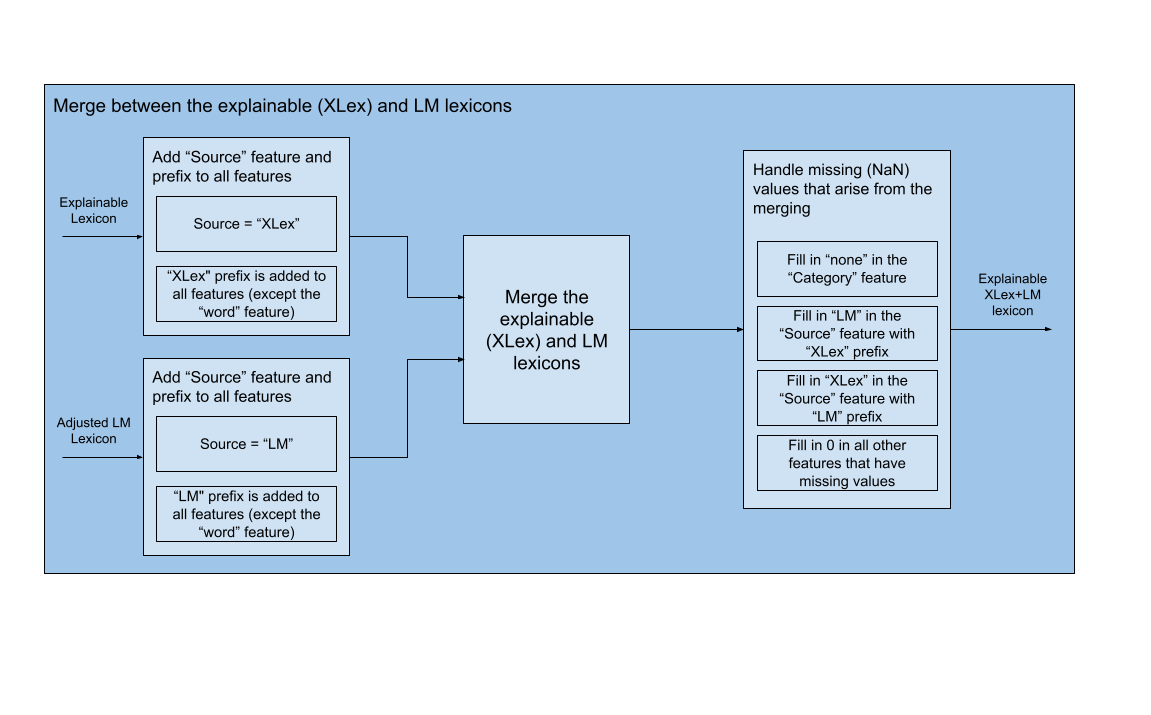}
    \caption{The explainable lexicon (XLex) and the LM lexicon are merged to form the combined XLex+LM lexicon. Before the merging process, the ``Source'' feature is introduced to both XLex and LM, and all features (excluding the ``word'' feature) are appropriately prefixed to enable identification of XLex features as well as LM features within the combined lexicon. Handling of missing values takes place subsequent to the merging.}
    \label{fig:merge_between_explainable_LM_lexicons}
\end{figure*}

\begin{figure*}[h]
    \centering
    \includegraphics[width=0.8\textwidth]{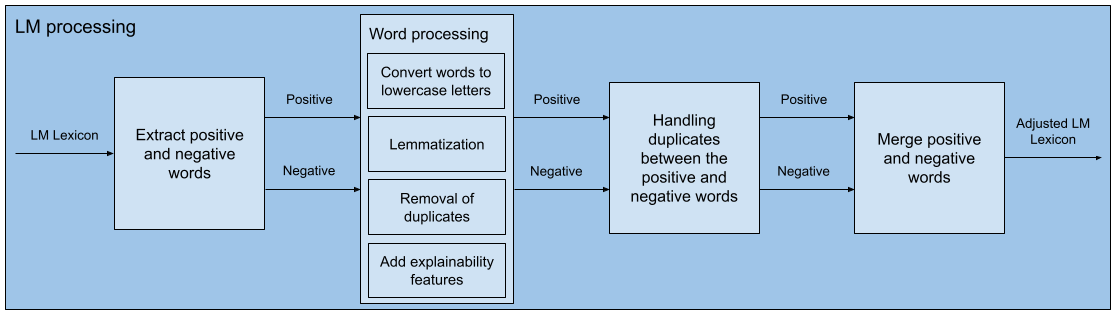}
    \caption{The LM lexicon undergoes a preparatory adjustment process to enable its seamless integration with the explainable XLex lexicon, resulting in the formation of the combined XLex+LM lexicon. This adjustment process includes the extraction of positive and negative words, subsequent word processing, handling of duplicates, and the final step of merging the positive and negative word sets.}
    \label{fig:LM_processing}
\end{figure*}

\begin{figure}[h]
    \centering
    \includegraphics[width=0.3\textwidth]{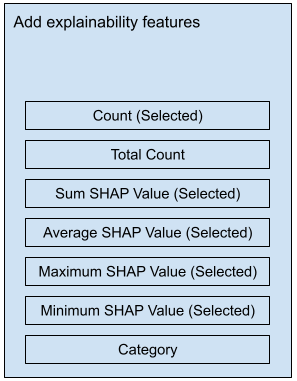}
    \caption{A list of explainability features based on SHAP added in the explainable and LM lexicons. For the LM lexicon, all features except ``Category'' are assigned the value of 1 as their default value.}
    \label{fig:add_explainability_features}
\end{figure}

\begin{figure*}[h]
    \centering
    \includegraphics[width=0.8\textwidth]{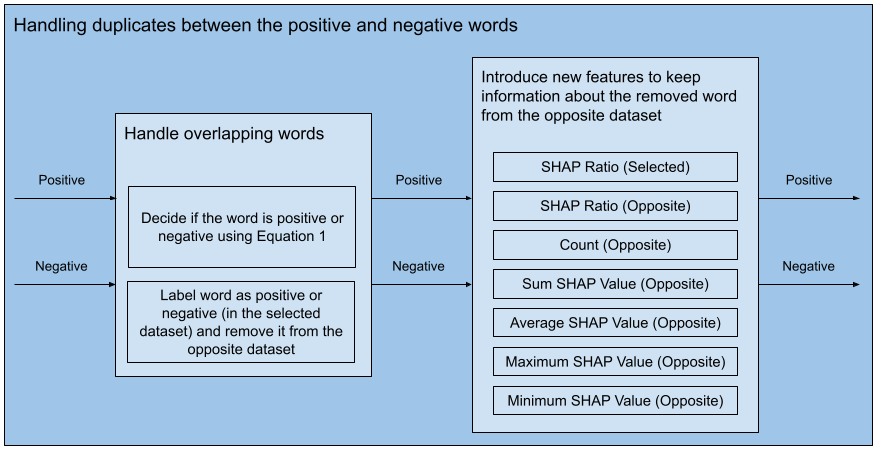}
    \caption{The process of dealing with duplicate entries between the positive and negative words for each of the explainable (XLex) and LM lexicons. In the case of the LM lexicon, features designated as ``Opposite'' are assigned a default value of 0. The ``Total Count'' feature can be obtained by deriving it from the values of ``Count (Selected)'' and ``Count (Opposite)''.}
    \label{fig:handling_duplicates}
\end{figure*}

\subsection{A Transformer-Based Model for Sentiment Analysis}
\label{subsec:3.1}

To develop an explainable lexicon, we begin by using a transformer-based model tailored for sentiment analysis. Specifically, FinBERT is a notable model in this domain, designed explicitly for analyzing financial texts \cite{huang2023finbert}. However, FinBERT is fine-tuned on a closed dataset comprising 10,000 sentences from analyst reports sourced from Thompson Reuters's proprietary Investext database. To ensure a controlled environment over the fine-tuning process, we take charge of it by using a publicly available financial sentiment dataset and one of the available base transformer models, specifically RoBERTa. According to a survey \cite{mishev2020evaluation}, the RoBERTa model demonstrates exceptional performance in various finance-related sentiment classification tasks, achieving an accuracy of 94\%. Therefore, we decided to use RoBERTa as our starting model.

To provide a comprehensive analysis, we also included results obtained using the FinBERT model. While both models produce comparable results (as shown in Table \ref{table:results_from_models}), the key advantage of using the fine-tuned RoBERTa model lies in the customized approach and rigorous control we have over the data and fine-tuning process, thereby boosting our flexibility to perform experiments.

The datasets utilized for learning the explainable dictionaries are presented in Table \ref{table:datasets_statistics}, where they are labeled as ``Source'' datasets. The results of the XLex model are then evaluated using the datasets labeled as ``Evaluation''.

To construct the sentiment dictionaries, we use SHAP to interpret the output of the pre-trained transformer model. This approach aids us in identifying individual words and classifying them as either positive or negative in sentiment. This approach is discussed in detail in Subsection \ref{subsec:3.2}.

The RoBERTa-Large model is originally trained in a self-supervised manner on an extensive corpus of English text\footnote{For a comprehensive explanation of the pretraining methodology employed for RoBERTa-Large, the reader is directed to the official Hugging Face documentation: https://huggingface.co/roberta-large}. The RoBERTa-Large model comprises 24 layers, 1024 hidden units, 16 attention heads and is based on a total of 355 million parameters. Subsequently, we fine-tune the foundational RoBERTa model using the approach outlined in \cite{mishev2020evaluation}. The fine-tuning is conducted on a merged dataset comprising the Financial PhraseBank \cite{malo2013learning} and SemEval-2017-Task5 datasets \cite{cortis2017semeval}. This fine-tuning procedure ensures that the resulting model is specialized for the domain of financial sentiment analysis. This fine-tuned model will also be referred to as the RoBERTa-based model for convenience in the following sections.

These two constituent datasets are composed of financial headlines extracted from two different sources. The sentences in the Financial PhraseBank corpus are selected using random sampling from English news on all listed companies in the OMX Helsinki stock index. The sampling is performed to ensure that the selected sentences represent both small and large companies, different industries as well as different news sources. The dataset contains 4846 sentences annotated with three polarities: positive, negative, and neutral. On the other hand, SemEval-2017-Task5 is the dataset used for the ``Fine-Grained Sentiment Analysis'' problem posed by Task 5 of the SemEval 2017 competition. It consists of approximately 1200 news headlines related to large companies operating worldwide. The headlines are extracted from various internet sources, including Yahoo Finance. The sentiment score of each sentence in the dataset is labeled with a real number ranging from -1 to 1. A summary of the statistics of the two datasets is given in Table \ref{table:phrasebank_semeval}.

As illustrated in Table \ref{table:phrasebank_semeval}, there is an imbalance between the number of positive and negative sentences in both datasets. The number of neutral sentences also differs drastically when compared to the number of positive or negative sentences. To address the problem, balancing is performed by extracting 1093 positive and 1093 negative sentences, which are then merged into one dataset. This dataset is used for training and evaluation of the model that we take from \cite{mishev2020evaluation}. The sentences in the dataset are shuffled and divided into 80\% training set and 20\% testing set. The training and testing sets contain 1748 and 438 sentences, respectively. Both the training and test sets are balanced, i.e., they contain the same number of positive and negative sentences. The statistics of the resulting dataset are shown in Table \ref{table:train_test_sets_statistics}.

\subsection{Extracting Words and Their Analysis With SHAP}
\label{subsec:3.2}

The first step in creating the lexicon involves extracting words from financial sentences and labeling them as positive or negative. For this purpose, we use the previously introduced model together with a tokenizer. The model classifies the sentiment of the input sentences, while the tokenizer deals with tokenization, i.e., dividing the sentences into component words. The model and tokenizer are then passed to the SHAP explainer, which generates explanations for the model decisions.

SHAP is considered a state-of-the-art technique for ML model explainability \cite{mazzanti2020shap}. Its approach uses Shapley values from game theory to explain the output of ML models \cite{shap}. Game theory is characterized by two elements: a game and players. In SHAP, the game consists of reproducing the results of the model being explained (in our case, that is, the NLP model for sentiment analysis), while the players are the features (the financial statement, i.e., its constituent words) that are passed as input to the model. SHAP evaluates the contribution of each feature to the model predictions and assigns each feature an importance value, called a SHAP value. SHAP values are calculated for each feature across all samples of the dataset to assess the contribution of individual features to the model's output \cite{lundberg2017unified}. It is important to note that SHAP explains the predictions locally, meaning that the contributions of the features (words) on the model prediction are related to a specific sample in the dataset. A different sample can yield other values for the features' contributions. However, due to the additive nature of SHAP values, it is also possible to aggregate them, allowing us to calculate global values for the overall contribution of the features across all samples.

\begin{table}[!ht]
\caption{Polarity distribution of sentences in the Financial PhraseBank and SemEval-2017-Task5 datasets.}
\centering
\begin{tabular}{|l|cc|}
\hline
\multicolumn{1}{|c|}{Sentiment}      & \multicolumn{2}{c|}{Dataset} \\ \cline{2-3} 
                           & \multicolumn{1}{l|}{Financial PhraseBank} & \multicolumn{1}{l|}{SemEval2017-Task5} \\ \hline
Neutral                    & \multicolumn{1}{c|}{2879}                 & 38                                     \\ \hline
Negative                   & \multicolumn{1}{c|}{604}                  & 451                                    \\ \hline
Positive                   & \multicolumn{1}{c|}{1363}                 & 654                                    \\ \hline
Total                      & \multicolumn{1}{c|}{4846}                 & 1143                                   \\ \hline
\end{tabular}
\label{table:phrasebank_semeval}
\end{table}

\begin{table}[!ht]
\caption{Statistics of the train and test sets used for fine-tuning the initial RoBERTa-Large model.}
\centering
\begin{tabular}{|l|ccc|}
\hline
\multicolumn{1}{|c|}{Polarity} & \multicolumn{3}{c|}{Financial PhraseBank \& SemEval2017-Task5}               \\ \cline{2-4} 
                          & \multicolumn{1}{c|}{Train set} & \multicolumn{1}{c|}{Test set} & Total \\ \hline
Negative                  & \multicolumn{1}{c|}{874}       & \multicolumn{1}{c|}{219}            & 1093  \\ \hline
Positive                  & \multicolumn{1}{c|}{874}       & \multicolumn{1}{c|}{219}            & 1093  \\ \hline
Total                     & \multicolumn{1}{c|}{1748}      & \multicolumn{1}{c|}{438}            & 2186  \\ \hline
\end{tabular}
\label{table:train_test_sets_statistics}
\end{table}

To evaluate the features' contributions to the model prediction for a given sample, SHAP creates a copy of the model for each combination of the input features. Each of these models is the same, the only difference is the combination of features passed to the model. In one of these combinations, none of the features is passed to the model. In that case, the model results in a mean value for the prediction; the value is obtained by averaging the labels of the dataset on which the model was trained. This value is called a \textit{base value}. The base value is the value that would be predicted if no features are known for the model’s current output \cite{lundberg2017unified}. In this way, by adding a certain feature to the input, the SHAP explainer can record the changes to the predicted value and can measure the contribution of that feature. Each of the features can increase or decrease the predicted value. Finally, to obtain the value predicted by the model (when all features are present), SHAP aggregates the contribution values (which can be positive or negative) for each feature and superposes the result to the base value (the prediction with no input features provided). Using this process, SHAP explains the contribution (importance) of each feature in a given sample. In other words, SHAP measures the difference in the predicted value caused by the presence or absence of a feature. The additive nature of the aggregation is where the name SHAP comes from, namely Shapley Additive exPlanations.

The input parameter passed to the SHAP explainer is a sentence. The NLP model evaluates the sentiment of the sentence by making a sentiment classification decision, while SHAP provides an explanation for the decision. The explanation of the SHAP explainer returns three arrays: \textit{base values array}, \textit{data array}, and \textit{values array}. The base values array contains two numeric values: a base value for the positive class and a base value for the negative class. These base values represent the values that would be predicted for a particular sentence if no input features are known. In this case, it is the mean value of the labels for each of the classes obtained across all instances (samples) on which the model was trained. The data array contains the tokens (the constituent words), which are obtained by applying the tokenizer to the input sentence. The elements of the values array represent the weights, that is, the contribution of each of the words (tokens) in the calculation of the sentiment of the sentence. The weights in the values array are real numbers ranging from -1 to 1. The weights represent the importance of a particular word (token) and its contribution to the final value predicted by the sentiment classification model. The data array and the values array have the same number of elements.

As mentioned earlier, the weights are additive, which allows them to be superposed. By adding the weights to the base value, the explainer arrives at the value predicted by the sentiment model. Visually, this superposition is represented using diagrams where the calculated weights ``push'' the base value to the ``right'' or to the ``left'', causing the model to increase or decrease its predicted value. By doing so, it is possible to explain how the model arrived at a given decision and how different parts of a sentence contributed to the model's output. Specifically, in terms of sentiment analysis with SHAP, this helps understand why a given NLP model classified a sentence as positive or negative and how each of the constituent words of the sentence contributed to that classification decision.

\begin{figure*}[!ht]
    \centering
    \includegraphics[width=\textwidth]{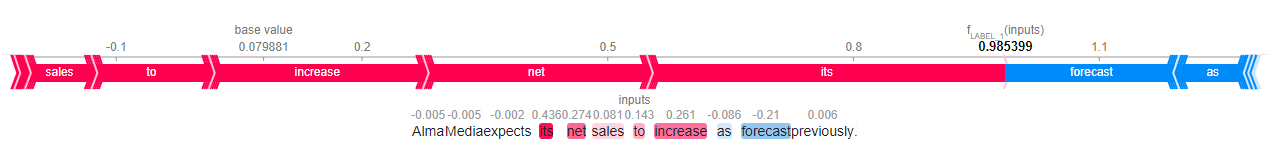}
    \caption{An example of using SHAP for evaluating the word contributions to the sentiment of a sentence.}
    \label{fig:shap_explanation_plot}
\end{figure*}

A visual example is given in Figure \ref{fig:shap_explanation_plot}. The figure shows that positive importance values, marked red, ``push'' the base value to the ``right'' (increasing the model's predicted value), while negative weights, marked blue, ``push'' the base value to the ``left'' (reducing the model's predicted value). The example is visualized from the perspective of the positive class, meaning that each "push" to the "right" increases the probability of predicting a positive sentiment for the given sentence. Each "push" to the "left" decreases this probability, that is, it increases the probability of predicting a negative sentiment for the sentence. Using these preliminaries about the SHAP explainer, we will next create two sets containing positive and negative words as explained in Subsection \ref{subsec:3.3}.

\subsection{Creating a Positive and Negative Dataset and Their Postprocessing}
\label{subsec:3.3}

The sentiment classification in the previous subsection is performed on datasets containing financial sentences. These datasets are denoted as source datasets in Table \ref{table:datasets_statistics}.  Using SHAP, each of the words in a sentence is marked as positive or negative in the given context. The decision to label a particular word as positive or negative depends on whether it contributes with a positive or negative weight to the final decision of the model. As a result, two new datasets are generated. One dataset contains all words across all sentences that contribute to the positive sentiment of each sentence (we refer to the words and dataset as ``positive'' words and ``positive'' dataset, respectively), while the other dataset contains all words across all sentences that contribute to the negative sentiment (``negative'' words, ``negative'' set). In addition to the words themselves, these datasets store a few additional parameters for each word, such as the mean value of the weights (importance values) obtained by the SHAP explainer for all of the word appearances, the sum of these values as well as their maximum and minimum values. The datasets also store the total number (count) of appearances of each of the words. All numerical entries in the datasets are represented by their absolute value.

After creating the positive and negative datasets, we perform post-processing to filter the extracted words. The goal is to keep only the words that are valid and have meaning. The word post-processing process is explained as follows.

The post-processing begins by transforming all words into lowercase letters. Then, all entries consisting of one or two letters are removed since they are of little sentiment utility to the datasets. These entries are typically fragments of words that are obtained due to the limitations of the tokenizer. The RoBERTa tokenizer is limited by the size and coverage of the vocabulary that is used to train the tokenizer. This leads to incorrect or imprecise tokenization of certain words that are either not sufficiently represented in the training vocabulary or are not represented at all. As a result, these words are not accurately represented since they are divided into parts based on more common entries found in the tokenizer's vocabulary. Thus, entries with one or two letters are deemed unnecessary and are removed due to their insufficient contribution to the sentiment analysis.

To obtain valid and useful words, we apply another filter to the datasets. Using a dictionary of English words, we remove all words that are not contained in the English dictionary. This is done to address the limitation of the tokenizer and also to provide a dataset containing only valid words. The last step in the post-processing removes auxiliary words that do not carry meaning in the sentence, such as adverbs, prepositions, pronouns, and conjunctions (stop words).

These preliminary steps and the data stored for each word are necessary to develop an explainable lexicon, which will be shown in Section \ref{sec:4}.

\section{XL\MakeLowercase{ex} Lexicon Creation Methodology}
\label{sec:4}

In the previous section, we demonstrated the use of a transformer model for sentiment analysis in combination with SHAP to process finance-related sentences, which resulted in the creation of two datasets. One dataset contains all words with positive sentiment in a given context (positive dataset), while the other contains all words with negative sentiment in a given context (negative dataset). These two datasets are used to create an explainable lexicon, as will be shown later on. We will evaluate the performance of the explainable lexicon employing the model proposed in the subsequent Section \ref{sec:5}. The evaluation results are presented in Section \ref{sec:6}.

This section explains in detail the methodology for generating the explainable lexicon as well as the process of merging it with the LM lexicon. The methodology encompasses four phases, as shown in Figure \ref{fig:data_processing_pipeline}.

\subsection{Phase 1: Lemmatization and Removal of Duplicate Words Within the Positive and Negative Datasets}

In this phase, we lemmatize the words in the positive and negative datasets obtained in Section \ref{sec:3}. As a result of the lemmatization, each of the words is replaced by its lemma, that is, by its basic form. The goal is to bring different forms of a certain word to their common lemma, thereby avoiding different interpretations of the same word. However, this causes duplicate words to appear in the datasets. After their lemmatization, different forms of a word (which until that moment are uniquely represented) can have the same lemma. The purpose of Phase 1 is to make the datasets consistent by removing duplicate words. Avoiding duplicates will also result in a single source of information related to a particular word.

Each of the duplicates may result in different values for the number of appearances, the average SHAP value, the sum SHAP value, as well as for the maximum and minimum SHAP values. Thus, the goal is to merge the duplicates so that each word is characterized by a single (unique) value for each of these features. The removal of duplicates is performed separately in each of the two datasets (the positive and the negative set). As will be shown, there are words that are labeled both as positive and negative, i.e., words that are present in both datasets. Dealing with these duplicates between the two datasets is done in Phase 2.

To calculate the unique values for the features of a particular word, it is necessary to aggregate the values of the features across all duplicates. The method for aggregating the duplicates for each feature (column) is shown in Table \ref{table:aggregation_functions}. The aggregation function represents how the values of all duplicates are combined (aggregated) for a certain feature. From the table, it can be seen that the feature indicating the number of appearances of the word is obtained as a summation of the number of appearances for each of the duplicates. The reason for this is that each of the duplicates represents the same word after lemmatization, so the number of occurrences of that word will be represented as the sum of all the occurrences of the duplicates. The same approach is applied to calculate the total (sum) SHAP value. After lemmatization, all duplicates of a word have the same form, so the sum SHAP value of all occurrences is the sum of the values of this feature across all duplicates. The maximum value is represented by an aggregation function that takes the maximum along this column for all duplicates. The maximum SHAP value of all duplicates is a suitable representative of the maximum SHAP value of that word. The minimum SHAP value is handled similarly. To obtain the minimum SHAP value for the word, it is necessary to aggregate it with a function that calculates the minimum SHAP value from all duplicates. As can be seen from Table \ref{table:aggregation_functions}, no aggregation is performed for the feature ``Average SHAP value'' because it is obtained by dividing the sum SHAP value by the number of occurrences of the word.

\begin{table*}[]
\caption{Aggregation functions to handle duplicates across the numerical features of the sentiment dataset. No aggregation is performed for the average SHAP value as it is obtained by dividing the sum SHAP value by the total number of word appearances. Another feature that is not aggregated is Category since it is a categorical variable.}
\centering
\begin{tabular}{|l|lllll|}
\hline
\multicolumn{1}{|c|}{}    & \multicolumn{5}{c|}{Features}                                                                                                                                                                                                                                                                                                                                                                                \\ \cline{2-6} 
                     & \multicolumn{1}{l|}{\begin{tabular}[c]{@{}l@{}}Number of\\ apperances\end{tabular}} & \multicolumn{1}{l|}{\begin{tabular}[c]{@{}l@{}}Sum\\ SHAP value\end{tabular}} & \multicolumn{1}{l|}{\begin{tabular}[c]{@{}l@{}}Average\\ SHAP value\end{tabular}} & \multicolumn{1}{l|}{\begin{tabular}[c]{@{}l@{}}Maximum\\ SHAP value\end{tabular}} & \begin{tabular}[c]{@{}l@{}}Minimum\\ SHAP value\end{tabular} \\ \hline
Aggregation function & \multicolumn{1}{l|}{Sum}                                                            & \multicolumn{1}{l|}{Sum}                                                        & \multicolumn{1}{l|}{N/A}                                                          & \multicolumn{1}{l|}{Max}                                                          & Min                                                          \\ \hline
\end{tabular}
\label{table:aggregation_functions}
\end{table*}

Table \ref{table:handling_duplicates_positive_dataset} illustrates an example of merging duplicate words in the positive dataset and getting unique values for the features. The example presents three different words that result in the same lemma after performing lemmatization, demonstrating how duplicate words are handled. In order to have only one instance of the word ``acquire'' in the dataset, it is necessary to merge these three instances into one. This is done as per the definition of the aggregation functions given in Table \ref{table:aggregation_functions}. In the example, the sum is calculated based on the number of occurrences of all duplicates (9 + 4 + 5 = 18), which results in a total of 18 occurrences of the word ``acquire''. The sum of the sum SHAP values across all duplicates (3.05 + 1.4 + 0.88 = 5.33) represents a sum SHAP value of 5.33 for the word. To get the average SHAP value, it is necessary to divide the sum SHAP value by the number of occurrences (5.33 ÷ 18 = 0.3), which gives an average SHAP value of 0.3 for the word ``acquire''. The maximum SHAP value for the word is 0.6, while the minimum SHAP value is 0.02.

\begin{table*}[]
\caption{An example for aggregating duplicates in the positive dataset. Duplicates are handled similarly in the negative dataset.}
\centering
\begin{tabular}{|ll|lllll|}
\hline
\multicolumn{2}{|c|}{Duplicates}                                                                            & \multicolumn{5}{c|}{Features}                                                                                                                                                                                                                                                                                                                                                                                         \\ \hline
\multicolumn{1}{|l|}{\begin{tabular}[c]{@{}l@{}}Original\\ word\end{tabular}}           & Lemma             & \multicolumn{1}{l|}{\begin{tabular}[c]{@{}l@{}}Number of\\ appearances\end{tabular}} & \multicolumn{1}{l|}{\begin{tabular}[c]{@{}l@{}}Total\\ SHAP\\ value\end{tabular}} & \multicolumn{1}{l|}{\begin{tabular}[c]{@{}l@{}}Average\\ SHAP\\ value\end{tabular}} & \multicolumn{1}{l|}{\begin{tabular}[c]{@{}l@{}}Maximum\\ SHAP\\ value\end{tabular}} & \begin{tabular}[c]{@{}l@{}}Minimum\\ SHAP\\ value\end{tabular} \\ \hline
\multicolumn{1}{|l|}{acquire}                                                           & acquire           & \multicolumn{1}{l|}{9}                                                               & \multicolumn{1}{l|}{3.05}                                                         & \multicolumn{1}{l|}{0.34}                                                           & \multicolumn{1}{l|}{0.6}                                                            & 0.05                                                           \\ \hline
\multicolumn{1}{|l|}{acquired}                                                          & acquire           & \multicolumn{1}{l|}{4}                                                               & \multicolumn{1}{l|}{1.4}                                                          & \multicolumn{1}{l|}{0.35}                                                           & \multicolumn{1}{l|}{0.5}                                                            & 0.23                                                           \\ \hline
\multicolumn{1}{|l|}{acquiring}                                                         & acquire           & \multicolumn{1}{l|}{5}                                                               & \multicolumn{1}{l|}{0.88}                                                         & \multicolumn{1}{l|}{0.18}                                                           & \multicolumn{1}{l|}{0.43}                                                           & 0.02                                                           \\ \hline
\multicolumn{2}{|l|}{\begin{tabular}[c]{@{}l@{}}Single instance\\ after aggregation (acquire)\end{tabular}} & \multicolumn{1}{l|}{18}                                                              & \multicolumn{1}{l|}{5.33}                                                         & \multicolumn{1}{l|}{0.3}                                                            & \multicolumn{1}{l|}{0.6}                                                            & 0.02                                                           \\ \hline
\end{tabular}
\label{table:handling_duplicates_positive_dataset}
\end{table*}

The method demonstrated in this phase is applied to each of the two datasets separately. The example above shows how this process is performed for one word in the positive dataset, but the same procedure is used for all other duplicate words in that dataset, as well as for all duplicate words in the negative dataset. This is indicated with the elements ``Lemmatization'' and ``Removal of duplicates'' in Figure \ref{fig:data_processing_pipeline}.

\subsection{Phase 2: Handling of Duplicate Words Between Datasets}

A particular word can be present in both the positive and negative datasets, leading to word overlaps between the datasets. Given our goal to generate a lexicon as a combination of the two datasets, each word should be represented by a single instance in the resulting lexicon. To overcome the overlaps, we use the following approach. If an overlapping word has a higher sum SHAP value in the positive dataset ($SHAP_{sum}^{pos}$) when compared to the negative one ($SHAP_{sum}^{neg}$), then the word is labeled as positive. Similarly, if $SHAP_{sum}^{neg}$ is higher than or equal to $SHAP_{sum}^{pos}$, then the word is labeled as negative. The decision criteria are shown in Equation \ref{eq:1}:

\begin{equation}
    selected \: dataset = 
    \begin{cases}
    \text{positive},    & SHAP_{sum}^{pos} > SHAP_{sum}^{neg} \\
    \text{negative},    & otherwise
    \end{cases}
    \label{eq:1}
\end{equation}

If a certain word is labeled as positive or negative (in the selected dataset), it is removed from the opposite dataset. To keep the information about the word removed from the opposite dataset, new columns are introduced in the datasets. The new columns are given in Table \ref{table:word_features_in_lexicons} under ``Features added in Phase 2''. A complete representation of the words in the two datasets, including their features from both polarities, is achieved by adding these columns. Using Equation \ref{eq:1} as a decision criterion and keeping information about the word removed from the opposite dataset is shown in Figure \ref{fig:handling_duplicates}.

Table \ref{table:word_features_in_lexicons} shows the features added in Phase 2 in addition to the existing features of the datasets. The table also consolidates brief explanations for each of the features. It should be noted that the label ``opposite'' represents the set that was not selected during the decision, in accordance with Equation \ref{eq:1}. Thus, if Equation \ref{eq:1} decides that an overlapping word belongs to the positive dataset, in that case, the ``opposite'' dataset is the negative dataset. This word is removed from the negative dataset, and all its values from the negative dataset are placed in the positive dataset, in the corresponding columns marked as ``opposite''. Similarly, if the decision criteria decide that the word belongs to the negative dataset, in that case, the ``opposite'' represents the positive dataset. This word is removed from the positive dataset, and all its values from the positive dataset are placed in the negative set in the corresponding columns marked ``opposite''.

\begin{table*}[]
\caption{Features of the words in the lexicons.}
\centering
\begin{tabular}{|ccc|}
\hline
\multicolumn{1}{|c|}{Feature name} &
  \multicolumn{1}{c|}{Feature notation} &
  Feature description \\ \hline
\multicolumn{3}{|c|}{Word feature} \\ \hline
\multicolumn{1}{|c|}{Word} &
  \multicolumn{1}{c|}{$word$} &
  A word that appears in the lexicon \\ \hline
\multicolumn{3}{|c|}{Initial features} \\ \hline
\multicolumn{1}{|c|}{Count (Selected)} &
  \multicolumn{1}{c|}{$count$} &
  Number of appearances of the word \\ \hline
\multicolumn{1}{|c|}{Sum SHAP Value (Selected)} &
  \multicolumn{1}{c|}{$SHAP_{sum}$} &
  Sum SHAP value of the word \\ \hline
\multicolumn{1}{|c|}{Average SHAP Value (Selected)} &
  \multicolumn{1}{c|}{$SHAP_{avg}$} &
  Average SHAP value of the word \\ \hline
\multicolumn{1}{|c|}{Maximum SHAP Value (Selected)} &
  \multicolumn{1}{c|}{$SHAP_{max}$} &
  Maximum SHAP value of the word \\ \hline
\multicolumn{1}{|c|}{Minimum SHAP Value (Selected)} &
  \multicolumn{1}{c|}{$SHAP_{min}$} &
  Minimum SHAP value of the word \\ \hline
\multicolumn{3}{|c|}{Features added in Phase 2} \\ \hline
\multicolumn{1}{|c|}{Total Count} &
  \multicolumn{1}{c|}{$count_{total}$} &
  \begin{tabular}[c]{@{}c@{}}Total number of appearances of the\\ word (in the two sentiments)\end{tabular} \\ \hline
\multicolumn{1}{|c|}{Count (Opposite)} &
  \multicolumn{1}{c|}{$count_{opp}$} &
  \begin{tabular}[c]{@{}c@{}}Total number of appearances\\ in the opposite sentiment\end{tabular} \\ \hline
\multicolumn{1}{|c|}{Sum SHAP Value (Opposite)} &
  \multicolumn{1}{c|}{$SHAP_{sum}^{opp}$} &
  \begin{tabular}[c]{@{}c@{}}Sum SHAP value of the word\\ in the opposite sentiment\end{tabular} \\ \hline
\multicolumn{1}{|c|}{Average SHAP Value (Opposite)} &
  \multicolumn{1}{c|}{$SHAP_{avg}^{opp}$} &
  \begin{tabular}[c]{@{}c@{}}Average SHAP value of the word\\ in the opposite sentiment\end{tabular} \\ \hline
\multicolumn{1}{|c|}{Maximum SHAP Value (Opposite)} &
  \multicolumn{1}{c|}{$SHAP_{max}^{opp}$} &
  \begin{tabular}[c]{@{}c@{}}Maximum SHAP value of the word\\ in the opposite sentiment\end{tabular} \\ \hline
\multicolumn{1}{|c|}{Minimum SHAP Value (Opposite)} &
  \multicolumn{1}{c|}{$SHAP_{min}^{opp}$} &
  \begin{tabular}[c]{@{}c@{}}Minimum SHAP value of the word\\ in the opposite sentiment\end{tabular} \\ \hline
\multicolumn{1}{|c|}{SHAP Ratio (Selected)} &
  \multicolumn{1}{c|}{$SHAP_{ratio}$} &
  \begin{tabular}[c]{@{}c@{}}Ratio between $SHAP_{avg}$\\ and the sum of $SHAP_{avg}$ and $SHAP_{avg}^{opp}$\end{tabular} \\ \hline
\multicolumn{1}{|c|}{SHAP Ratio (Opposite)} &
  \multicolumn{1}{c|}{$SHAP_{ratio}^{opp}$} &
  \begin{tabular}[c]{@{}c@{}}Ratio between $SHAP_{avg}^{opp}$\\ and the sum of $SHAP_{avg}$ and $SHAP_{avg}^{opp}$\end{tabular} \\ \hline
\multicolumn{3}{|c|}{Features added in Phase 3} \\ \hline
\multicolumn{1}{|c|}{Category} &
  \multicolumn{1}{c|}{$category$} &
  Category of the word (positive or negative) \\ \hline
\multicolumn{3}{|c|}{Features added in Phase 4} \\ \hline
\multicolumn{1}{|c|}{Source} &
  \multicolumn{1}{c|}{$src$} &
  \begin{tabular}[c]{@{}c@{}}Source lexicon from which\\ the word originates from\end{tabular} \\ \hline
\end{tabular}
\label{table:word_features_in_lexicons}
\end{table*}

If a word is decided to belong to the positive set, then the SHAP ratio ($SHAP_{ratio}$) is calculated as the ratio between the average SHAP value of the word from the positive dataset and the sum of the average SHAP values of the word from the positive and negative datasets. This is shown in Equation \ref{eq:2}:
\begin{equation}
    SHAP_{ratio} = \frac{SHAP_{avg}^{pos}}{SHAP_{avg}^{pos} + SHAP_{avg}^{neg}}
    \label{eq:2}
\end{equation}
The opposite value of the SHAP ratio is expressed as the ratio between the average SHAP value of the word from the opposite dataset (in this case, it is the negative dataset) and the sum of the average SHAP values of the word from the positive and negative sets. This is shown in Equation \ref{eq:3}.
\begin{equation}
    SHAP_{ratio}^{opp} = \frac{SHAP_{avg}^{neg}}{SHAP_{avg}^{pos} + SHAP_{avg}^{neg}} = 1 - SHAP_{ratio}
    \label{eq:3}
\end{equation}
Similar steps are taken if the word is decided to belong to the negative dataset. The only difference is that $SHAP_{ratio}$ is calculated based on the average SHAP value of the negative dataset ($SHAP_{avg}^{neg}$), while $SHAP_{ratio}^{opp}$ is calculated based on the average SHAP value of the positive set ($SHAP_{avg}^{pos}$).

Table \ref{table:word_features_example} shows an illustrative example with the word ``option'' that appears in both datasets (positive and negative). As can be seen, this word has a sum SHAP value in the positive and negative dataset of $SHAP_{sum}^{pos} = 0.39$ and $SHAP_{sum}^{neg} = 0.023$, respectively. Given that $SHAP_{sum}^{pos} > SHAP_{sum}^{neg}$, it is decided that the word belongs to the positive dataset and is removed from the negative dataset. Before removing the word from the negative dataset, the values of its features from the negative dataset are added to the positive dataset in the corresponding columns labeled as ``opposite''. The values added in the “opposite” columns are given as follows: $count_{opp} = 7$, $SHAP_{sum}^{opp} = 0.023$, $SHAP_{avg}^{opp} = 0.0033$, $SHAP_{max}^{opp} = 0.009$, $SHAP_{min}^{opp} = 0.0001$. The SHAP ratio of the word in the two datasets is calculated as $SHAP_{ratio} = \frac{0.026}{0.026+0.0033} = 0.887$; $SHAP_{ratio}^{opp} = \frac{0.0033}{0.026+0.0033} = 0.113$. All other features of the word in the selected (positive) dataset remain unchanged.

\begin{table*}[]
\caption{An example with the word ``option'' and its features in the dataset.}
\centering
\begin{tabular}{|c|ccccc|}
\hline
Word                     & \multicolumn{5}{c|}{option}                                                                                                       \\ \hline
\multicolumn{1}{|c|}{Dataset} & \multicolumn{5}{c|}{Features (columns)}                                                                                           \\ \cline{2-6} 
                         & \multicolumn{1}{c|}{$count$} & \multicolumn{1}{c|}{$SHAP_{sum}$} & \multicolumn{1}{c|}{$SHAP_{avg}$} & \multicolumn{1}{c|}{$SHAP_{max}$} & $SHAP_{min}$ \\ \hline
Positive                 & \multicolumn{1}{c|}{15}    & \multicolumn{1}{c|}{0.39}    & \multicolumn{1}{c|}{0.026}   & \multicolumn{1}{c|}{0.2}     & 0.07    \\ \hline
Negative                 & \multicolumn{1}{c|}{7}     & \multicolumn{1}{c|}{0.023}   & \multicolumn{1}{c|}{0.0033}  & \multicolumn{1}{c|}{0.009}   & 0.0001  \\ \hline
\end{tabular}
\label{table:word_features_example}
\end{table*}

If a word appears only in one of the datasets, a zero value is assigned to each of the features labeled as ``opposite'' because that word does not appear in the opposite sentiment. Also, according to Equation \ref{eq:2}, $SHAP_{ratio}$ evaluates to 1 since $SHAP_{avg}^{opp}$ for the corresponding word is 0.

\subsection{Phase 3: Merging the Positive and Negative Datasets}

In this phase, the two datasets, positive and negative, are merged into a single dataset. The feature ``Category'' is important for the merging. Possible values for this feature are ``positive'' and ``negative'', depending on whether the word is in the positive or negative dataset. All words from the positive dataset have ``positive'' as the value for this feature, while all words from the negative dataset have the value ``negative''. The purpose of the ``Category'' feature is to delineate positive words from negative words in the resulting explainable lexicon. Using this feature, the two datasets are merged by simply adding all the data points (i.e., all words together with all their features) from the negative set to the positive one. An excerpt of the explainable lexicon after the merge is shown in Table \ref{table:excerpt_explainable_lexicon}.\footnote{To ensure the diagrams fit within the page limits, only a subset of the dataset features are depicted in this and subsequent lexicon-related diagrams.} This finalizes the creation of the explainable lexicon containing words that are automatically extracted with the help of transformers and SHAP. A distinctive property of this lexicon is the usage of SHAP values, especially $SHAP_{avg}$, which will be later used to perform sentiment analysis. As will be shown by the results in Section \ref{sec:6}, $SHAP_{avg}$ is a good indicator of the sentiment of a particular word. In addition to this feature, $SHAP_{ratio}$ and $count$ are also introduced as parameters that will be used in sentiment analysis when determining the polarity of a particular sentence. This is explained in detail in Section \ref{sec:5}.

\begin{table*}[]
\caption{An excerpt of selected features of the explainable lexicon.}
\centering
\begin{tabular}{|c|c|c|c|c|c|c|c|c|}
\hline
Word &
  \begin{tabular}[c]{@{}c@{}}Count\\ (Selected)\end{tabular} &
  Total &
  \begin{tabular}[c]{@{}c@{}}Count\\ (Opposite)\end{tabular} &
  Category &
  \begin{tabular}[c]{@{}c@{}}Average\\ SHAP Value\\ (Selected)\end{tabular} &
  \begin{tabular}[c]{@{}c@{}}Average\\ SHAP Value\\ (Opposite)\end{tabular} &
  \begin{tabular}[c]{@{}c@{}}Sum\\ SHAP Value\\ (Selected)\end{tabular} &
  \begin{tabular}[c]{@{}c@{}}Sum\\ SHAP Value\\ (Opposite)\end{tabular} \\ \hline
new     & 426 & 577 & 151 & positive & 0.090629 & 0.021480 & 38.608165 & 3.243527 \\ \hline
amp     & 311 & 568 & 257 & negative & 0.032435 & 0.037868 & 10.087175 & 9.732118 \\ \hline
world   & 267 & 513 & 246 & positive & 0.046991 & 0.027468 & 12.546591 & 6.757057 \\ \hline
year    & 384 & 461 & 77  & negative & 0.050818 & 0.032912 & 19.514242 & 2.534230 \\ \hline
china   & 213 & 426 & 213 & positive & 0.042231 & 0.032894 & 8.995204  & 7.006360 \\ \hline
group   & 297 & 398 & 101 & positive & 0.079659 & 0.026396 & 23.658838 & 2.666006 \\ \hline
company & 291 & 387 & 96  & positive & 0.076470 & 0.030276 & 22.252625 & 2.906498 \\ \hline
energy  & 327 & 357 & 30  & positive & 0.109173 & 0.019223 & 35.699520 & 0.576700 \\ \hline
bank    & 202 & 351 & 149 & positive & 0.069276 & 0.036797 & 13.993703 & 5.482755 \\ \hline
power   & 312 & 328 & 16  & positive & 0.136059 & 0.019770 & 42.450354 & 0.316315 \\ \hline
state   & 258 & 301 & 43  & negative & 0.039640 & 0.011132 & 10.227108 & 0.478668 \\ \hline
million & 168 & 285 & 117 & negative & 0.048877 & 0.050262 & 8.211418  & 5.880613 \\ \hline
use     & 147 & 284 & 137 & positive & 0.042997 & 0.025001 & 6.320554  & 3.425115 \\ \hline
country & 240 & 280 & 40  & negative & 0.037619 & 0.023715 & 9.028585  & 0.948618 \\ \hline
market  & 177 & 273 & 96  & positive & 0.048187 & 0.043074 & 8.529055  & 4.135104 \\ \hline
trump   & 231 & 269 & 38  & negative & 0.086255 & 0.022430 & 19.924996 & 0.852340 \\ \hline
time    & 192 & 264 & 72  & negative & 0.045261 & 0.026059 & 8.690184  & 1.876225 \\ \hline
apple   & 224 & 255 & 31  & positive & 0.112486 & 0.022463 & 25.196845 & 0.696366 \\ \hline
chinese & 139 & 248 & 109 & negative & 0.035228 & 0.026792 & 4.896641  & 2.920374 \\ \hline
global  & 173 & 240 & 67  & positive & 0.078297 & 0.040759 & 13.545395 & 2.730843 \\ \hline
billion & 168 & 233 & 65  & negative & 0.055175 & 0.048924 & 9.269427  & 3.180079 \\ \hline
ban     & 149 & 228 & 79  & negative & 0.066532 & 0.032665 & 9.913209  & 2.580523 \\ \hline
day     & 180 & 223 & 43  & negative & 0.064980 & 0.026559 & 11.696429 & 1.142023 \\ \hline
\end{tabular}
\label{table:excerpt_explainable_lexicon}
\end{table*}

Our aim is to use the explainable lexicon to improve and extend the LM lexicon. To compare their performance, these two lexicons are combined into a final lexicon, which for the remainder of the paper will be interchangeably referred to as the combined lexicon or XLex+LM lexicon. We will use the combined lexicon to perform sentiment analysis on financial sentences, thereby evaluating the possible improvement of the combined lexicon over the plain vanilla LM lexicon. The results obtained by analyzing the combined lexicon will be shown and discussed in Section \ref{sec:6}. In the next and last phase, Phase 4, we explain the process of combining the explainable and LM lexicons into the combined lexicon.

\subsection{Phase 4: Merging with the Loughran-McDonald Dictionary}

In this last phase, we combine the explainable lexicon with the LM lexicon. However, before this can be done, it is necessary that the words in the LM lexicon undergo similar processing as in the case of the explainable lexicon so that the LM words obtain the same set of features. The processing of words of the LM lexicon is given in Figure \ref{fig:LM_processing} and is explained as follows.

\subsubsection{Processing of the LM Lexicon}

While the Loughran-McDonald lexicon consists of seven sentiment datasets, only its positive and negative components (datasets) are of interest to the combined (XLex+LM) lexicon. Similarly to the datasets used to create the explainable lexicon, the words from the LM datasets are first transformed into lowercase letters and then lemmatized. Duplicate words are obtained due to lemmatization. These datasets consist only of words without any other additional features (columns), so there is no need to aggregate the duplicates for a particular word. Instead, all duplicates are removed, leaving only one instance of the word in the datasets. To be able to combine this lexicon with the explainable lexicon, it is necessary to ensure they have the same features. Thus, all features from the explainable lexicon (shown in Table \ref{table:word_features_in_lexicons}) are added to the LM datasets.

As a first step, the initial features and the features introduced in Phase 2 are added to each of the LM datasets. These newly added features (except for those labeled as \textit{“opposite”}) are assigned a value of 1 as their main (default) value. Since these words do not contain values for the corresponding features, it is necessary to assign them a specific value. The value 1 is chosen as the main value to indicate if the word is present in the given dataset. While 1 is a high value to be assigned to $SHAP_{avg}$, this default value assignment is compensated with the model coefficients that are introduced in Section \ref{sec:5}. On the other hand, those features labeled as opposite are assigned a value of 0 since there are no words from one dataset that overlap with the other dataset. This assignment of values is a consequence of the fact that the words originating from the LM datasets are not obtained in an explainable way using SHAP; thus, they do not have the characteristics shown in Table \ref{table:word_features_in_lexicons}.

In addition, the feature ``Category'' is added to all the words from the LM datasets. For the words from the positive and negative LM dataset, this column is filled with the value ``positive'' or ``negative'' respectively. As was the case with the datasets from the explainable lexicon, the purpose of the ``Category'' feature in the LM datasets is to be able to identify the origin of a given word in the merged LM lexicon, i.e., whether the word originates from the positive or negative LM dataset. After this, the two LM datasets are merged into a single consolidated dataset by simply adding the data points (i.e., the words together with all their features) from the negative to the positive LM dataset. This concludes the processing of the LM lexicon. In the next subsection, the LM lexicon will be merged with the explainable lexicon to arrive at the combined (XLex+LM) lexicon. A visual overview of the LM lexicon after merging the positive and negative LM datasets, along with some of the added features, is shown in Table \ref{table:excerpt_lm_lexicon}.

\begin{table*}[]
\caption{An excerpt of selected features of the LM lexicon.}
\centering
\begin{tabular}{|c|c|c|c|c|c|c|c|c|}
\hline
Word &
  \begin{tabular}[c]{@{}c@{}}Count\\ (Selected)\end{tabular} &
  Total &
  \begin{tabular}[c]{@{}c@{}}Count\\ (Opposite)\end{tabular} &
  Category &
  \begin{tabular}[c]{@{}c@{}}Average\\ SHAP Value\\ (Selected)\end{tabular} &
  \begin{tabular}[c]{@{}c@{}}Average\\ SHAP Value\\ (Opposite)\end{tabular} &
  \begin{tabular}[c]{@{}c@{}}Sum\\ SHAP Value\\ (Selected)\end{tabular} &
  \begin{tabular}[c]{@{}c@{}}Sum\\ SHAP Value\\ (Opposite)\end{tabular} \\ \hline
surpasses    & 1 & 1 & 0 & positive & 1 & 0 & 1 & 0 \\ \hline
transparency & 1 & 1 & 0 & positive & 1 & 0 & 1 & 0 \\ \hline
tremendous   & 1 & 1 & 0 & positive & 1 & 0 & 1 & 0 \\ \hline
tremendously & 1 & 1 & 0 & positive & 1 & 0 & 1 & 0 \\ \hline
unmatched    & 1 & 1 & 0 & positive & 1 & 0 & 1 & 0 \\ \hline
unparalleled & 1 & 1 & 0 & positive & 1 & 0 & 1 & 0 \\ \hline
unsurpassed  & 1 & 1 & 0 & positive & 1 & 0 & 1 & 0 \\ \hline
upturn       & 1 & 1 & 0 & positive & 1 & 0 & 1 & 0 \\ \hline
valuable     & 1 & 1 & 0 & positive & 1 & 0 & 1 & 0 \\ \hline
versatile    & 1 & 1 & 0 & positive & 1 & 0 & 1 & 0 \\ \hline
versatility  & 1 & 1 & 0 & positive & 1 & 0 & 1 & 0 \\ \hline
vibrancy     & 1 & 1 & 0 & positive & 1 & 0 & 1 & 0 \\ \hline
vibrant      & 1 & 1 & 0 & positive & 1 & 0 & 1 & 0 \\ \hline
win          & 1 & 1 & 0 & positive & 1 & 0 & 1 & 0 \\ \hline
winner       & 1 & 1 & 0 & positive & 1 & 0 & 1 & 0 \\ \hline
worthy       & 1 & 1 & 0 & positive & 1 & 0 & 1 & 0 \\ \hline
abandon      & 1 & 1 & 0 & negative & 1 & 0 & 1 & 0 \\ \hline
abandonment  & 1 & 1 & 0 & negative & 1 & 0 & 1 & 0 \\ \hline
abdicate     & 1 & 1 & 0 & negative & 1 & 0 & 1 & 0 \\ \hline
abdicates    & 1 & 1 & 0 & negative & 1 & 0 & 1 & 0 \\ \hline
abdication   & 1 & 1 & 0 & negative & 1 & 0 & 1 & 0 \\ \hline
aberrant     & 1 & 1 & 0 & negative & 1 & 0 & 1 & 0 \\ \hline
aberration   & 1 & 1 & 0 & negative & 1 & 0 & 1 & 0 \\ \hline
\end{tabular}
\label{table:excerpt_lm_lexicon}
\end{table*}

\subsubsection{Obtaining the XLex+LM Lexicon by Merging the XLex and LM Lexicons}

As a final step, we merge the explainable lexicon (XLex) created in Phase 3 with the LM lexicon. We make two changes in the lexicons before merging them. We introduce a new feature (column) called \textit{src} (``source'') as shown in Table \ref{table:word_features_in_lexicons}. Since two different lexicons will be merged into one, the purpose of this feature is to indicate the origin of a certain word in the merged lexicon, i.e., whether the word originates from the explainable or LM lexicon. The feature is filled in with the value ``XLex'' and ``LM'' if the word originates from the explainable and LM lexicon, respectively. The $src$ feature allows flexibility in selecting the lexicon that is used by the sentiment analysis model in the evaluation process. Thus, it is possible to select the explainable lexicon (XLex), the LM lexicon, or the combined (XLex+LM) lexicon.

We also add a prefix to all features in the lexicons (i.e., all features indicated in Table \ref{table:word_features_in_lexicons}). The only exception is the column that contains the word itself (``word'' column) since that column is used to merge the two lexicons. Adding the prefix is done with the same purpose, namely to have the flexibility to select a lexicon for the sentiment analysis model. Selecting a certain lexicon means taking into account only its words and features in the sentiment analysis and not the words and features of the other lexicon. Before merging the lexicons, they have the same names for the features, so to distinguish these features in the combined lexicon, it is necessary to name them differently. We add prefixes ``XLex'' and ``LM'' to denote the columns from the explainable and LM lexicon, respectively. This is shown in Tables \ref{table:explainable_lexicon_with_prefix}-\ref{table:LM_lexicon_with_prefix} in the Appendix \ref{Appendix_A}. The prefix ``XLex'' stands for ``eXplainable Lexicon'' and indicates that the lexicon is created using explainability tools. The prefix ``LM'' is an abbreviation for the Loughran-McDonald lexicon, indicating that these features are related to the LM lexicon. With these two changes, it is possible to completely extract the explainable or LM lexicon from the combined lexicon.


After merging the lexicons, all words will appear with one instance in the combined dataset, including words that appear in both lexicons. The features of a given word in the combined lexicon will contain the feature values of both the explainable and LM lexicons for that word. This is shown in Table  \ref{table:combined_lexicon_words_in_both_explainable_and_LM}. If a particular word does not appear in both lexicons, it will also be represented by a single instance in the combined lexicon, but its instance will be populated only with the features of that word from the lexicon in which it exists, not the features from the other lexicon. This is shown in Table \ref{table:either_explainable_or_LM}. Words of this type do not appear in both lexicons and therefore, for the lexicon in which they do not appear, there is no value that can be assigned to them. This is the reason why after merging the lexicons, there are words with missing feature values for certain columns, as indicated with ``NaN'' (missing value) in Table \ref{table:either_explainable_or_LM}. For different columns, the missing values are handled differently. The feature ``XLex Category'' is filled with the value ``none'' because that word does not appear in the explainable lexicon. Similarly, the feature ``LM Category'' is filled in with the value ``none'' because the word is not present in the LM lexicon. If the column ``XLex Source'' has the value ``NaN'', then it means that the word is from the LM lexicon, so the column ``LM Source'' has the value ``LM''. To indicate that the word does not appear in the explainable lexicon, the ``NaN'' value of the ``XLex Source'' column is replaced by the ``LM'' value. Similarly, if the ``LM Source'' column has the value ``NaN'', then it means that the word is from the explainable lexicon, so the ``XLex Source'' column has the value ``XLex''. To indicate that the word is not contained in the LM lexicon, the value ``NaN'' of the column ``LM Source'' is replaced by the value ``XLex''. All other columns that contain ``NaN'' values (for the corresponding lexicon in which the given word does not appear) are assigned the value of 0. If a word does not appear in a given lexicon, the value for all its features is 0. Figure \ref{fig:merge_between_explainable_LM_lexicons} summarizes the handling of missing (``NaN'') values that arise due to the merging of the two lexicons.


After handling the invalid feature values, an excerpt of the lexicon’s content is shown in Table \ref{table:combined_lexicon_handled_invalid_values}. Comparing Table \ref{table:either_explainable_or_LM} and Table \ref{table:combined_lexicon_handled_invalid_values} can reveal the effect of replacing invalid values for certain columns. A normalized version of the combined lexicon is then created. To obtain the normalized lexicon, the values of each of the numerical features are modified according to Equation \ref{eq:4}:
\begin{equation}
    v_{norm} = \frac{v(f)}{max(f)}
    \label{eq:4}
\end{equation}
where $v(f)$ represents the value of a feature for a given word, while $max(f)$ is the maximum value of that feature across all words. This step concludes the creation of the combined XLex+LM lexicon, which can now be used as a basis for performing sentiment analysis.


\begin{table*}[]
\caption{The combined lexicon after handling invalid values.}
\centering
\resizebox{\textwidth}{!}{%
\begin{tabular}{|c|c|c|c|c|c|c|c|c|c|c|}
\hline
Word &
  \begin{tabular}[c]{@{}c@{}}XLex Count\\ (Selected)\end{tabular} &
  XLex Total &
  \begin{tabular}[c]{@{}c@{}}XLex Count\\ (Opposite)\end{tabular} &
  \begin{tabular}[c]{@{}c@{}}XLex Average\\ SHAP Value\\ (Selected)\end{tabular} &
  XLex Source &
  \begin{tabular}[c]{@{}c@{}}LM Average\\ SHAP Value\\ (Selected)\end{tabular} &
  LM Category &
  \begin{tabular}[c]{@{}c@{}}LM Sum\\ SHAP Value\\ (Opposite)\end{tabular} &
  LM Source &
  \begin{tabular}[c]{@{}c@{}}LM Max\\ SHAP Value\\ (Opposite)\end{tabular} \\ \hline
abide      & 1.0 & 1.0 & 0.0 & 0.003838 & XLex & 0 & none     & 0 & XLex & 0 \\ \hline
abo        & 1.0 & 1.0 & 0.0 & 0.000976 & XLex & 0 & none     & 0 & XLex & 0 \\ \hline
aboard     & 2.0 & 2.0 & 0.0 & 0.122640 & XLex & 0 & none     & 0 & XLex & 0 \\ \hline
abolition  & 1.0 & 1.0 & 0.0 & 0.006430 & XLex & 0 & none     & 0 & XLex & 0 \\ \hline
abroad     & 3.0 & 3.0 & 0.0 & 0.039073 & XLex & 0 & none     & 0 & XLex & 0 \\ \hline
writeoff   & 0.0 & 0.0 & 0.0 & 0.000000 & LM         & 1 & negative & 0 & LM         & 0 \\ \hline
writeoffs  & 0.0 & 0.0 & 0.0 & 0.000000 & LM         & 1 & negative & 0 & LM         & 0 \\ \hline
wrongful   & 0.0 & 0.0 & 0.0 & 0.000000 & LM         & 1 & negative & 0 & LM         & 0 \\ \hline
wrongfully & 0.0 & 0.0 & 0.0 & 0.000000 & LM         & 1 & negative & 0 & LM         & 0 \\ \hline
wrongly    & 0.0 & 0.0 & 0.0 & 0.000000 & LM         & 1 & negative & 0 & LM         & 0 \\ \hline
\end{tabular}%
}
\label{table:combined_lexicon_handled_invalid_values}
\end{table*}

In the next section, we define a model for sentiment analysis based on the combined lexicon.

\section{Model for Sentiment Analysis Based on Explainable Lexicons}
\label{sec:5}

In this section, we develop a model for sentiment analysis. The model is designed to make lexicon-based decisions, namely using the combined XLex+LM lexicon. To perform sentiment classification, the model can also use the explainable or LM lexicon as input since both can be extracted from the combined lexicon. To determine the sentiment of sentences, it is necessary to pass the following input parameters to the model: the combined lexicon, the lexicon’s features that will be used to make decisions about the sentiment of the sentences, as well as the source of the words, that is, which of the lexicons will be used in the analysis (explainable, LM or combined lexicon). There are three features used for decision-making purposes: $SHAP_{avg}$, $SHAP_{ratio}$, and $count$. Each of these characteristics can make the decision individually, but they can also be used together in any combination. The details of how these decision features are used together are explained later in this section.

After defining the model and its input parameters, we use the model to perform sentiment classification of financial sentences. The datasets used in the process of sentiment classification, and the corresponding results are outlined in Section \ref{sec:6}. We use the evaluation method of our model. The input parameters passed to the method are the sentences to be evaluated, the actual labels (sentiment) of those sentences, as well as 4 or 2 coefficients, depending on whether the combined lexicon or any of the constituent lexicons is used individually. The purpose of these coefficients is to control how much each of the lexicons will contribute to the decision, as well as how much importance will be given to the selected category relative to the opposite category.

We now explain how to calculate the sentiment of a certain sentence using the sentiment analysis model, relying on the combined lexicon. The explanation applies to one sentence, but the same process is applied to every sentence in the dataset. To determine the sentiment of a particular sentence, it is first split into its component words using a tokenizer. We employ two types of tokenizers in our methodology. For XLex, we use the corresponding RoBERTa/FinBERT tokenizers, and for LM, we use NLTK, which is a standard rule-based tokenizer. After applying the tokenizers, every word is transformed into lowercase letters and lemmatized. All words in the combined lexicon are lemmatized, so in order to follow an identical approach, we also lemmatize the words from the evaluation sentences. Each of the sentences is represented as a set of words $w_i$, $1 \leq i \leq n$:

\begin{equation}
    sentence = \{ w_1, w_2, ..., w_n \}
    \label{eq:5}
\end{equation}

We calculate the sentiment value of every word $w_i$ in a given sentence. Before calculating this sentiment value, it is necessary to calculate a cumulative value for each of the lexicons selected in the sentiment analysis (explainable and LM) and for each of the word categories (positive and negative).  The term ``cumulative value'' refers to the sum of the values of the word's features. For a specific word and for a specific lexicon, the cumulative value for the positive category is calculated as per Equation \ref{eq:6}:
\begin{equation}
    {v_c}^{pos}(w_i) = \sum_{i=1}^n v^{pos}(x_i)
    \label{eq:6}
\end{equation}
where $x_i$, $1 \leq i \leq n$, are the features selected to make the decision in the sentiment analysis. These features are the same for each of the selected lexicons and for each of the word categories. $v^{pos}(x_i)$ is the value of the feature $x_i$ of the positive word category. The sum of all these features represents the cumulative value of a given word in the selected lexicon as per the positive category. As previously mentioned, these decision-making features can be at most three ($SHAP_{avg}$, $SHAP_{ratio}$, and $count$) and at least one. At the same time, it is possible to use any other combination of them. Similarly to the positive word category, the cumulative value of a given word in the selected lexicon with respect to the negative category is calculated as per Equation \ref{eq:7}:
\begin{equation}
    {v_c}^{neg}(w_i) = (-1) \sum_{i=1}^n v^{neg}(x_i)
    \label{eq:7}
\end{equation}
where $x_i$, $1 \leq i \leq n$, are the selected features used to make sentiment decisions while $v^{neg}(x_i)$ is the value of the feature $x_i$ in the negative word category. The sum across all the selected features represents the cumulative value of a given word in the negative category for the selected lexicon. As can be seen in Equation \ref{eq:7}, the sum is multiplied by $-1$, ensuring that the cumulative value for the negative word category is always negative. Initially, all feature values are positive in each of the two lexicons as well as in the combined lexicon, i.e., given by their absolute values. Thus, it is necessary to multiply the cumulative value by $-1$ for the negative category. As will be pointed out later in this subsection, this facilitates the calculation of the sentiment value of the analyzed word. It should be noted that if a certain word does not exist in one of the categories or in one of the lexicons, the cumulative value evaluates to 0 according to Equations \ref{eq:6}-\ref{eq:7}.

The sentiment value of a given word can be calculated after determining the cumulative value for each lexicon and each category. If the combined XLex+LM lexicon is chosen for performing the sentiment analysis, the sentiment value of the word is obtained using Equation \ref{eq:8}:
\begin{equation}
    \begin{split}
        v_{sent}(w_i) & = c_{xlp}*{v_c}^{xl}(w_i) + c_{xlo}*{v_c}^{xl,opp}(w_i) + \\
        & + c_{lmp}*{v_c}^{lm}(w_i) + c_{lmo}*{v_c}^{lm,opp}(w_i)
    \end{split}
    \label{eq:8}
\end{equation}
The variables used in Equation \ref{eq:8} are summarized and explained in Table \ref{table:explanations_variables_equations}. The coefficients (parameters) in Equation \ref{eq:8} are introduced to control the contribution of each lexicon and each category on the sentiment classification decision. As will be explained in Section \ref{sec:6}, the parameters can be fine-tuned to investigate which values lead to improved sentiment classification performance.

If only the explainable lexicon is passed to the sentiment analysis model when determining the sentiment of the sentences, the sentiment value of a given word is calculated using Equation \ref{eq:9}:
\begin{equation}
    v_{sent}(w_i) = c_{xlp}*{v_c}^{xl}(w_i) + c_{xlo}*{v_c}^{xl,opp}(w_i)
    \label{eq:9}
\end{equation}
As can be seen, only two coefficients are used in this equation instead of four since only one of the lexicons is selected.

On the other hand, if only the LM lexicon is selected as an input to the sentiment analysis model, the sentiment value of a given word is calculated by Equation \ref{eq:10}:
\begin{equation}
    v_{sent}(w_i) = c_{lmp}*{v_c}^{lm}(w_i) + c_{lmo}*{v_c}^{lm,opp}(w_i)
    \label{eq:10}
\end{equation}
Equation \ref{eq:10} also has only two parameters instead of four since only one of the lexicons is selected.

\begin{table*}[]
\caption{Explanations of the variables used in the equations for calculating the sentiment value of a given word (Equations \ref{eq:8}-\ref{eq:10}).}
\centering
\begin{tabular}{|l|l|}
\hline
Variable                         & Description                                                                                                                                                \\ \hline
Primary category                 & \begin{tabular}[c]{@{}l@{}}The category selected as the primary category in\\ Phase 2 (positive or negative)\end{tabular}                                  \\ \hline
Opposite category                & \begin{tabular}[c]{@{}l@{}}The category that was not selected as the primary\\ category in Phase 2 (positive or negative)\end{tabular}                     \\ \hline
$c_{xlp}$                        & \begin{tabular}[c]{@{}l@{}}Coefficient of influence on the cumulative value\\ of the primary category in relation to the explainable lexicon\end{tabular}  \\ \hline
${{v_c}^{xl}}$           & \begin{tabular}[c]{@{}l@{}}Cumulative value of the primary category in relation\\ to the explainable lexicon\end{tabular}                        \\ \hline
$c_{xlo}$                        & \begin{tabular}[c]{@{}l@{}}Coefficient of influence on the cumulative value\\ of the opposite category in relation to the explainable lexicon\end{tabular} \\ \hline
${{v_c}^{xl, opp}}$  & \begin{tabular}[c]{@{}l@{}}Cumulative value of the opposite category in relation to the\\ explainable lexicon\end{tabular}                                 \\ \hline
$c_{lmp}$                        & \begin{tabular}[c]{@{}l@{}}Coefficient of influence on the cumulative value\\ of the primary category in relation to the LM lexicon\end{tabular}           \\ \hline
${{v_c}^{lm}}$            & \begin{tabular}[c]{@{}l@{}}Cumulative value of the primary category in relation\\ to the LM lexicon\end{tabular}                                           \\ \hline
$c_{lmo}$                        & \begin{tabular}[c]{@{}l@{}}Coefficient of influence on the cumulative value\\ of the opposite category in relation to the LM lexicon\end{tabular}          \\ \hline
${{v_c}^{lm, opp}}$  & \begin{tabular}[c]{@{}l@{}}Cumulative value of the opposite category\\ in relation to the LM lexicon\end{tabular}                                          \\ \hline
\end{tabular}
\label{table:explanations_variables_equations}
\end{table*}

After calculating the sentiment value of every word in a sentence using the above equations, the sentiment value of the sentence is evaluated as the sum of the sentiment value of each of the constituent words. This is given in Equation \ref{eq:11}:
\begin{equation}
    {v_{sent}}(sentence) = \sum_{i=1}^n {v_{sent}}(w_i)
    \label{eq:11}
\end{equation}
where $sentence$ is represented as a set of words (Equation \ref{eq:5}). In this way, the sentiment value of a certain sentence is calculated. Next, we determine the polarity of a sentence, i.e., whether it is positive, negative, or neutral. To calculate the sentiment polarity $s_{pol}$ of a sentence from its sentiment value, we check whether the sentiment value is positive, negative, or equal to 0 as follows:

\begin{equation}
    {s_{pol}}(sentence) = 
    \begin{cases}
        positive & : {v_{sent}}(sentence) > 0 \\
        negative & : {v_{sent}}(sentence) < 0 \\
        neutral  & : otherwise        
    \end{cases}
    \label{eq:12}
\end{equation}

\begin{figure}[ht]
    \centering
    \includegraphics[width=0.5\textwidth]{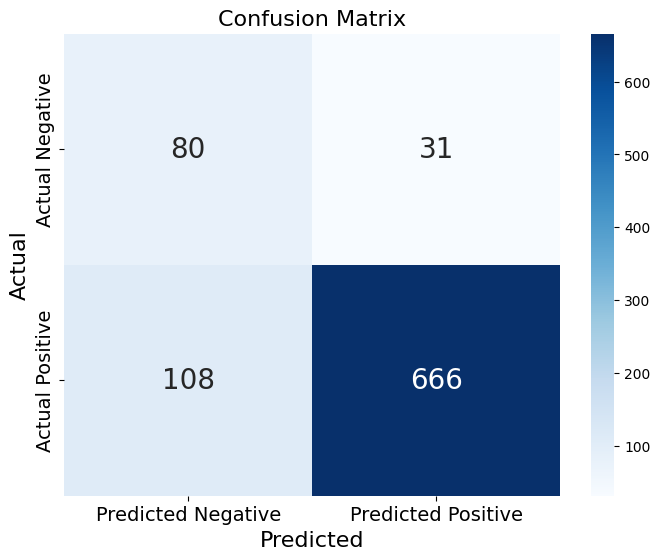}
    \caption{Confusion matrix showcasing the performance of the XLex-based model (using RoBERTa) based on the combined XLex+LM lexicon. The model is constructed using the lexicon created with the $nasdaq$ dataset as its source dataset. The model evaluation is performed on the $financial\_phrase\_bank$ dataset. This model achieves the highest accuracy among all XLex+LM models tested, achieving an accuracy rate of 84.3\%.}
    \label{fig:confusion_matrix}
\end{figure}

The sentiment model uses Equation \ref{eq:12} to calculate the sentiment of each sentence that is subject to sentiment analysis. After calculating the sentiment of each sentence, we evaluate the sentiment classification performance of the model. The evaluation is performed using the predicted and actual sentiments of each of the sentences. For this purpose, we use standard classification metrics such as accuracy, F1 score, and MCC. We also generate a classification report and confusion matrix. The results regarding the accuracy metric are presented in Table \ref{table:results_from_models}, while the F1 and MCC scores are given in Appendix \ref{Appendix_B}. The confusion matrix and classification report are presented in Figure \ref{fig:confusion_matrix} and Table \ref{table:classification_report}, respectively. The confusion matrix and classification report are generated for the XLex+LM model achieving the highest accuracy across the experiments performed. As shown later in the paper, the XLex+LM model achieves its highest accuracy of $84.3\%$ when constructed with the $nasdaq$ dataset as its source dataset and evaluated on the $financial\_phrase\_bank$ dataset.

Equations \ref{eq:11}-\ref{eq:12} show why it is necessary to involve multiplication by $-1$ in Equation \ref{eq:7}. The sentiment value of a certain sentence is the sum of the sentiment values of the constituent words of that sentence. The sentiment of the sentence depends on whether its sentiment value is positive or negative. Thus, it is important to ensure that a positive word leads to a positive sentiment value while a negative word leads to a negative sentiment value. This is achieved by the equations for calculating the cumulative value (Equations \ref{eq:6}-\ref{eq:7}).

The methodology explained in this section completes the entire process – from automatic word extraction, word classification, and postprocessing to creating an explainable lexicon with SHAP, combining it with the manually annotated LM lexicon, and finally creating a model that will classify the sentiment of finance-related sentences. The results obtained by applying this model to different datasets of financial sentences are shown in the next section.

\section{Results and Discussion}
\label{sec:6}

We present results obtained by the model introduced in the previous section using the combined XLex+LM lexicon.

\subsection{Used Datasets}

Tables \ref{table:descriptions_datasets}-\ref{table:datasets_statistics} present the datasets used to build the explainable lexicons as well as the datasets on which these lexicons are evaluated. Table \ref{table:descriptions_datasets} summarizes these datasets by giving their descriptions, while Table \ref{table:datasets_statistics} contains summary statistics about the datasets. Each of the datasets consists of financial sentences, where each sentence is labeled with its sentiment polarity. It should be noted that these datasets do not contain the sentences that were used to train the initial model given in Section \ref{sec:3} with the goal of avoiding bias in the experiments. In addition, the evaluation datasets do not include any sentences that are present in the source datasets.

\begin{table*}[]
\caption{Descriptions of the datasets that are used in the evaluation of XLex methodology.}
\centering
\begin{tabular}{|l|l|}
\hline
Dataset                     & Description                                                                                                                                                   \\ \hline
sentfin                     & Sentfin: Dataset of financial news with entity-sentiment annotations                                                                                                               \\ \hline
fiqa\_labeled\_df           & Fiqa: Aspect-based dataset of financial sentences                                                                                                             \\ \hline
sem\_eval                     & Financially relevant news headlines annotated for fine-grained sentiment                                                                                                        \\ \hline
financial\_phrase\_bank     & \begin{tabular}[c]{@{}l@{}}Financial PhraseBank: Manually annotated financial\\ sentences about companies listed on the OMX Helsinki stock index\end{tabular} \\ \hline
fpb\_fiqa                   & Financial PhraseBank + Fiqa                                                                                                                                   \\ \hline
nasdaq                      & Financial news about companies listed on the NASDAQ index                                                                                                     \\ \hline
fiqa\_fpb\_sentfin\_neutral & All neutral sentences from Fiqa, Financial PhraseBank and Sentfin                                                                                             \\ \hline
\end{tabular}
\label{table:descriptions_datasets}
\end{table*}

\begin{table*}[]
\caption{Statistics of the datasets used in the evaluation of XLex methodology. The number of positive, negative, and neutral sentences, as well as the purpose of the datasets (used as a source or for evaluation), are shown.}
\centering
\begin{tabular}{|c|c|c|c|c|c|}
\hline
Label                       & \begin{tabular}[c]{@{}c@{}}Total number\\ of sentences\end{tabular} & Positive & Negative & Neutral & Purpose                                                        \\ \hline
fiqa\_labeled\_df           & 201 & 139 & 62 & 0 & Evaluation                                                     \\ \hline
sem\_eval                     & 353 & 270 & 83 & 0 & Evaluation                                                     \\ \hline
fpb\_fiqa                   & 1542 & 1236 & 306 & 0 & Evaluation                                                    \\ \hline
financial\_phrase\_bank     & 885 & 774 & 111 & 0 & Evaluation \\ \hline
financial\_phrase\_bank     & 2960 & 89 & 0 & 2871 & Source \\ \hline
nasdaq                      & 9202 & 3067 & 5903 & 232 & Source                                                         \\ \hline
fiqa\_fpb\_sentfin\_neutral & 6086 & 0 & 0 & 6086     & Source                                                         \\ \hline
\end{tabular}
\label{table:datasets_statistics}
\end{table*}

The label ``Source'' in the ``Purpose'' column in Table \ref{table:datasets_statistics} denotes that the corresponding dataset is used to extract words with SHAP and to generate an explainable lexicon. The label ``Evaluation'' in the ``Purpose'' column in Table \ref{table:datasets_statistics} denotes that the corresponding dataset is used to evaluate the generated explainable lexicons. Details about generating the explainable lexicons from these datasets are shown as follows.

The datasets utilized in this study were primarily obtained from Kaggle, with the exception of the SemEval-2017-Task5 dataset, which was accessed from the official page of the SemEval competition. For extracting words with SHAP, we conducted a thorough search on Kaggle to find suitable financial-related datasets comprising textual statements, sentences, or news headlines for sentiment analysis. Our selection criteria included datasets studied in the literature or containing relevant data about companies listed on the stock market, ensuring diverse sources for extracting positive and negative words for building the explainable lexicon. As for the evaluation datasets, we not only considered financial textual data but also ensured that they were appropriately annotated by financial experts. This selection process was implemented to ensure the validity and robustness of these datasets for evaluation purposes.

\subsection{Generated Lexicons}

We generate three different explainable lexicons. The words in the lexicons are generated from three different sources. The datasets serving as the sources of the lexicons are marked as ``Source'' in the ``Purpose'' column of Table \ref{table:datasets_statistics}. Each of the lexicons is created using the method described in Sections \ref{sec:3}-\ref{sec:4}. The purpose of using different sources is to verify the ability of the method presented in this paper to successfully generate explainable lexicons under different conditions (given that different sources exhibit varied data).

\begin{table*}[]
\caption{Statistics of the lexicons on which sentiment analysis is performed. The lexicons are obtained using the RoBERTa transformer model.}
\centering
\begin{tabular}{|c|c|c|c|}
\hline
Lexicon                     & \begin{tabular}[c]{@{}c@{}}Total number\\ of words\end{tabular} & Positive & Negative \\ \hline
fiqa\_fpb\_sentfin\_neutral & 3313 & 1635 & 1678     \\ \hline
nasdaq                      & 5751 & 2537 & 3214     \\ \hline
financial\_phrase\_bank     & 2729 & 1342 & 1387     \\ \hline
Loughran–McDonald           & 1731                                                            & 246      & 1485     \\ \hline
\end{tabular}
\label{table:statistics_lexicons_sentiment_analysis}
\end{table*}

In Table \ref{table:datasets_statistics}, the Sentfin dataset is denoted as a ``Source'' dataset because it is used only for the purposes of word extraction. The Sentfin dataset comprises headlines and corresponding sentiment labels for the financial entities mentioned in the headlines. Each financial entity in a headline is assigned a sentiment label. However, the dataset lacks sentiment labels specifically for the headlines themselves, which renders it unsuitable for evaluation purposes. Nonetheless, we utilize this dataset as a ``Source'' since the sentiment labels for the headlines are not required for our word extraction and classification process.

We also want to evaluate the effectiveness of the methodology to label the words with the appropriate sentiments automatically. Summary data of the explainable lexicons is shown in Table \ref{table:statistics_lexicons_sentiment_analysis}, which also gives information about the LM lexicon. Each of the explainable lexicons from Table \ref{table:statistics_lexicons_sentiment_analysis} is combined with the LM lexicon, and the resulting lexicons are used in the process of evaluating the model performance. The results of the analysis are shown in the next subsection.

\subsection{Results from the Sentiment Analysis}

The model takes two parameters that can be fine-tuned: decision coefficients and features that will be used to make sentiment decisions. We perform a grid search to find the optimal values of the model parameters that maximize the accuracy, F1 and MCC.

Although it is possible to use all three decision features ($SHAP_{avg}$, $SHAP_{ratio}$, and $count$), we conducted a grid search to identify the most effective combination. Our results revealed that $SHAP_{avg}$ has the dominant impact on accuracy, F1, and MCC, and we, therefore, selected it as the primary decision feature. Then we performed a second grid search by using both the standard (without normalization) and normalized versions of the explainable lexicons in order to find the optimal values of the coefficients $c_{xlp}$, $c_{xlo}$, $c_{lmp}$ and $c_{lmo}$. We chose $0.1$, $0.3$, $0.5$, $0.7$, and $0.9$ as possible values for these coefficients to distinguish between different levels of impact ($0.1$ denotes weak impact, while $0.9$ denotes strong impact). By applying permutations with repetition, all permutations of the values for the coefficients are obtained. There are five possible values that can be assigned to three coefficients (we exclude the $c_{lmo}$ coefficient as obsolete because there is no word shared between positive and negative words in the LM dictionary; thus, there are no ``opposite'' words in the LM dictionary). Thus, the total number of permutations is $5^3 = 125$. These permutations are combined with each of the three explainable lexicons (using both standard and normalized versions of the lexicons), and each of the four evaluation datasets. The only exceptions are the financial phrase bank explainable lexicon and the financial phrase bank evaluation dataset. We do not evaluate this combination to avoid biased results. As a result, we arrive at a total of 2750 models that are created for the purpose of grid search (125 permutations $\times$ 2 explainable lexicons $\times$ 2 lexicon versions $\times$ 4 evaluation datasets + 125 permutations $\times$ 1 explainable lexicon $\times$ 2 lexicon versions $\times$ 3 evaluation datasets). For each of the models in the grid search set of models, we are using the $SHAP_{avg}$ feature as the decision maker.

After obtaining the results, our primary goal is to identify the combination of coefficients that yields the highest aggregated average across the accuracy, F1, and MCC metrics. To calculate this aggregated average, we first determine the average values for accuracy, F1, and MCC scores across all experiments. We compute these average values for each combination of coefficients, considering both the explainable lexicons XLex and XLex+LM obtained from the three distinct source datasets ($nasdaq$, $fpb$, and $sentfin$). This results in a total of $3 \times 2 \times 3 = 18$ average values for each coefficient combination. The aggregated average for a specific combination of coefficients is obtained by summing these 18 average values, allowing us to represent each coefficient combination using a single consolidated parameter.

Through the grid search procedure, we discovered that the coefficients $(c_{xlp}, c_{xlo}, c_{lmp}, c_{lmo}) = (0.3, 0.1, 0.1, 0.5)$ form the combination that achieves the highest aggregated average, thus defining the optimal model parameters. It is worth noting that the choice of the value for the $c_{lmo}$ coefficient does not impact the grid search procedure, as there are no ``opposite'' words in the LM dictionary. Hence, we can safely assume that $c_{lmo} = 0.5$ without affecting the outcome of the grid search.

\begin{table*}[]
    \caption{Accuracy obtained using the XLex methodology based on the RoBERTa and FinBERT transformer models. The columns ``LM'', ``XLex'', and ``XLex+LM'' show the accuracy of the models on the corresponding dataset. The columns ``LM on LM'', ``XLex on LM'', ``(XLex+LM) on LM'' show the accuracy on portions of the datasets that have a recall from LM (i.e., all instances where the LM either provided no answer or were unable to make a decision are removed from the datasets). The approach with the highest accuracy among the three approaches is represented in bold.}
    \centering
    \begin{subtable}[c]{\textwidth}
    \subcaption{RoBERTa-based results}
    \centering
    \small
    \setlength\tabcolsep{2pt}
    \begin{tabular}{|c|c|c|ccc|ccc|}
    \hline
    \multicolumn{1}{|c|}{Source of the lexicon} &
      \multicolumn{1}{|c|}{Normalized} &
      \multicolumn{1}{|c|}{Evaluation set} &
      \multicolumn{3}{c|}{Accuracy on whole dataset} & \multicolumn{3}{c|}{\begin{tabular}[c]{@{}c@{}}Accuracy only on the part of the dataset for\\which the LM-based model has an answer
      \end{tabular}} \\ 
      \cline{4-9} 
     &
       &
       &
      \multicolumn{1}{c|}{LM} &
      \multicolumn{1}{c|}{XLex} &
      \multicolumn{1}{c|}{\begin{tabular}[c]{@{}c@{}}XLex+\\ LM\end{tabular}} &
      \multicolumn{1}{c|}{\begin{tabular}[c]{@{}c@{}}LM on\\ LM\end{tabular}} &
      \multicolumn{1}{c|}{\begin{tabular}[c]{@{}c@{}}XLex on\\ LM\end{tabular}} &
      \begin{tabular}[c]{@{}c@{}}(XLex+LM)\\ on LM\end{tabular} \\ \hline
    nasdaq &
Yes &
financial\_phrase\_bank &
\multicolumn{1}{c|}{0.303} &
\multicolumn{1}{c|}{0.836} &
\multicolumn{1}{c|}{{\textbf{0.843}}} &
\multicolumn{1}{c|}{0.753} &
\multicolumn{1}{c|}{0.784} &
\multicolumn{1}{c|}{{\textbf{0.801}}} \\ \hline
nasdaq &
Yes &
fiqa\_labeled\_df &
\multicolumn{1}{c|}{0.313} &
\multicolumn{1}{c|}{0.721} &
\multicolumn{1}{c|}{{\textbf{0.731}}} &
\multicolumn{1}{c|}{0.808} &
\multicolumn{1}{c|}{0.795} &
\multicolumn{1}{c|}{{\textbf{0.821}}} \\ \hline
nasdaq &
Yes &
fpb\_fiqa &
\multicolumn{1}{c|}{0.296} &
\multicolumn{1}{c|}{0.761} &
\multicolumn{1}{c|}{{\textbf{0.776}}} &
\multicolumn{1}{c|}{0.748} &
\multicolumn{1}{c|}{0.726} &
\multicolumn{1}{c|}{{\textbf{0.766}}} \\ \hline
nasdaq &
Yes &
sem\_eval &
\multicolumn{1}{c|}{0.275} &
\multicolumn{1}{c|}{0.745} &
\multicolumn{1}{c|}{{\textbf{0.765}}} &
\multicolumn{1}{c|}{0.752} &
\multicolumn{1}{c|}{0.721} &
\multicolumn{1}{c|}{{\textbf{0.775}}} \\ \hline
nasdaq &
No &
financial\_phrase\_bank &
\multicolumn{1}{c|}{0.303} &
\multicolumn{1}{c|}{0.837} &
\multicolumn{1}{c|}{{\textbf{0.843}}} &
\multicolumn{1}{c|}{0.753} &
\multicolumn{1}{c|}{0.781} &
\multicolumn{1}{c|}{{\textbf{0.795}}} \\ \hline
nasdaq &
No &
fiqa\_labeled\_df &
\multicolumn{1}{c|}{0.313} &
\multicolumn{1}{c|}{0.697} &
\multicolumn{1}{c|}{{\textbf{0.706}}} &
\multicolumn{1}{c|}{0.808} &
\multicolumn{1}{c|}{0.795} &
\multicolumn{1}{c|}{{\textbf{0.821}}} \\ \hline
nasdaq &
No &
fpb\_fiqa &
\multicolumn{1}{c|}{0.296} &
\multicolumn{1}{c|}{0.75} &
\multicolumn{1}{c|}{{\textbf{0.768}}} &
\multicolumn{1}{c|}{0.748} &
\multicolumn{1}{c|}{0.72} &
\multicolumn{1}{c|}{{\textbf{0.764}}} \\ \hline
nasdaq &
No &
sem\_eval &
\multicolumn{1}{c|}{0.275} &
\multicolumn{1}{c|}{0.745} &
\multicolumn{1}{c|}{{\textbf{0.756}}} &
\multicolumn{1}{c|}{0.752} &
\multicolumn{1}{c|}{0.729} &
\multicolumn{1}{c|}{{\textbf{0.76}}} \\ \hline
fiqa\_fpb\_sentfin\_neutral &
Yes &
financial\_phrase\_bank &
\multicolumn{1}{c|}{0.303} &
\multicolumn{1}{c|}{{\textbf{0.808}}} &
\multicolumn{1}{c|}{{\textbf{0.808}}} &
\multicolumn{1}{c|}{0.753} &
\multicolumn{1}{c|}{{\textbf{0.801}}} &
\multicolumn{1}{c|}{{\textbf{0.801}}} \\ \hline
fiqa\_fpb\_sentfin\_neutral &
Yes &
fiqa\_labeled\_df &
\multicolumn{1}{c|}{0.313} &
\multicolumn{1}{c|}{0.697} &
\multicolumn{1}{c|}{{\textbf{0.711}}} &
\multicolumn{1}{c|}{0.808} &
\multicolumn{1}{c|}{0.808} &
\multicolumn{1}{c|}{{\textbf{0.846}}} \\ \hline
fiqa\_fpb\_sentfin\_neutral &
Yes &
fpb\_fiqa &
\multicolumn{1}{c|}{0.296} &
\multicolumn{1}{c|}{0.733} &
\multicolumn{1}{c|}{{\textbf{0.746}}} &
\multicolumn{1}{c|}{0.748} &
\multicolumn{1}{c|}{0.748} &
\multicolumn{1}{c|}{{\textbf{0.779}}} \\ \hline
fiqa\_fpb\_sentfin\_neutral &
Yes &
sem\_eval &
\multicolumn{1}{c|}{0.275} &
\multicolumn{1}{c|}{0.725} &
\multicolumn{1}{c|}{{\textbf{0.756}}} &
\multicolumn{1}{c|}{0.752} &
\multicolumn{1}{c|}{0.69} &
\multicolumn{1}{c|}{{\textbf{0.775}}} \\ \hline
fiqa\_fpb\_sentfin\_neutral &
No &
financial\_phrase\_bank &
\multicolumn{1}{c|}{0.303} &
\multicolumn{1}{c|}{0.81} &
\multicolumn{1}{c|}{{\textbf{0.811}}} &
\multicolumn{1}{c|}{0.753} &
\multicolumn{1}{c|}{0.798} &
\multicolumn{1}{c|}{{\textbf{0.801}}} \\ \hline
fiqa\_fpb\_sentfin\_neutral &
No &
fiqa\_labeled\_df &
\multicolumn{1}{c|}{0.313} &
\multicolumn{1}{c|}{0.672} &
\multicolumn{1}{c|}{{\textbf{0.692}}} &
\multicolumn{1}{c|}{0.808} &
\multicolumn{1}{c|}{0.795} &
\multicolumn{1}{c|}{{\textbf{0.846}}} \\ \hline
fiqa\_fpb\_sentfin\_neutral &
No &
fpb\_fiqa &
\multicolumn{1}{c|}{0.296} &
\multicolumn{1}{c|}{0.733} &
\multicolumn{1}{c|}{{\textbf{0.746}}} &
\multicolumn{1}{c|}{0.748} &
\multicolumn{1}{c|}{0.744} &
\multicolumn{1}{c|}{{\textbf{0.777}}} \\ \hline
fiqa\_fpb\_sentfin\_neutral &
No &
sem\_eval &
\multicolumn{1}{c|}{0.275} &
\multicolumn{1}{c|}{0.722} &
\multicolumn{1}{c|}{{\textbf{0.759}}} &
\multicolumn{1}{c|}{0.752} &
\multicolumn{1}{c|}{0.674} &
\multicolumn{1}{c|}{{\textbf{0.775}}} \\ \hline
financial\_phrase\_bank &
Yes &
fiqa\_labeled\_df &
\multicolumn{1}{c|}{0.313} &
\multicolumn{1}{c|}{0.632} &
\multicolumn{1}{c|}{{\textbf{0.682}}} &
\multicolumn{1}{c|}{0.808} &
\multicolumn{1}{c|}{0.731} &
\multicolumn{1}{c|}{{\textbf{0.859}}} \\ \hline
financial\_phrase\_bank &
Yes &
fpb\_fiqa &
\multicolumn{1}{c|}{0.296} &
\multicolumn{1}{c|}{0.676} &
\multicolumn{1}{c|}{{\textbf{0.707}}} &
\multicolumn{1}{c|}{0.748} &
\multicolumn{1}{c|}{0.687} &
\multicolumn{1}{c|}{{\textbf{0.764}}} \\ \hline
financial\_phrase\_bank &
Yes &
sem\_eval &
\multicolumn{1}{c|}{0.275} &
\multicolumn{1}{c|}{0.671} &
\multicolumn{1}{c|}{{\textbf{0.697}}} &
\multicolumn{1}{c|}{0.752} &
\multicolumn{1}{c|}{0.698} &
\multicolumn{1}{c|}{{\textbf{0.767}}} \\ \hline
financial\_phrase\_bank &
No &
fiqa\_labeled\_df &
\multicolumn{1}{c|}{0.313} &
\multicolumn{1}{c|}{0.662} &
\multicolumn{1}{c|}{{\textbf{0.692}}} &
\multicolumn{1}{c|}{0.808} &
\multicolumn{1}{c|}{0.744} &
\multicolumn{1}{c|}{{\textbf{0.821}}} \\ \hline
financial\_phrase\_bank &
No &
fpb\_fiqa &
\multicolumn{1}{c|}{0.296} &
\multicolumn{1}{c|}{0.691} &
\multicolumn{1}{c|}{{\textbf{0.716}}} &
\multicolumn{1}{c|}{0.748} &
\multicolumn{1}{c|}{0.7} &
\multicolumn{1}{c|}{{\textbf{0.762}}} \\ \hline
financial\_phrase\_bank &
No &
sem\_eval &
\multicolumn{1}{c|}{0.275} &
\multicolumn{1}{c|}{0.671} &
\multicolumn{1}{c|}{{\textbf{0.694}}} &
\multicolumn{1}{c|}{0.752} &
\multicolumn{1}{c|}{0.705} &
\multicolumn{1}{c|}{{\textbf{0.767}}} \\ \hline\hline
\multicolumn{3}{|c|}{Average accuracy} &
\multicolumn{1}{c|}{0.296} &
\multicolumn{1}{c|}{0.727} &
\multicolumn{1}{c|}{{\textbf{0.746}}} &
\multicolumn{1}{c|}{0.766} &
\multicolumn{1}{c|}{0.744} &
\multicolumn{1}{c|}{{\textbf{0.793}}} \\ \hline
    \end{tabular}
    \label{table:results_from_models_roberta}
    \end{subtable}
    \newline
    \vspace*{2 mm}
    \newline
    \begin{subtable}[c]{\textwidth}
    \subcaption{FinBERT-based results}
    \centering
    \small
    \setlength\tabcolsep{2pt}
    \begin{tabular}{|c|c|c|ccc|ccc|}
    \hline
    \multicolumn{1}{|c|}{Source of the lexicon} &
      \multicolumn{1}{|c|}{Normalized} &
      \multicolumn{1}{|c|}{Evaluation set} &
      \multicolumn{3}{c|}{Accuracy on whole dataset} & \multicolumn{3}{c|}{\begin{tabular}[c]{@{}c@{}}Accuracy only on the part of the dataset for\\which the LM-based model has an answer
      \end{tabular}} \\ 
      \cline{4-9} 
     &
       &
       &
      \multicolumn{1}{c|}{LM} &
      \multicolumn{1}{c|}{XLex} &
      \multicolumn{1}{c|}{\begin{tabular}[c]{@{}c@{}}XLex+\\ LM\end{tabular}} &
      \multicolumn{1}{c|}{\begin{tabular}[c]{@{}c@{}}LM on\\ LM\end{tabular}} &
      \multicolumn{1}{c|}{\begin{tabular}[c]{@{}c@{}}XLex on\\ LM\end{tabular}} &
      \begin{tabular}[c]{@{}c@{}}(XLex+LM)\\ on LM\end{tabular} \\ \hline
    nasdaq &
Yes &
financial\_phrase\_bank &
\multicolumn{1}{c|}{0.303} &
\multicolumn{1}{c|}{0.768} &
\multicolumn{1}{c|}{{\textbf{0.779}}} &
\multicolumn{1}{c|}{0.753} &
\multicolumn{1}{c|}{0.736} &
\multicolumn{1}{c|}{{\textbf{0.761}}} \\ \hline
nasdaq &
Yes &
fiqa\_labeled\_df &
\multicolumn{1}{c|}{0.313} &
\multicolumn{1}{c|}{0.761} &
\multicolumn{1}{c|}{{\textbf{0.771}}} &
\multicolumn{1}{c|}{0.808} &
\multicolumn{1}{c|}{0.833} &
\multicolumn{1}{c|}{{\textbf{0.859}}} \\ \hline
nasdaq &
Yes &
fpb\_fiqa &
\multicolumn{1}{c|}{0.296} &
\multicolumn{1}{c|}{0.747} &
\multicolumn{1}{c|}{{\textbf{0.752}}} &
\multicolumn{1}{c|}{0.748} &
\multicolumn{1}{c|}{0.754} &
\multicolumn{1}{c|}{{\textbf{0.767}}} \\ \hline
nasdaq &
Yes &
sem\_eval &
\multicolumn{1}{c|}{0.275} &
\multicolumn{1}{c|}{0.725} &
\multicolumn{1}{c|}{{\textbf{0.728}}} &
\multicolumn{1}{c|}{0.752} &
\multicolumn{1}{c|}{0.775} &
\multicolumn{1}{c|}{{\textbf{0.783}}} \\ \hline
nasdaq &
No &
financial\_phrase\_bank &
\multicolumn{1}{c|}{0.303} &
\multicolumn{1}{c|}{0.694} &
\multicolumn{1}{c|}{{\textbf{0.722}}} &
\multicolumn{1}{c|}{0.753} &
\multicolumn{1}{c|}{0.694} &
\multicolumn{1}{c|}{{\textbf{0.764}}} \\ \hline
nasdaq &
No &
fiqa\_labeled\_df &
\multicolumn{1}{c|}{0.313} &
\multicolumn{1}{c|}{{\textbf{0.761}}} &
\multicolumn{1}{c|}{{\textbf{0.761}}} &
\multicolumn{1}{c|}{0.808} &
\multicolumn{1}{c|}{{\textbf{0.846}}} &
\multicolumn{1}{c|}{{\textbf{0.846}}} \\ \hline
nasdaq &
No &
fpb\_fiqa &
\multicolumn{1}{c|}{0.296} &
\multicolumn{1}{c|}{0.709} &
\multicolumn{1}{c|}{{\textbf{0.724}}} &
\multicolumn{1}{c|}{0.748} &
\multicolumn{1}{c|}{0.73} &
\multicolumn{1}{c|}{{\textbf{0.767}}} \\ \hline
nasdaq &
No &
sem\_eval &
\multicolumn{1}{c|}{0.275} &
\multicolumn{1}{c|}{0.714} &
\multicolumn{1}{c|}{{\textbf{0.722}}} &
\multicolumn{1}{c|}{0.752} &
\multicolumn{1}{c|}{0.767} &
\multicolumn{1}{c|}{{\textbf{0.791}}} \\ \hline
fiqa\_fpb\_sentfin\_neutral &
Yes &
financial\_phrase\_bank &
\multicolumn{1}{c|}{0.303} &
\multicolumn{1}{c|}{0.831} &
\multicolumn{1}{c|}{{\textbf{0.836}}} &
\multicolumn{1}{c|}{0.753} &
\multicolumn{1}{c|}{0.756} &
\multicolumn{1}{c|}{{\textbf{0.77}}} \\ \hline
fiqa\_fpb\_sentfin\_neutral &
Yes &
fiqa\_labeled\_df &
\multicolumn{1}{c|}{0.313} &
\multicolumn{1}{c|}{0.791} &
\multicolumn{1}{c|}{{\textbf{0.811}}} &
\multicolumn{1}{c|}{0.808} &
\multicolumn{1}{c|}{0.795} &
\multicolumn{1}{c|}{{\textbf{0.846}}} \\ \hline
fiqa\_fpb\_sentfin\_neutral &
Yes &
fpb\_fiqa &
\multicolumn{1}{c|}{0.296} &
\multicolumn{1}{c|}{0.78} &
\multicolumn{1}{c|}{{\textbf{0.796}}} &
\multicolumn{1}{c|}{0.748} &
\multicolumn{1}{c|}{0.736} &
\multicolumn{1}{c|}{{\textbf{0.777}}} \\ \hline
fiqa\_fpb\_sentfin\_neutral &
Yes &
sem\_eval &
\multicolumn{1}{c|}{0.275} &
\multicolumn{1}{c|}{0.756} &
\multicolumn{1}{c|}{{\textbf{0.785}}} &
\multicolumn{1}{c|}{0.752} &
\multicolumn{1}{c|}{0.721} &
\multicolumn{1}{c|}{{\textbf{0.798}}} \\ \hline
fiqa\_fpb\_sentfin\_neutral &
No &
financial\_phrase\_bank &
\multicolumn{1}{c|}{0.303} &
\multicolumn{1}{c|}{0.816} &
\multicolumn{1}{c|}{{\textbf{0.823}}} &
\multicolumn{1}{c|}{0.753} &
\multicolumn{1}{c|}{0.753} &
\multicolumn{1}{c|}{{\textbf{0.77}}} \\ \hline
fiqa\_fpb\_sentfin\_neutral &
No &
fiqa\_labeled\_df &
\multicolumn{1}{c|}{0.313} &
\multicolumn{1}{c|}{0.786} &
\multicolumn{1}{c|}{{\textbf{0.801}}} &
\multicolumn{1}{c|}{0.808} &
\multicolumn{1}{c|}{0.795} &
\multicolumn{1}{c|}{{\textbf{0.833}}} \\ \hline
fiqa\_fpb\_sentfin\_neutral &
No &
fpb\_fiqa &
\multicolumn{1}{c|}{0.296} &
\multicolumn{1}{c|}{0.774} &
\multicolumn{1}{c|}{{\textbf{0.791}}} &
\multicolumn{1}{c|}{0.748} &
\multicolumn{1}{c|}{0.736} &
\multicolumn{1}{c|}{{\textbf{0.779}}} \\ \hline
fiqa\_fpb\_sentfin\_neutral &
No &
sem\_eval &
\multicolumn{1}{c|}{0.275} &
\multicolumn{1}{c|}{0.756} &
\multicolumn{1}{c|}{{\textbf{0.785}}} &
\multicolumn{1}{c|}{0.752} &
\multicolumn{1}{c|}{0.729} &
\multicolumn{1}{c|}{{\textbf{0.806}}} \\ \hline
financial\_phrase\_bank &
Yes &
fiqa\_labeled\_df &
\multicolumn{1}{c|}{0.313} &
\multicolumn{1}{c|}{0.751} &
\multicolumn{1}{c|}{{\textbf{0.771}}} &
\multicolumn{1}{c|}{0.808} &
\multicolumn{1}{c|}{0.782} &
\multicolumn{1}{c|}{{\textbf{0.833}}} \\ \hline
financial\_phrase\_bank &
Yes &
fpb\_fiqa &
\multicolumn{1}{c|}{0.296} &
\multicolumn{1}{c|}{0.792} &
\multicolumn{1}{c|}{{\textbf{0.799}}} &
\multicolumn{1}{c|}{0.748} &
\multicolumn{1}{c|}{0.749} &
\multicolumn{1}{c|}{{\textbf{0.766}}} \\ \hline
financial\_phrase\_bank &
Yes &
sem\_eval &
\multicolumn{1}{c|}{0.275} &
\multicolumn{1}{c|}{0.734} &
\multicolumn{1}{c|}{{\textbf{0.776}}} &
\multicolumn{1}{c|}{0.752} &
\multicolumn{1}{c|}{0.698} &
\multicolumn{1}{c|}{{\textbf{0.814}}} \\ \hline
financial\_phrase\_bank &
No &
fiqa\_labeled\_df &
\multicolumn{1}{c|}{0.313} &
\multicolumn{1}{c|}{0.741} &
\multicolumn{1}{c|}{{\textbf{0.751}}} &
\multicolumn{1}{c|}{0.808} &
\multicolumn{1}{c|}{0.795} &
\multicolumn{1}{c|}{{\textbf{0.821}}} \\ \hline
financial\_phrase\_bank &
No &
fpb\_fiqa &
\multicolumn{1}{c|}{0.296} &
\multicolumn{1}{c|}{0.778} &
\multicolumn{1}{c|}{{\textbf{0.788}}} &
\multicolumn{1}{c|}{0.748} &
\multicolumn{1}{c|}{0.743} &
\multicolumn{1}{c|}{{\textbf{0.767}}} \\ \hline
financial\_phrase\_bank &
No &
sem\_eval &
\multicolumn{1}{c|}{0.275} &
\multicolumn{1}{c|}{0.739} &
\multicolumn{1}{c|}{{\textbf{0.776}}} &
\multicolumn{1}{c|}{0.752} &
\multicolumn{1}{c|}{0.705} &
\multicolumn{1}{c|}{{\textbf{0.806}}} \\ \hline\hline
\multicolumn{3}{|c|}{Average accuracy} &
\multicolumn{1}{c|}{0.296} &
\multicolumn{1}{c|}{0.759} &
\multicolumn{1}{c|}{{\textbf{0.775}}} &
\multicolumn{1}{c|}{0.766} &
\multicolumn{1}{c|}{0.756} &
\multicolumn{1}{c|}{{\textbf{0.797}}} \\ \hline
    \end{tabular}
    \label{table:results_from_models_finbert}
    \end{subtable}
\label{table:results_from_models}
\end{table*}

Once the optimal model parameters are selected, we proceed by generating the results. For this purpose, we use the combined lexicons from Table \ref{table:statistics_lexicons_sentiment_analysis} that are also available in their normalized form. Each of these lexicons serves as a basis for performing sentiment analysis using the model proposed in Section \ref{sec:5}. Each of the created models is evaluated on all evaluation datasets in Table \ref{table:datasets_statistics} (these are the datasets containing the label ``Evaluation'' in the ``Purpose'' column). Only the model that uses the Financial PhraseBank dataset as a source for the combined lexicon is not evaluated on that same dataset in order to avoid model bias.
The results about accuracy are shown in Table \ref{table:results_from_models_roberta}. Additional classification metrics, such as the F1 and MCC scores, are given in Tables \ref{F1_Results}-\ref{MCC_Results} in Appendix \ref{Appendix_B}.

\subsection{Discussion}

Table \ref{table:results_from_models_roberta} reveals that the model achieves overall best accuracy results in sentiment analysis when the combined lexicon XLex+LM is used as a basis for the analysis. The same applies when the explainable lexicon XLex is used as a basis. The highest accuracy of the model based on the combined XLex+LM lexicon is 0.843 (column 6 in Table \ref{table:results_from_models_roberta}); if sentiment classification is performed using only the explainable lexicon under the same conditions (i.e., the same source of the lexicon and the same evaluation dataset), the obtained accuracy evaluates to 0.837 (column 5 in Table \ref{table:results_from_models_roberta}). For the same experiment, the accuracy evaluates to 0.303 (column 4 in Table \ref{table:results_from_models_roberta}) if only the LM lexicon is taken as a basis for sentiment classification. The reason for this result is the insufficient word coverage of the LM lexicon since it does not contain the words that make up a large part of the sentences of the evaluation dataset. Therefore, those expressions remain unanswered. As can be seen from Table \ref{table:number_sentences_across_models}, the sentiment analysis performed with the LM lexicon leads to a large number of unanswered sentences for each of the datasets. The percentage of unanswered sentences is about 60\% for each dataset. These unanswered expressions are considered wrongly answered, leading to very low accuracy of the LM lexicon. On the other hand, Table \ref{table:number_sentences_across_models} shows that there are almost no unanswered sentences when using explainable lexicons combined with the LM lexicon. Hence, the results show that explainable lexicons are advantageous over manually annotated lexicons as they achieve larger vocabulary coverage and higher accuracy in sentiment analysis. Explainable lexicons are able to achieve larger vocabulary coverage because they can automatically extract words and classify them using explainable ML models. Consequently, the combined lexicon also leads to a larger vocabulary coverage.

As evidenced by the data presented in Table \ref{table:results_from_models}, it can be observed that XLex consistently surpasses LM in all experiments, resulting in an overall increase of 0.431 in terms of classification accuracy. This improvement remains evident when we extend LM with XLex. The combined XLex+LM dictionary leads to an overall 0.450 increase in accuracy over LM. Moreover, as observed by Tables \ref{F1_Results}-\ref{MCC_Results} in Appendix \ref{Appendix_B}, it is noteworthy to highlight that the XLex+LM model consistently outperforms the LM model in terms of both F1 and MCC scores across all conducted experiments. Notably, the explainable lexicon XLex alone exhibits improvements over LM, leading to a 0.155 increase in F1 and a 0.090 enhancement in MCC. Even higher are the results achieved by the combined lexicon, XLex+LM, which demonstrates an increase of 0.226 in F1 and a 0.190 rise in MCC compared to LM. The enhancements in performance are determined by computing the difference between the average of the respective metric (accuracy, F1, or MCC) for XLex (XLex+LM) and the same metric averaged for LM. The metric's average is derived by averaging its values across all experiments. Table \ref{table:xlex_improvements_over_lm_summary} consolidates the performance enhancements of XLex and XLex+LM over LM in terms of accuracy, F1, and MCC.

\begin{table*}[]
\caption{Number of sentences on which the models based on a given lexicon did not give answers. The results are obtained based on the lexicons generated using the RoBERTa-based transformer model.}
\centering
\begin{tabular}{|l|c|cccc|}
\hline
\multicolumn{1}{|c|}{Evaluation dataset} & \multicolumn{1}{|c|}{\begin{tabular}[c]{@{}c@{}}Number of\\ sentences\end{tabular}} & \multicolumn{4}{c|}{Lexicons}                                                                                                            \\ \cline{3-6} 
                                    &                                                                                & \multicolumn{1}{c|}{fiqa\_fpb\_sentfin} & \multicolumn{1}{c|}{nasdaq} & \multicolumn{1}{c|}{financial\_phrase\_bank} & Loughran-McDonald \\ \hline
fiqa\_labeled\_df                   & 201                                                                            & \multicolumn{1}{c|}{3}                  & \multicolumn{1}{c|}{2}      & \multicolumn{1}{c|}{6}                       & 130               \\ \hline
sem\_eval                             & 353                                                                            & \multicolumn{1}{c|}{1}                  & \multicolumn{1}{c|}{0}      & \multicolumn{1}{c|}{2}                       & 224               \\ \hline
financial\_phrase\_bank             & 885                                                                            & \multicolumn{1}{c|}{0}                  & \multicolumn{1}{c|}{0}      & \multicolumn{1}{c|}{/}                       & 522               \\ \hline
fpb\_fiqa                           & 1542                                                                           & \multicolumn{1}{c|}{4}                  & \multicolumn{1}{c|}{3}      & \multicolumn{1}{c|}{8}                       & 937               \\ \hline
\end{tabular}
\label{table:number_sentences_across_models}
\end{table*}

\begin{table}[]
\caption{Average improvements of XLex and XLex+LM over LM in terms of accuracy, F1, and MCC. The values represent the differences in average metric scores (accuracy, F1, or MCC) between XLex (XLex+LM) and LM, both calculated by averaging the values of the corresponding metrics across all experiments.}
\centering
\resizebox{0.9\linewidth}{!}{%
\begin{tabular}{|l|ccc|}
\hline
        & \multicolumn{3}{c|}{Improvement over LM}                           \\ \hline
Model   & \multicolumn{1}{c|}{Acc}    & \multicolumn{1}{c|}{F1}     & MCC    \\ \hline
XLex (RoBERTa-based) & \multicolumn{1}{c|}{0.431} & \multicolumn{1}{c|}{0.155} & 0.090 \\ \hline
XLex+LM (RoBERTa-based) & \multicolumn{1}{c|}{0.450} & \multicolumn{1}{c|}{0.226} & 0.190 \\ \hline
XLex (FinBERT-based)& \multicolumn{1}{c|}{0.463} & \multicolumn{1}{c|}{0.227} & 0.183 \\ \hline
XLex+LM (FinBERT-based)& \multicolumn{1}{c|}{0.479} & \multicolumn{1}{c|}{0.250} & 0.253 \\ \hline

\end{tabular}%
}
\label{table:xlex_improvements_over_lm_summary}
\end{table}

To test XLex methodology's effectiveness in the worst-case scenario, we conducted an evaluation only on portions of the datasets where we have a recall from LM (i.e., filtering out all instances where the LM either provided no answer or were unable to make a decision). This creates a dataset with a strong bias in favor of LM, ultimately resulting in higher classification accuracy than the case when using LM on the whole dataset. For example, LM did not provide answers for 522 out of 885 sentences in the $financial\_phrase\_bank$ dataset. This means that the evaluation, in this case, was conducted on only 363 sentences, effectively reducing the original dataset by 59\% (Table \ref{table:number_sentences_across_models} illustrates the reduction in the size of the evaluation datasets). 

On this heavily constrained dataset towards LM, first, we tested the accuracy of the LM dictionary, and the results are shown in the ``LM on LM'' column in Table \ref{table:results_from_models_roberta}. Then, we applied the XLex on this LM-contained dataset, and we obtained slightly worse results (column ``XLex on LM''). The accuracy decreased by only 1\% in the case of the FinBERT-based model and by 2.2\% for the RoBERTa-based model. This experiment indicates that despite the fact that XLex is trained on a general dataset and is automatically created, it produces comparable results to the state-of-the-art expert-annotated dictionary in this worst-case scenario.

Furthermore, we wanted to explore if XLex could be used to extend LM in this worst-case scenario, so we evaluated the performance of the combined dictionary XLex+LM on the LM-constrained datasets. Our findings reveal that the combined dictionary always leads to improvement of the results (column ``(XLex+LM) on LM'' in Table \ref{table:results_from_models_roberta}). The average accuracy increased by 3\% for the FinBERT-based model and 2.65\% for the RoBERTa-based model. These results show that the proposed methodology can also be effectively used as an automated dictionary enhancement methodology that can help in extending the expert annotated dictionaries.

\begin{table*}[]
\caption{Comparison of the XLex-based model with the RoBERTa and FinBERT transformer models in terms of model speed and size. The XLex-based model utilizes various source datasets and undergoes evaluation across different evaluation datasets. The execution speed of the models is assessed in a CPU environment available within the free tier of Google Colab to ensure a fair comparison under identical conditions. The CPU environment uses an Intel Xeon CPU with one physical and two logical cores running at 2.20GHz, equipped with 12GB of RAM.}
\centering
\resizebox{\linewidth}{!}{%
\begin{tabular}{|l|c|c|c|c|c|c|c|}
\hline
Model &
  \begin{tabular}[c]{@{}c@{}}Source\\ dataset\end{tabular} &
  Is normalized? &
  \begin{tabular}[c]{@{}c@{}}Num. of words\\ in source dataset\end{tabular} &
  Eval dataset &
  \begin{tabular}[c]{@{}c@{}}Num. of sentences\\ in eval dataset\end{tabular} &
  Model size &
  \begin{tabular}[c]{@{}c@{}}Processing time (CPU)\\ in seconds\end{tabular} \\ \hline
\multirow{22}{*}{XLex-based model}         & nasdaq  & No  & 7094 & fpb\_fiqa               & 1542               & 363 KB  & 11.48  \\ \cline{2-8} 
                                           & nasdaq  & No  & 7094 & fiqa\_labeled\_df       & 201                & 363 KB  & 1.44   \\ \cline{2-8} 
                                           & nasdaq  & No  & 7094 & financial\_phrase\_bank & 885                & 363 KB  & 7.47   \\ \cline{2-8} 
                                           & nasdaq  & No  & 7094 & sem\_eval                 & 353                & 363 KB  & 1.56   \\ \cline{2-8} 
                                           & nasdaq  & Yes & 7094 & fpb\_fiqa               & 1542               & 363 KB  & 11.68   \\ \cline{2-8} 
                                           & nasdaq  & Yes & 7094 & fiqa\_labeled\_df       & 201                & 363 KB  & 1.44   \\ \cline{2-8} 
                                           & nasdaq  & Yes & 7094 & financial\_phrase\_bank & 885                & 363 KB  & 7.48   \\ \cline{2-8} 
                                           & nasdaq  & Yes & 7094 & sem\_eval                 & 353                & 363 KB  & 1.59   \\ \cline{2-8} 
                                           & financial\_phrase\_bank     & No  & 4344 & fpb\_fiqa               & 1542               & 202 KB  & 11.71   \\ \cline{2-8} 
                                           & financial\_phrase\_bank     & No  & 4344 & fiqa\_labeled\_df       & 201                & 202 KB  & 1.21   \\ \cline{2-8} 
                                           & financial\_phrase\_bank     & No  & 4344 & sem\_eval                 & 353                & 202 KB  & 1.46   \\ \cline{2-8} 
                                           & financial\_phrase\_bank     & Yes & 4344 & fpb\_fiqa               & 1542               & 202 KB  & 11.61   \\ \cline{2-8} 
                                           & financial\_phrase\_bank     & Yes & 4344 & fiqa\_labeled\_df       & 201                & 202 KB  & 1.44   \\ \cline{2-8} 
                                           & financial\_phrase\_bank     & Yes & 4344 & sem\_eval                 & 353                & 202 KB  & 1.44   \\ \cline{2-8} 
                                           & fiqa\_fpb\_sentfin\_neutral & No  & 4868 & fpb\_fiqa               & 1542               & 233 KB  & 11.7   \\ \cline{2-8} 
                                           & fiqa\_fpb\_sentfin\_neutral & No  & 4868 & fiqa\_labeled\_df       & 201                & 233 KB  & 1.29   \\ \cline{2-8} 
                                           & fiqa\_fpb\_sentfin\_neutral & No  & 4868 & financial\_phrase\_bank & 885                & 233 KB  & 7.53   \\ \cline{2-8} 
                                           & fiqa\_fpb\_sentfin\_neutral & No  & 4868 & sem\_eval                 & 353                & 233 KB  & 1.6   \\ \cline{2-8} 
                                           & fiqa\_fpb\_sentfin\_neutral & Yes & 4868 & fpb\_fiqa               & 1542               & 233 KB  & 11.74   \\ \cline{2-8} 
                                           & fiqa\_fpb\_sentfin\_neutral & Yes & 4868 & fiqa\_labeled\_df       & 201                & 233 KB  & 1.14   \\ \cline{2-8} 
                                           & fiqa\_fpb\_sentfin\_neutral & Yes & 4868 & financial\_phrase\_bank & 885                & 233 KB  & 7.51   \\ \cline{2-8} 
                                           & fiqa\_fpb\_sentfin\_neutral & Yes & 4868 & sem\_eval                 & 353                & 233 KB  & 1.6
                                           
                                           \\ \hline
\multirow{4}{*}{RoBERTa transformer model} & \NA     & \NA & \NA  & fpb\_fiqa               & 1542               & 1.32 GB & 960.05 \\ \cline{2-8} 
                                           & \NA     & \NA & \NA  & fiqa\_labeled\_df       & 201                & 1.32 GB & 111.5 \\ \cline{2-8} 
                                           & \NA     & \NA & \NA  & financial\_phrase\_bank & 885                & 1.32 GB & 600.12 \\ \cline{2-8} 
                                           & \NA     & \NA & \NA  & sem\_eval                 & 353                & 1.32 GB & 183.45  
                                           
                                           \\ \hline
\multirow{4}{*}{FinBERT transformer model} & \NA     & \NA & \NA  & fpb\_fiqa               & 1542               & 417.8 MB & 232.26 \\ \cline{2-8} 
                                           & \NA     & \NA & \NA  & fiqa\_labeled\_df       & 201                & 417.8 MB & 27.95 \\ \cline{2-8} 
                                           & \NA     & \NA & \NA  & financial\_phrase\_bank & 885                & 417.8 MB & 143.99 \\ \cline{2-8} 
                                           & \NA     & \NA & \NA  & sem\_eval                 & 353                & 417.8 MB & 38.02  \\ \hline                                           
\end{tabular}%
}
\label{table:model_comparison_speed_size}
\end{table*}

To gain a deeper understanding of the decision-making processes of XLex and LM, we present a detailed analysis of several text instances. We will use the test setup involving the standard XLex-based model (without normalization) in combination with $nasdaq$ and $financial\_phrase \_bank$ used as the source and evaluation dataset. As expected, our observations revealed that LM's errors stem from its insufficient word coverage. For instance, in the sentences \textit{``Finnish engineering and technology company Metso Oyj said on May 27, 2008, it completed the acquisition of paper machinery technology from Japanese engineering company Mitsubishi Heavy Industries (MHI) for an undisclosed sum''} and \textit{``Nokia also noted the average selling price of handsets declined during the period, though its mobile phone profit margin rose to more than 22 percent from 13 percent in the year-ago quarter''}, LM incorrectly places emphasis solely on the words ``undisclosed'' and ``decline'', respectively. LM classifies both words with a negative sentiment, disregarding the rest of the words in the respective sentences, which leads to inaccurate predictions. In contrast, XLex exhibits a larger vocabulary coverage and can assign sentiment scores to other relevant words in these sentences, resulting in accurate predictions. We also analyzed cases where XLex made errors while LM produced correct predictions, such as the sentences \textit{``Finnish airline Finnair is starting the temporary layoffs of cabin crews in February 2010''} and \textit{``The financial impact is estimated to be an annual improvement of EUR 2.0 m in the division's results, as of fiscal year 2008''}. In these cases, LM correctly classified the words ``layoffs'' and ``improvement'', respectively, which had a dominant impact on the sentiment classification for the two sentences. While XLex formed its decision score on more words, its classification ultimately resulted in an inaccurate prediction. However, for these cases, the combined lexicon XLex+LM resulted in accurate predictions.

It is worth mentioning that we explored two distinct approaches in our initial experimental setup, one utilizing FinBERT and the other employing RoBERTa. FinBERT is a pre-trained large language model specifically designed for financial text analysis \cite{huang2023finbert}. It is based on the BERT architecture and trained on a large corpus of financial text data to better understand and analyze financial language and documents. It has been reported that FinBERT performs well in financial applications compared to general-purpose language models. In our analysis, the FinBERT model demonstrated similar performance to RoBERTa with a slight accuracy improvement when utilizing XLex and XLex+LM. We want to note that we have made a deliberate choice to base our main analysis on RoBERTa, as FinBERT is fine-tuned on a closed proprietary dataset. In contrast, our model (using RoBERTa) is fine-tuned on publicly available datasets, enabling us to exercise precise control over the fine-tuning process while still achieving satisfactory results. All the details regarding the RoBERTa and FinBERT-based models are shown in Table \ref{table:results_from_models}, Table \ref{table:xlex_improvements_over_lm_summary}, 
Table \ref{table:model_comparison_speed_size} and Appendix \ref{Appendix_B}.\footnote{The source code and results about the comparison between RoBERTa and FinBERT can be found at: https://github.com/hristijanpeshov/SHAP-\\Explainable-Lexicon-Model/tree/master/notebooks} The results for FinBERT were obtained following the identical grid search procedure as that applied to RoBERTa.

Besides the achieved high accuracy and increased vocabulary coverage, the proposed explainable lexicons also lead to two additional benefits: speed and size. The speed for processing sentences is an important factor in real-time production systems. If NLP processing is worth doing at all in a system, it is worth doing it fast \cite{honnibal2019spacy}.

Table \ref{table:model_comparison_speed_size} shows a comparative analysis of model sizes and sentiment classification execution times for three models: the XLex-based model, featured in Section \ref{sec:5}, the fine-tuned RoBERTa transformer model, presented in Section \ref{sec:3}, and the FinBERT model. The analysis involves conducting experiments using the XLex-based model with lexicons learned from various source datasets, assessing its performance in sentiment classification across different evaluation datasets. We explore both normalized and non-normalized versions of the lexicons. In contrast, the RoBERTa and FinBERT-based models do not rely on source datasets by design, as they are pre-trained models that can be readily used for sentiment classification. Consequently, the corresponding entries in columns 2-4 of Table \ref{table:model_comparison_speed_size} remain empty.

The execution speed of the models is evaluated across the evaluation datasets using a central processing unit (CPU) in Google Colab to ensure a fair comparison under identical conditions. The execution time of each model is determined by calculating the average of the times recorded from 10 experimental runs. For the experiments, we used Google Colab's free tier, which provides an Intel Xeon CPU with one physical and two logical cores running at 2.20GHz paired with 12GB of RAM. As can be seen in Table \ref{table:model_comparison_speed_size}, the results reveal a substantial difference in the execution time of the models. The XLex-based model leads to a significantly smaller execution time compared to the RoBERTa and FinBERT transformer models by a factor of about 87 and 21, respectively. The factor is determined by dividing the average CPU speed of the RoBERTa (FinBERT) model by that of the XLex-based model (the averages are calculated considering the respective experiments for each of the two models). This makes the lexicon-based model suitable for tasks that need to be performed quickly and in real-time and still lead to reasonably accurate predictions.

\begin{table}[ht]
    \caption{Classification report showcasing the performance of the XLex-based model (using RoBERTa) based on the combined XLex+LM lexicon. The model is constructed using the lexicon created with the $nasdaq$ dataset as its source dataset. The model evaluation is performed on the $financial\_phrase\_bank$ dataset. This model achieves the highest accuracy among all XLex+LM models tested, achieving an accuracy rate of 84.3\%.}
    \centering
    \resizebox{\columnwidth}{!}{%
    \begin{tabular}{rrrrr}
        \multicolumn{1}{l}{}\phantom{C1, C2, C3, C4} & precision & recall & f1-score & support                          \\
        \multicolumn{1}{l}{} & \multicolumn{1}{l}{} & \multicolumn{1}{l}{} & \multicolumn{1}{l}{} & \multicolumn{1}{l}{} \\
        Negative sentences                         & 0.43 & 0.72  & 0.54 & 111  \\
        Positive sentences                         & 0.96 & 0.86  & 0.91 & 774  \\
                                                   &      &       &      &      \\
        accuracy                                   &      &       & 0.84 & 885  \\
        macro avg                                  & 0.69 & 0.79  & 0.72 & 885  \\
        weighted avg                               & 0.89 & 0.84  & 0.86 & 885  \\
    \end{tabular}%
}
\label{table:classification_report}
\end{table}

Similar to other neural network models, RoBERTa can leverage parallel architectures to enhance its processing speed. Employing RoBERTa on a GPU, as opposed to a CPU, yields a substantial reduction in execution time, achieving an overall speedup factor of approximately 26. The GPU tests were performed on the NVIDIA Tesla T4 GPU computing environment available within Google Colab's free tier. Although the XLex-based model can be parallelized, its current implementation lacks the necessary capabilities for parallel processing. As a result, evaluating its performance in a GPU environment would lead to an unfair comparison. Consequently, Table \ref{table:model_comparison_speed_size} presents results only for the CPU comparison. The parallelization of the XLex-based model goes beyond the scope of this paper; however, it can serve as a potential avenue for future work.

In the GPU environment, we perform the word extraction process using SHAP and transformer models. The word extraction process with the RoBERTa and FinBERT models on 9202 sentences from the $nasdaq$ cf dataset took approximately 23 and 10 hours, respectively. The extended time required for word extraction is primarily attributed to the slow performance and time-consuming nature of SHAP's operations.

The XLex methodology provides flexibility in selecting the underlying transformer model, allowing for the easy integration of any preferred model. Thus, we were also interested in assessing the speed performance of the FinBERT model. We conducted tests in the same CPU environment. Table \ref{table:model_comparison_speed_size} shows that the FinBERT model leads to better execution speeds than the RoBERTa model. However, the FinBERT model exhibits a smaller size compared to RoBERTa. Similarly to RoBERTa, the FinBERT model also underwent testing on a GPU, resulting in a nearly 12-fold improvement in execution speed compared to running it on a CPU. As expected, the XLex-based model outperforms the FinBERT model in terms of speed.

The second important aspect of the lexicon-based model is the size. The model size is an important factor to consider when deciding which model to be used in production systems. Transformer models are trained on large datasets and are often larger than the free disk space available on resource-constrained devices. For the lexicon-based model proposed in Section \ref{sec:5}, the model size is actually represented by the size of the lexicon. The size comparison between the RoBERTa transformer model and the lexicon-based model is shown in Table \ref{table:model_comparison_speed_size}. The difference in size is considerable. As can be seen, the lexicon-based model is about three orders of magnitude smaller than the RoBERTa transformer model. The XLex-based model exhibits an identical size for each specific source dataset, with the XLex-based model registering sizes of 363KB, 202KB, and 233KB for the $nasdaq$, $financial\_phrase\_bank$ ($fpb$), and $fiqa\_fpb\_sentfin\_neutral$ source datasets, respectively. Additionally, the RoBERTa-based model maintains a size of 1.32GB, regardless of the evaluation dataset used. The size of the FinBERT model is 417.8MB. Although there are approaches to make transformer-based models smaller \cite{lathia2019when}, transformers are not suitable for certain use cases, such as environments with limited computational resources or embedded devices. On the other hand, Table \ref{table:model_comparison_speed_size} shows that lexicon-based models have a size that is suitable for such applications.

Another important advantage of lexicon-based approaches is their interpretability. Lexicon-based sentiment models are generally more interpretable than transformer-based sentiment models because they rely on a pre-defined set of rules that are easy to understand and interpret. In a lexicon-based sentiment model, each word is assigned a sentiment score based on its associated sentiment value in the sentiment dictionary, and the overall sentiment of the text is calculated based on the sum or average of the sentiment scores of the words in the text. This makes it easy to understand why a particular text was classified as positive, negative, or neutral, as the sentiment scores assigned to each word in the text are transparent and interpretable. Moreover, the sentiment dictionary can be customized for specific domains or use cases, allowing for more accurate and relevant sentiment analysis. In contrast, transformer-based sentiment models are based on more complex deep learning architectures that are more difficult to interpret. Transformer models use large neural networks to learn the context and meaning of words in a sentence and assign a sentiment score to the sentence based on this understanding. While transformer-based sentiment models can achieve higher accuracy than dictionary-based models, the sentiment scores assigned to each word or phrase in the sentence are not as transparent or interpretable as they are generated by a complex neural network that learns its own set of rules based on the training data. This lack of interpretability can be a limitation for applications where it is important to understand why a particular text was classified as positive, negative, or neutral. Transformer-based sentiment models can be useful for tasks that require a more nuanced understanding of sentiment, as they can capture the complex relationships between words and the context in which they are used. However, it is important to note that while explainable AI (XAI) methods like SHAP can provide some level of interpretability for transformer-based models, they may not always provide a complete understanding of how the model works due to the black-box nature of its inner workings. Additionally, the interpretability of XAI methods is often limited by the complexity of the model and may not be able to fully capture the nuances of natural language. Therefore, it is important to use XAI methods in conjunction with other approaches to ensure accurate and reliable sentiment analysis.

Due to their inherent advantages, explainable lexicons could potentially be used to replace standard lexicons that are nowadays still established in various domains. For example, the proposed lexicon in this paper could be used to replace the LM lexicon in the domain of finance. However, it is essential to use domain experts before nominating an explainable lexicon as the new standard for a specific application or domain. The domain experts can give an expert opinion when validating the sentiment scores of its constituent words. The involvement of domain experts in the lexicon review process could be a possible direction of future research as it could improve transparency and objectivity. Only then we can have a lexicon that is not only superior in terms of speed, size, and interpretability but also validated by human experts. This is especially important in critical applications where the quality of the results directly affects people's lives or safety. Examples of such applications may include not only finance but also knowledge extraction in medicine, legal document analysis, and risk assessment. By having an expert review validate the lexicon, the dictionary becomes more accurate and reliable, enhancing its usefulness and value to the users. It also ensures that the lexicon is consistent with the standard conventions of the language while meeting the needs and expectations of the intended audience.

\section{Conclusion}
\label{sec:7}

In this paper, we present a novel XLex methodology that leverages NLP transformer models and SHAP explainability to automatically enhance the vocabulary coverage of the Loughran-McDonald (LM) lexicon in sentiment analysis scenarios for financial applications. Our results demonstrate that standard domain-specific lexicons, such as the LM lexicon, can be expanded in an explainable way with new words without the need for laborious annotation involvement of human experts, a process that is both expensive and time-consuming.

To ensure the robustness of our findings, we employ a multi-faceted validation strategy that integrates multiple datasets. Specifically, we learn the lexicon on one dataset and subsequently test its effectiveness on various other datasets. We have conducted 22 separate experiments, and in all of them, the proposed XLex methodology leads to increased performance compared to LM. Specifically, we evaluated the XLex methodology in two separate instances: one employing a fine-tuned RoBERTa-based model and the other utilizing a pre-trained FinBERT model. The results yielded are largely comparable while also emphasizing the robustness of the XLex methodology and its effectiveness in working with various transformer models.

The use of generated (XLex) or combined lexicons (XLex+LM) leads to significant improvements in sentiment analysis results compared to using the manually annotated lexicon alone. This improvement is demonstrated by higher accuracy and larger vocabulary coverage, directly addressing the limitations of standard, manually annotated lexicons.

Overall, the proposed XLex methodology holds great promise in advancing the field of sentiment analysis, particularly in applications where interpretability is of utmost importance. Unlike transformer models that rely on complex inner workings of neural networks, lexicon models depend on pre-defined rules, making it easy to interpret why a particular text is classified as positive, negative, or neutral. The enhanced interpretability provided by explainable lexicons makes them especially well-suited for critical applications where the quality of the results directly affects people's lives or safety. Examples of such applications include finance, medicine, legal document analysis, and risk assessment. In these areas, the transparency and explainability of the analysis process are essential for building trust and ensuring the responsible use of AI technologies.

Our study highlights the performance improvements of XLex, achieved even with a simple optimization algorithm based on a grid search procedure characterized by a relatively low granularity. One avenue for future work is to improve the grid search by increasing granularity or incorporating more advanced techniques, such as Bayesian optimization. This would enable a more precise discovery of optimal model parameters, potentially resulting in additional gains in accuracy, F1, and MCC. Furthermore, the parallelization of the XLex-based model could serve as a viable direction for future research given its potential to further increase the computational efficiency of XLex.

Additionally, it would be beneficial to investigate the integration of explainable lexicons with other NLP techniques, to further enhance the performance and applicability of sentiment analysis. It is also essential to evaluate the robustness of explainable lexicons against various challenges, such as changes in language use, evolving domains, and the presence of adversarial examples.

The proposed methodology is general and adaptable, offering opportunities for future research to explore its application across other domains beyond finance. Adopting the XLex methodology for different domains has the potential to significantly impact various industries, enhancing the accuracy and interpretability of sentiment analysis results while reducing the time and cost associated with manual lexicon development.

\section{ACKNOWLEDGMENT}
This work is partially based on COST Action CA18209 – NexusLinguarum ``European network for Web-centred linguistic data
science'', supported by COST (European Cooperation in Science and Technology).

\onecolumn
\clearpage
\appendices
\section{ }
\label{Appendix_A}

\setcounter{table}{0}
\renewcommand\thetable{\thesection.\arabic{table}}

\begin{table}[h]
\caption{Features of the explainable lexicon after adding the XLex prefix.}
\centering
\resizebox{\columnwidth}{!}{%
\begin{tabular}{|c|c|c|c|c|c|c|c|c|c|c|c|}
\hline
Word &
  \begin{tabular}[c]{@{}c@{}}XLex Count\\ (Selected)\end{tabular} &
  XLex Total &
  \begin{tabular}[c]{@{}c@{}}XLex Count\\ (Opposite)\end{tabular} &
  XLex Category &
  \begin{tabular}[c]{@{}c@{}}XLex Sum\\ SHAP Value\\ (Selected)\end{tabular} &
  \begin{tabular}[c]{@{}c@{}}XLex Average\\ SHAP Value\\ (Selected)\end{tabular} &
  \begin{tabular}[c]{@{}c@{}}XLex Max\\ SHAP Value\\ (Selected)\end{tabular} &
  \begin{tabular}[c]{@{}c@{}}XLex Min\\ SHAP Value\\ (Selected)\end{tabular} &
  \begin{tabular}[c]{@{}c@{}}XLex Ratio\\ (Selected)\end{tabular} &
  \begin{tabular}[c]{@{}c@{}}XLex Sum\\ SHAP Value\\ (Opposite)\end{tabular} &
  \begin{tabular}[c]{@{}c@{}}XLex Average\\ SHAP Value\\ (Opposite)\end{tabular} \\ \hline
abet      & 1  & 1  & 0  & positive & 0.025090 & 0.025090 & 0.025090 & 0.025090 & 1.000000 & 0.000000 & 0.000000 \\ \hline
abide     & 1  & 1  & 0  & positive & 0.003838 & 0.003838 & 0.003838 & 0.003838 & 1.000000 & 0.000000 & 0.000000 \\ \hline
abo       & 1  & 1  & 0  & positive & 0.000976 & 0.000976 & 0.000976 & 0.000976 & 1.000000 & 0.000000 & 0.000000 \\ \hline
aboard    & 2  & 2  & 0  & positive & 0.245279 & 0.122640 & 0.210718 & 0.034561 & 1.000000 & 0.000000 & 0.000000 \\ \hline
abolition & 1  & 1  & 0  & positive & 0.006430 & 0.006430 & 0.006430 & 0.006430 & 1.000000 & 0.000000 & 0.000000 \\ \hline
ken       & 4  & 7  & 3  & negative & 0.114558 & 0.028640 & 0.086224 & 0.003522 & 0.504271 & 0.084463 & 0.028154 \\ \hline
peter     & 9  & 15 & 6  & negative & 0.146561 & 0.016285 & 0.068352 & 0.000112 & 0.477515 & 0.106909 & 0.017818 \\ \hline
uri       & 1  & 2  & 1  & negative & 0.016012 & 0.016012 & 0.016012 & 0.016012 & 0.565765 & 0.012289 & 0.012289 \\ \hline
military  & 76 & 98 & 22 & negative & 2.958678 & 0.038930 & 0.169050 & 0.001473 & 0.773876 & 0.250255 & 0.011375 \\ \hline
depth     & 1  & 2  & 1  & positive & 0.018833 & 0.018833 & 0.018833 & 0.018833 & 0.997290 & 0.000051 & 0.000051 \\ \hline
\end{tabular}%
}
\label{table:explainable_lexicon_with_prefix}
\end{table}

\begin{table}[h]
\caption{Features of the LM lexicon after adding the LM prefix.}
\centering
\resizebox{\columnwidth}{!}{%
\begin{tabular}{|c|c|c|c|c|c|c|c|c|c|c|c|}
\hline
Word &
  \begin{tabular}[c]{@{}c@{}}LM Count\\ (Selected)\end{tabular} &
  LM Total &
  \begin{tabular}[c]{@{}c@{}}LM Count\\ (Opposite)\end{tabular} &
  LM Category &
  \begin{tabular}[c]{@{}c@{}}LM Sum\\ SHAP Value\\ (Selected)\end{tabular} &
  \begin{tabular}[c]{@{}c@{}}LM Average\\ SHAP Value\\ (Selected)\end{tabular} &
  \begin{tabular}[c]{@{}c@{}}LM Max\\ SHAP Value\\ (Selected)\end{tabular} &
  \begin{tabular}[c]{@{}c@{}}LM Min\\ SHAP Value\\ (Selected)\end{tabular} &
  \begin{tabular}[c]{@{}c@{}}LM Ratio\\ (Selected)\end{tabular} &
  \begin{tabular}[c]{@{}c@{}}LM Sum\\ SHAP Value\\ (Opposite)\end{tabular} &
  \begin{tabular}[c]{@{}c@{}}LM Average\\ SHAP Value\\ (Opposite)\end{tabular} \\ \hline
abet         & 1 & 1 & 0 & negative & 1 & 1 & 1 & 1 & 1 & 0 & 0 \\ \hline
accomplish   & 1 & 1 & 0 & positive & 1 & 1 & 1 & 1 & 1 & 0 & 0 \\ \hline
advance      & 1 & 1 & 0 & positive & 1 & 1 & 1 & 1 & 1 & 0 & 0 \\ \hline
advantage    & 1 & 1 & 0 & positive & 1 & 1 & 1 & 1 & 1 & 0 & 0 \\ \hline
advantageous & 1 & 1 & 0 & positive & 1 & 1 & 1 & 1 & 1 & 0 & 0 \\ \hline
writeoff     & 1 & 1 & 0 & negative & 1 & 1 & 1 & 1 & 1 & 0 & 0 \\ \hline
writeoffs    & 1 & 1 & 0 & negative & 1 & 1 & 1 & 1 & 1 & 0 & 0 \\ \hline
wrongful     & 1 & 1 & 0 & negative & 1 & 1 & 1 & 1 & 1 & 0 & 0 \\ \hline
wrongfully   & 1 & 1 & 0 & negative & 1 & 1 & 1 & 1 & 1 & 0 & 0 \\ \hline
wrongly      & 1 & 1 & 0 & negative & 1 & 1 & 1 & 1 & 1 & 0 & 0 \\ \hline
\end{tabular}%
}
\label{table:LM_lexicon_with_prefix}
\end{table}

\begin{table}[h]
\caption{Values of selected features of the combined lexicon for words that appear in both the explainable and LM lexicon.}
\centering
\resizebox{\columnwidth}{!}{%
\begin{tabular}{|c|c|c|c|c|c|c|c|c|c|c|}
\hline
Word &
  \begin{tabular}[c]{@{}c@{}}XLex Count\\ (Selected)\end{tabular} &
  XLex Total &
  \begin{tabular}[c]{@{}c@{}}XLex Count\\ (Opposite)\end{tabular} &
  \begin{tabular}[c]{@{}c@{}}XLex Average\\ SHAP Value\\ (Selected)\end{tabular} &
  XLex Source &
  \begin{tabular}[c]{@{}c@{}}LM Average\\ SHAP Value\\ (Selected)\end{tabular} &
  LM Category &
  \begin{tabular}[c]{@{}c@{}}LM Sum\\ SHAP Value\\ (Opposite)\end{tabular} &
  LM Source &
  \begin{tabular}[c]{@{}c@{}}LM Max\\ SHAP Value\\ (Opposite)\end{tabular} \\ \hline
abet         & 1.0  & 1.0  & 0.0  & 0.025090 & XLex & 1 & negative & 0 & LM & 0 \\ \hline
accomplish   & 1.0  & 1.0  & 0.0  & 0.078244 & XLex & 1 & positive & 0 & LM & 0 \\ \hline
advance      & 9.0  & 9.0  & 0.0  & 0.113294 & XLex & 1 & positive & 0 & LM & 0 \\ \hline
advantage    & 7.0  & 7.0  & 0.0  & 0.431898 & XLex & 1 & positive & 0 & LM & 0 \\ \hline
advantageous & 1.0  & 1.0  & 0.0  & 0.441719 & XLex & 1 & positive & 0 & LM & 0 \\ \hline
good         & 65.0 & 68.0 & 3.0  & 0.149197 & XLex & 1 & positive & 0 & LM & 0 \\ \hline
pose         & 10.0 & 20.0 & 10.0 & 0.027118 & XLex & 1 & negative & 0 & LM & 0 \\ \hline
gain         & 14.0 & 16.0 & 2.0  & 0.157810 & XLex & 1 & positive & 0 & LM & 0 \\ \hline
evasion      & 2.0  & 3.0  & 1.0  & 0.032876 & XLex & 1 & negative & 0 & LM & 0 \\ \hline
defeat       & 10.0 & 12.0 & 2.0  & 0.077834 & XLex & 1 & negative & 0 & LM & 0 \\ \hline
\end{tabular}%
}
\label{table:combined_lexicon_words_in_both_explainable_and_LM}
\end{table}

\begin{table}[h]
\caption{Values of selected features of the combined lexicon for words that appear in either the explainable or the LM lexicon.}
\centering
\resizebox{\columnwidth}{!}{%
\begin{tabular}{|c|c|c|c|c|c|c|c|c|c|c|}
\hline
Word &
  \begin{tabular}[c]{@{}c@{}}XLex Count\\ (Selected)\end{tabular} &
  XLex Total &
  \begin{tabular}[c]{@{}c@{}}XLex Count\\ (Opposite)\end{tabular} &
  \begin{tabular}[c]{@{}c@{}}XLex Average\\ SHAP Value\\ (Selected)\end{tabular} &
  XLex Source &
  \begin{tabular}[c]{@{}c@{}}LM Average\\ SHAP Value\\ (Selected)\end{tabular} &
  LM Category &
  \begin{tabular}[c]{@{}c@{}}LM Sum\\ SHAP Value\\ (Opposite)\end{tabular} &
  LM Source &
  \begin{tabular}[c]{@{}c@{}}LM Max\\ SHAP Value\\ (Opposite)\end{tabular} \\ \hline
abide      & 1.0 & 1.0 & 0.0 & 0.003838 & XLex & NaN & NaN      & NaN & NaN & NaN \\ \hline
abo        & 1.0 & 1.0 & 0.0 & 0.000976 & XLex & NaN & NaN      & NaN & NaN & NaN \\ \hline
aboard     & 2.0 & 2.0 & 0.0 & 0.122640 & XLex & NaN & NaN      & NaN & NaN & NaN \\ \hline
abolition  & 1.0 & 1.0 & 0.0 & 0.006430 & XLex & NaN & NaN      & NaN & NaN & NaN \\ \hline
abroad     & 3.0 & 3.0 & 0.0 & 0.039073 & XLex & NaN & NaN      & NaN & NaN & NaN \\ \hline
writeoff   & NaN & NaN & NaN & NaN      & NaN        & 1   & negative & 0   & LM  & 0   \\ \hline
writeoffs  & NaN & NaN & NaN & NaN      & NaN        & 1   & negative & 0   & LM  & 0   \\ \hline
wrongful   & NaN & NaN & NaN & NaN      & NaN        & 1   & negative & 0   & LM  & 0   \\ \hline
wrongfully & NaN & NaN & NaN & NaN      & NaN        & 1   & negative & 0   & LM  & 0   \\ \hline
wrongly    & NaN & NaN & NaN & NaN      & NaN        & 1   & negative & 0   & LM  & 0   \\ \hline
\end{tabular}%
}
\label{table:either_explainable_or_LM}
\end{table}

\section{ }
\label{Appendix_B}

\setcounter{table}{0}
\renewcommand\thetable{\thesection.\arabic{table}}

\begin{table*}[]
\caption{F1 score obtained across different experiments using the XLex methodology based on the RoBERTa transformer model. The highest values are represented in bold.}
\centering
\small
\setlength\tabcolsep{2pt}

\begin{tabular}{|c|c|c|ccc|ccc|}
\hline
\multicolumn{1}{|c|}{Source of the lexicon} &
  \multicolumn{1}{|c|}{Normalized} &
  \multicolumn{1}{|c|}{Evaluation set} &
  \multicolumn{3}{c|}{F1 on whole dataset} & \multicolumn{3}{c|}{\begin{tabular}[c]{@{}c@{}}F1 only on the part of the dataset for\\which the LM-based model has an answer
  \end{tabular}} \\ 
  \cline{4-9} 
 &
   &
   &
  \multicolumn{1}{c|}{LM} &
  \multicolumn{1}{c|}{XLex} &
  \multicolumn{1}{c|}{\begin{tabular}[c]{@{}c@{}}XLex+\\ LM\end{tabular}} &
  \multicolumn{1}{c|}{\begin{tabular}[c]{@{}c@{}}LM on\\ LM\end{tabular}} &
  \multicolumn{1}{c|}{\begin{tabular}[c]{@{}c@{}}XLex on\\ LM\end{tabular}} &
  \begin{tabular}[c]{@{}c@{}}(XLex+LM)\\ on LM\end{tabular} \\ \hline
nasdaq &
Yes &
financial\_phrase\_bank &
\multicolumn{1}{c|}{0.287} &
\multicolumn{1}{c|}{0.451} &
\multicolumn{1}{c|}{{\textbf{0.714}}} &
\multicolumn{1}{c|}{0.688} &
\multicolumn{1}{c|}{0.448} &
\multicolumn{1}{c|}{{\textbf{0.731}}} \\ \hline
nasdaq &
Yes &
fiqa\_labeled\_df &
\multicolumn{1}{c|}{0.297} &
\multicolumn{1}{c|}{0.457} &
\multicolumn{1}{c|}{{\textbf{0.473}}} &
\multicolumn{1}{c|}{0.789} &
\multicolumn{1}{c|}{0.739} &
\multicolumn{1}{c|}{{\textbf{0.802}}} \\ \hline
nasdaq &
Yes &
fpb\_fiqa &
\multicolumn{1}{c|}{0.295} &
\multicolumn{1}{c|}{0.434} &
\multicolumn{1}{c|}{{\textbf{0.466}}} &
\multicolumn{1}{c|}{0.726} &
\multicolumn{1}{c|}{0.437} &
\multicolumn{1}{c|}{{\textbf{0.739}}} \\ \hline
nasdaq &
Yes &
sem\_eval &
\multicolumn{1}{c|}{0.305} &
\multicolumn{1}{c|}{0.664} &
\multicolumn{1}{c|}{{\textbf{0.721}}} &
\multicolumn{1}{c|}{0.752} &
\multicolumn{1}{c|}{0.704} &
\multicolumn{1}{c|}{{\textbf{0.775}}} \\ \hline
nasdaq &
No &
financial\_phrase\_bank &
\multicolumn{1}{c|}{0.287} &
\multicolumn{1}{c|}{0.457} &
\multicolumn{1}{c|}{{\textbf{0.72}}} &
\multicolumn{1}{c|}{0.688} &
\multicolumn{1}{c|}{0.446} &
\multicolumn{1}{c|}{{\textbf{0.726}}} \\ \hline
nasdaq &
No &
fiqa\_labeled\_df &
\multicolumn{1}{c|}{0.297} &
\multicolumn{1}{c|}{0.437} &
\multicolumn{1}{c|}{{\textbf{0.454}}} &
\multicolumn{1}{c|}{0.789} &
\multicolumn{1}{c|}{0.739} &
\multicolumn{1}{c|}{{\textbf{0.802}}} \\ \hline
nasdaq &
No &
fpb\_fiqa &
\multicolumn{1}{c|}{0.295} &
\multicolumn{1}{c|}{0.426} &
\multicolumn{1}{c|}{{\textbf{0.459}}} &
\multicolumn{1}{c|}{0.726} &
\multicolumn{1}{c|}{0.433} &
\multicolumn{1}{c|}{{\textbf{0.738}}} \\ \hline
nasdaq &
No &
sem\_eval &
\multicolumn{1}{c|}{0.305} &
\multicolumn{1}{c|}{0.672} &
\multicolumn{1}{c|}{{\textbf{0.713}}} &
\multicolumn{1}{c|}{0.752} &
\multicolumn{1}{c|}{0.717} &
\multicolumn{1}{c|}{{\textbf{0.759}}} \\ \hline
fiqa\_fpb\_sentfin\_neutral &
Yes &
financial\_phrase\_bank &
\multicolumn{1}{c|}{0.287} &
\multicolumn{1}{c|}{0.451} &
\multicolumn{1}{c|}{{\textbf{0.689}}} &
\multicolumn{1}{c|}{0.688} &
\multicolumn{1}{c|}{0.474} &
\multicolumn{1}{c|}{{\textbf{0.731}}} \\ \hline
fiqa\_fpb\_sentfin\_neutral &
Yes &
fiqa\_labeled\_df &
\multicolumn{1}{c|}{0.297} &
\multicolumn{1}{c|}{0.434} &
\multicolumn{1}{c|}{{\textbf{0.456}}} &
\multicolumn{1}{c|}{0.789} &
\multicolumn{1}{c|}{0.752} &
\multicolumn{1}{c|}{{\textbf{0.827}}} \\ \hline
fiqa\_fpb\_sentfin\_neutral &
Yes &
fpb\_fiqa &
\multicolumn{1}{c|}{0.295} &
\multicolumn{1}{c|}{0.424} &
\multicolumn{1}{c|}{{\textbf{0.448}}} &
\multicolumn{1}{c|}{0.726} &
\multicolumn{1}{c|}{0.458} &
\multicolumn{1}{c|}{{\textbf{0.751}}} \\ \hline
fiqa\_fpb\_sentfin\_neutral &
Yes &
sem\_eval &
\multicolumn{1}{c|}{0.305} &
\multicolumn{1}{c|}{0.434} &
\multicolumn{1}{c|}{{\textbf{0.476}}} &
\multicolumn{1}{c|}{0.752} &
\multicolumn{1}{c|}{0.676} &
\multicolumn{1}{c|}{{\textbf{0.775}}} \\ \hline
fiqa\_fpb\_sentfin\_neutral &
No &
financial\_phrase\_bank &
\multicolumn{1}{c|}{0.287} &
\multicolumn{1}{c|}{0.448} &
\multicolumn{1}{c|}{{\textbf{0.694}}} &
\multicolumn{1}{c|}{0.688} &
\multicolumn{1}{c|}{0.463} &
\multicolumn{1}{c|}{{\textbf{0.731}}} \\ \hline
fiqa\_fpb\_sentfin\_neutral &
No &
fiqa\_labeled\_df &
\multicolumn{1}{c|}{0.297} &
\multicolumn{1}{c|}{0.416} &
\multicolumn{1}{c|}{{\textbf{0.439}}} &
\multicolumn{1}{c|}{0.789} &
\multicolumn{1}{c|}{0.747} &
\multicolumn{1}{c|}{{\textbf{0.827}}} \\ \hline
fiqa\_fpb\_sentfin\_neutral &
No &
fpb\_fiqa &
\multicolumn{1}{c|}{0.295} &
\multicolumn{1}{c|}{0.425} &
\multicolumn{1}{c|}{{\textbf{0.448}}} &
\multicolumn{1}{c|}{0.726} &
\multicolumn{1}{c|}{0.457} &
\multicolumn{1}{c|}{{\textbf{0.748}}} \\ \hline
fiqa\_fpb\_sentfin\_neutral &
No &
sem\_eval &
\multicolumn{1}{c|}{0.305} &
\multicolumn{1}{c|}{0.434} &
\multicolumn{1}{c|}{{\textbf{0.48}}} &
\multicolumn{1}{c|}{0.752} &
\multicolumn{1}{c|}{0.66} &
\multicolumn{1}{c|}{{\textbf{0.775}}} \\ \hline
financial\_phrase\_bank &
Yes &
fiqa\_labeled\_df &
\multicolumn{1}{c|}{0.297} &
\multicolumn{1}{c|}{0.409} &
\multicolumn{1}{c|}{{\textbf{0.447}}} &
\multicolumn{1}{c|}{0.789} &
\multicolumn{1}{c|}{0.688} &
\multicolumn{1}{c|}{{\textbf{0.84}}} \\ \hline
financial\_phrase\_bank &
Yes &
fpb\_fiqa &
\multicolumn{1}{c|}{0.295} &
\multicolumn{1}{c|}{0.4} &
\multicolumn{1}{c|}{{\textbf{0.43}}} &
\multicolumn{1}{c|}{0.726} &
\multicolumn{1}{c|}{0.428} &
\multicolumn{1}{c|}{{\textbf{0.739}}} \\ \hline
financial\_phrase\_bank &
Yes &
sem\_eval &
\multicolumn{1}{c|}{0.305} &
\multicolumn{1}{c|}{0.416} &
\multicolumn{1}{c|}{{\textbf{0.445}}} &
\multicolumn{1}{c|}{0.752} &
\multicolumn{1}{c|}{0.462} &
\multicolumn{1}{c|}{{\textbf{0.767}}} \\ \hline
financial\_phrase\_bank &
No &
fiqa\_labeled\_df &
\multicolumn{1}{c|}{0.297} &
\multicolumn{1}{c|}{0.43} &
\multicolumn{1}{c|}{{\textbf{0.456}}} &
\multicolumn{1}{c|}{0.789} &
\multicolumn{1}{c|}{0.692} &
\multicolumn{1}{c|}{{\textbf{0.802}}} \\ \hline
financial\_phrase\_bank &
No &
fpb\_fiqa &
\multicolumn{1}{c|}{0.295} &
\multicolumn{1}{c|}{0.413} &
\multicolumn{1}{c|}{{\textbf{0.439}}} &
\multicolumn{1}{c|}{0.726} &
\multicolumn{1}{c|}{0.436} &
\multicolumn{1}{c|}{{\textbf{0.738}}} \\ \hline
financial\_phrase\_bank &
No &
sem\_eval &
\multicolumn{1}{c|}{0.305} &
\multicolumn{1}{c|}{0.416} &
\multicolumn{1}{c|}{{\textbf{0.444}}} &
\multicolumn{1}{c|}{0.752} &
\multicolumn{1}{c|}{0.467} &
\multicolumn{1}{c|}{{\textbf{0.767}}} \\ \hline

\end{tabular}
\label{F1_Results}
\end{table*}

\begin{table*}[]
\caption{MCC obtained across different experiments using the XLex methodology based on the RoBERTa transformer model. The highest values are represented in bold.}
\centering
\small
\setlength\tabcolsep{2pt}

\begin{tabular}{|c|c|c|ccc|ccc|}
\hline
\multicolumn{1}{|c|}{Source of the lexicon} &
  \multicolumn{1}{|c|}{Normalized} &
  \multicolumn{1}{|c|}{Evaluation set} &
  \multicolumn{3}{c|}{MCC on whole dataset} & \multicolumn{3}{c|}{\begin{tabular}[c]{@{}c@{}}MCC only on the part of the dataset for\\which the LM-based model has an answer
  \end{tabular}} \\ 
  \cline{4-9} 
 &
   &
   &
  \multicolumn{1}{c|}{LM} &
  \multicolumn{1}{c|}{XLex} &
  \multicolumn{1}{c|}{\begin{tabular}[c]{@{}c@{}}XLex+\\ LM\end{tabular}} &
  \multicolumn{1}{c|}{\begin{tabular}[c]{@{}c@{}}LM on\\ LM\end{tabular}} &
  \multicolumn{1}{c|}{\begin{tabular}[c]{@{}c@{}}XLex on\\ LM\end{tabular}} &
  \begin{tabular}[c]{@{}c@{}}(XLex+LM)\\ on LM\end{tabular} \\ \hline
nasdaq &
Yes &
financial\_phrase\_bank &
\multicolumn{1}{c|}{0.192} &
\multicolumn{1}{c|}{0.362} &
\multicolumn{1}{c|}{{\textbf{0.453}}} &
\multicolumn{1}{c|}{0.489} &
\multicolumn{1}{c|}{0.364} &
\multicolumn{1}{c|}{{\textbf{0.539}}} \\ \hline
nasdaq &
Yes &
fiqa\_labeled\_df &
\multicolumn{1}{c|}{0.216} &
\multicolumn{1}{c|}{0.372} &
\multicolumn{1}{c|}{{\textbf{0.432}}} &
\multicolumn{1}{c|}{0.594} &
\multicolumn{1}{c|}{0.492} &
\multicolumn{1}{c|}{{\textbf{0.615}}} \\ \hline
nasdaq &
Yes &
fpb\_fiqa &
\multicolumn{1}{c|}{0.209} &
\multicolumn{1}{c|}{0.306} &
\multicolumn{1}{c|}{{\textbf{0.418}}} &
\multicolumn{1}{c|}{0.519} &
\multicolumn{1}{c|}{0.312} &
\multicolumn{1}{c|}{{\textbf{0.524}}} \\ \hline
nasdaq &
Yes &
sem\_eval &
\multicolumn{1}{c|}{0.265} &
\multicolumn{1}{c|}{0.332} &
\multicolumn{1}{c|}{{\textbf{0.477}}} &
\multicolumn{1}{c|}{0.567} &
\multicolumn{1}{c|}{0.417} &
\multicolumn{1}{c|}{{\textbf{0.601}}} \\ \hline
nasdaq &
No &
financial\_phrase\_bank &
\multicolumn{1}{c|}{0.192} &
\multicolumn{1}{c|}{0.383} &
\multicolumn{1}{c|}{{\textbf{0.471}}} &
\multicolumn{1}{c|}{0.489} &
\multicolumn{1}{c|}{0.359} &
\multicolumn{1}{c|}{{\textbf{0.531}}} \\ \hline
nasdaq &
No &
fiqa\_labeled\_df &
\multicolumn{1}{c|}{0.216} &
\multicolumn{1}{c|}{0.313} &
\multicolumn{1}{c|}{{\textbf{0.374}}} &
\multicolumn{1}{c|}{0.594} &
\multicolumn{1}{c|}{0.492} &
\multicolumn{1}{c|}{{\textbf{0.615}}} \\ \hline
nasdaq &
No &
fpb\_fiqa &
\multicolumn{1}{c|}{0.209} &
\multicolumn{1}{c|}{0.281} &
\multicolumn{1}{c|}{{\textbf{0.399}}} &
\multicolumn{1}{c|}{0.519} &
\multicolumn{1}{c|}{0.301} &
\multicolumn{1}{c|}{{\textbf{0.525}}} \\ \hline
nasdaq &
No &
sem\_eval &
\multicolumn{1}{c|}{0.265} &
\multicolumn{1}{c|}{0.349} &
\multicolumn{1}{c|}{{\textbf{0.465}}} &
\multicolumn{1}{c|}{0.567} &
\multicolumn{1}{c|}{0.437} &
\multicolumn{1}{c|}{{\textbf{0.578}}} \\ \hline
fiqa\_fpb\_sentfin\_neutral &
Yes &
financial\_phrase\_bank &
\multicolumn{1}{c|}{0.192} &
\multicolumn{1}{c|}{0.392} &
\multicolumn{1}{c|}{{\textbf{0.434}}} &
\multicolumn{1}{c|}{0.489} &
\multicolumn{1}{c|}{0.46} &
\multicolumn{1}{c|}{{\textbf{0.539}}} \\ \hline
fiqa\_fpb\_sentfin\_neutral &
Yes &
fiqa\_labeled\_df &
\multicolumn{1}{c|}{0.216} &
\multicolumn{1}{c|}{0.303} &
\multicolumn{1}{c|}{{\textbf{0.371}}} &
\multicolumn{1}{c|}{0.594} &
\multicolumn{1}{c|}{0.523} &
\multicolumn{1}{c|}{{\textbf{0.659}}} \\ \hline
fiqa\_fpb\_sentfin\_neutral &
Yes &
fpb\_fiqa &
\multicolumn{1}{c|}{0.209} &
\multicolumn{1}{c|}{0.286} &
\multicolumn{1}{c|}{{\textbf{0.376}}} &
\multicolumn{1}{c|}{0.519} &
\multicolumn{1}{c|}{0.375} &
\multicolumn{1}{c|}{{\textbf{0.54}}} \\ \hline
fiqa\_fpb\_sentfin\_neutral &
Yes &
sem\_eval &
\multicolumn{1}{c|}{0.265} &
\multicolumn{1}{c|}{0.308} &
\multicolumn{1}{c|}{{\textbf{0.466}}} &
\multicolumn{1}{c|}{0.567} &
\multicolumn{1}{c|}{0.355} &
\multicolumn{1}{c|}{{\textbf{0.593}}} \\ \hline
fiqa\_fpb\_sentfin\_neutral &
No &
financial\_phrase\_bank &
\multicolumn{1}{c|}{0.192} &
\multicolumn{1}{c|}{0.376} &
\multicolumn{1}{c|}{{\textbf{0.443}}} &
\multicolumn{1}{c|}{0.489} &
\multicolumn{1}{c|}{0.412} &
\multicolumn{1}{c|}{{\textbf{0.539}}} \\ \hline
fiqa\_fpb\_sentfin\_neutral &
No &
fiqa\_labeled\_df &
\multicolumn{1}{c|}{0.216} &
\multicolumn{1}{c|}{0.249} &
\multicolumn{1}{c|}{{\textbf{0.32}}} &
\multicolumn{1}{c|}{0.594} &
\multicolumn{1}{c|}{0.499} &
\multicolumn{1}{c|}{{\textbf{0.659}}} \\ \hline
fiqa\_fpb\_sentfin\_neutral &
No &
fpb\_fiqa &
\multicolumn{1}{c|}{0.209} &
\multicolumn{1}{c|}{0.288} &
\multicolumn{1}{c|}{{\textbf{0.373}}} &
\multicolumn{1}{c|}{0.519} &
\multicolumn{1}{c|}{0.373} &
\multicolumn{1}{c|}{{\textbf{0.531}}} \\ \hline
fiqa\_fpb\_sentfin\_neutral &
No &
sem\_eval &
\multicolumn{1}{c|}{0.265} &
\multicolumn{1}{c|}{0.31} &
\multicolumn{1}{c|}{{\textbf{0.481}}} &
\multicolumn{1}{c|}{0.567} &
\multicolumn{1}{c|}{0.322} &
\multicolumn{1}{c|}{{\textbf{0.593}}} \\ \hline
financial\_phrase\_bank &
Yes &
fiqa\_labeled\_df &
\multicolumn{1}{c|}{0.216} &
\multicolumn{1}{c|}{0.25} &
\multicolumn{1}{c|}{{\textbf{0.363}}} &
\multicolumn{1}{c|}{0.594} &
\multicolumn{1}{c|}{0.375} &
\multicolumn{1}{c|}{{\textbf{0.683}}} \\ \hline
financial\_phrase\_bank &
Yes &
fpb\_fiqa &
\multicolumn{1}{c|}{0.209} &
\multicolumn{1}{c|}{0.245} &
\multicolumn{1}{c|}{{\textbf{0.351}}} &
\multicolumn{1}{c|}{0.519} &
\multicolumn{1}{c|}{0.303} &
\multicolumn{1}{c|}{{\textbf{0.528}}} \\ \hline
financial\_phrase\_bank &
Yes &
sem\_eval &
\multicolumn{1}{c|}{0.265} &
\multicolumn{1}{c|}{0.296} &
\multicolumn{1}{c|}{{\textbf{0.422}}} &
\multicolumn{1}{c|}{0.567} &
\multicolumn{1}{c|}{0.386} &
\multicolumn{1}{c|}{{\textbf{0.581}}} \\ \hline
financial\_phrase\_bank &
No &
fiqa\_labeled\_df &
\multicolumn{1}{c|}{0.216} &
\multicolumn{1}{c|}{0.306} &
\multicolumn{1}{c|}{{\textbf{0.398}}} &
\multicolumn{1}{c|}{0.594} &
\multicolumn{1}{c|}{0.385} &
\multicolumn{1}{c|}{{\textbf{0.615}}} \\ \hline
financial\_phrase\_bank &
No &
fpb\_fiqa &
\multicolumn{1}{c|}{0.209} &
\multicolumn{1}{c|}{0.284} &
\multicolumn{1}{c|}{{\textbf{0.383}}} &
\multicolumn{1}{c|}{0.519} &
\multicolumn{1}{c|}{0.325} &
\multicolumn{1}{c|}{{\textbf{0.529}}} \\ \hline
financial\_phrase\_bank &
No &
sem\_eval &
\multicolumn{1}{c|}{0.265} &
\multicolumn{1}{c|}{0.296} &
\multicolumn{1}{c|}{{\textbf{0.418}}} &
\multicolumn{1}{c|}{0.567} &
\multicolumn{1}{c|}{0.4} &
\multicolumn{1}{c|}{{\textbf{0.581}}} \\ \hline

\end{tabular}
\label{MCC_Results}
\end{table*}

\begin{table*}[]
\caption{F1 score obtained across different experiments using the XLex methodology based on the FinBERT transformer model. The highest values are represented in bold.}
\centering
\small
\setlength\tabcolsep{2pt}

\begin{tabular}{|c|c|c|ccc|ccc|}
\hline
\multicolumn{1}{|c|}{Source of the lexicon} &
  \multicolumn{1}{|c|}{Normalized} &
  \multicolumn{1}{|c|}{Evaluation set} &
  \multicolumn{3}{c|}{F1 on whole dataset} & \multicolumn{3}{c|}{\begin{tabular}[c]{@{}c@{}}F1 only on the part of the dataset for\\which the LM-based model has an answer
  \end{tabular}} \\ 
  \cline{4-9} 
 &
   &
   &
  \multicolumn{1}{c|}{LM} &
  \multicolumn{1}{c|}{XLex} &
  \multicolumn{1}{c|}{\begin{tabular}[c]{@{}c@{}}XLex+\\ LM\end{tabular}} &
  \multicolumn{1}{c|}{\begin{tabular}[c]{@{}c@{}}LM on\\ LM\end{tabular}} &
  \multicolumn{1}{c|}{\begin{tabular}[c]{@{}c@{}}XLex on\\ LM\end{tabular}} &
  \begin{tabular}[c]{@{}c@{}}(XLex+LM)\\ on LM\end{tabular} \\ \hline
nasdaq &
Yes &
financial\_phrase\_bank &
\multicolumn{1}{c|}{0.287} &
\multicolumn{1}{c|}{0.421} &
\multicolumn{1}{c|}{{\textbf{0.432}}} &
\multicolumn{1}{c|}{0.688} &
\multicolumn{1}{c|}{0.441} &
\multicolumn{1}{c|}{{\textbf{0.694}}} \\ \hline
nasdaq &
Yes &
fiqa\_labeled\_df &
\multicolumn{1}{c|}{0.297} &
\multicolumn{1}{c|}{0.497} &
\multicolumn{1}{c|}{{\textbf{0.505}}} &
\multicolumn{1}{c|}{0.789} &
\multicolumn{1}{c|}{0.802} &
\multicolumn{1}{c|}{{\textbf{0.836}}} \\ \hline
nasdaq &
Yes &
fpb\_fiqa &
\multicolumn{1}{c|}{0.295} &
\multicolumn{1}{c|}{0.448} &
\multicolumn{1}{c|}{{\textbf{0.458}}} &
\multicolumn{1}{c|}{0.726} &
\multicolumn{1}{c|}{0.477} &
\multicolumn{1}{c|}{{\textbf{0.74}}} \\ \hline
nasdaq &
Yes &
sem\_eval &
\multicolumn{1}{c|}{0.305} &
\multicolumn{1}{c|}{0.684} &
\multicolumn{1}{c|}{{\textbf{0.693}}} &
\multicolumn{1}{c|}{0.752} &
\multicolumn{1}{c|}{0.774} &
\multicolumn{1}{c|}{{\textbf{0.783}}} \\ \hline
nasdaq &
No &
financial\_phrase\_bank &
\multicolumn{1}{c|}{0.287} &
\multicolumn{1}{c|}{0.389} &
\multicolumn{1}{c|}{{\textbf{0.408}}} &
\multicolumn{1}{c|}{0.688} &
\multicolumn{1}{c|}{0.419} &
\multicolumn{1}{c|}{{\textbf{0.696}}} \\ \hline
nasdaq &
No &
fiqa\_labeled\_df &
\multicolumn{1}{c|}{0.297} &
\multicolumn{1}{c|}{0.498} &
\multicolumn{1}{c|}{{\textbf{0.5}}} &
\multicolumn{1}{c|}{0.789} &
\multicolumn{1}{c|}{0.815} &
\multicolumn{1}{c|}{{\textbf{0.823}}} \\ \hline
nasdaq &
No &
fpb\_fiqa &
\multicolumn{1}{c|}{0.295} &
\multicolumn{1}{c|}{0.428} &
\multicolumn{1}{c|}{{\textbf{0.442}}} &
\multicolumn{1}{c|}{0.726} &
\multicolumn{1}{c|}{0.464} &
\multicolumn{1}{c|}{{\textbf{0.74}}} \\ \hline
nasdaq &
No &
sem\_eval &
\multicolumn{1}{c|}{0.305} &
\multicolumn{1}{c|}{0.676} &
\multicolumn{1}{c|}{{\textbf{0.692}}} &
\multicolumn{1}{c|}{0.752} &
\multicolumn{1}{c|}{0.766} &
\multicolumn{1}{c|}{{\textbf{0.791}}} \\ \hline
fiqa\_fpb\_sentfin\_neutral &
Yes &
financial\_phrase\_bank &
\multicolumn{1}{c|}{0.287} &
\multicolumn{1}{c|}{0.708} &
\multicolumn{1}{c|}{{\textbf{0.724}}} &
\multicolumn{1}{c|}{0.688} &
\multicolumn{1}{c|}{0.673} &
\multicolumn{1}{c|}{{\textbf{0.702}}} \\ \hline
fiqa\_fpb\_sentfin\_neutral &
Yes &
fiqa\_labeled\_df &
\multicolumn{1}{c|}{0.297} &
\multicolumn{1}{c|}{0.522} &
\multicolumn{1}{c|}{{\textbf{0.537}}} &
\multicolumn{1}{c|}{0.789} &
\multicolumn{1}{c|}{0.765} &
\multicolumn{1}{c|}{{\textbf{0.823}}} \\ \hline
fiqa\_fpb\_sentfin\_neutral &
Yes &
fpb\_fiqa &
\multicolumn{1}{c|}{0.295} &
\multicolumn{1}{c|}{0.462} &
\multicolumn{1}{c|}{{\textbf{0.484}}} &
\multicolumn{1}{c|}{0.726} &
\multicolumn{1}{c|}{0.458} &
\multicolumn{1}{c|}{{\textbf{0.75}}} \\ \hline
fiqa\_fpb\_sentfin\_neutral &
Yes &
sem\_eval &
\multicolumn{1}{c|}{0.305} &
\multicolumn{1}{c|}{0.69} &
\multicolumn{1}{c|}{{\textbf{0.741}}} &
\multicolumn{1}{c|}{0.752} &
\multicolumn{1}{c|}{0.715} &
\multicolumn{1}{c|}{{\textbf{0.798}}} \\ \hline
fiqa\_fpb\_sentfin\_neutral &
No &
financial\_phrase\_bank &
\multicolumn{1}{c|}{0.287} &
\multicolumn{1}{c|}{0.696} &
\multicolumn{1}{c|}{{\textbf{0.711}}} &
\multicolumn{1}{c|}{0.688} &
\multicolumn{1}{c|}{0.673} &
\multicolumn{1}{c|}{{\textbf{0.702}}} \\ \hline
fiqa\_fpb\_sentfin\_neutral &
No &
fiqa\_labeled\_df &
\multicolumn{1}{c|}{0.297} &
\multicolumn{1}{c|}{0.518} &
\multicolumn{1}{c|}{{\textbf{0.53}}} &
\multicolumn{1}{c|}{0.789} &
\multicolumn{1}{c|}{0.765} &
\multicolumn{1}{c|}{{\textbf{0.811}}} \\ \hline
fiqa\_fpb\_sentfin\_neutral &
No &
fpb\_fiqa &
\multicolumn{1}{c|}{0.295} &
\multicolumn{1}{c|}{0.461} &
\multicolumn{1}{c|}{{\textbf{0.483}}} &
\multicolumn{1}{c|}{0.726} &
\multicolumn{1}{c|}{0.459} &
\multicolumn{1}{c|}{{\textbf{0.751}}} \\ \hline
fiqa\_fpb\_sentfin\_neutral &
No &
sem\_eval &
\multicolumn{1}{c|}{0.305} &
\multicolumn{1}{c|}{0.696} &
\multicolumn{1}{c|}{{\textbf{0.742}}} &
\multicolumn{1}{c|}{0.752} &
\multicolumn{1}{c|}{0.724} &
\multicolumn{1}{c|}{{\textbf{0.806}}} \\ \hline
financial\_phrase\_bank &
Yes &
fiqa\_labeled\_df &
\multicolumn{1}{c|}{0.297} &
\multicolumn{1}{c|}{0.486} &
\multicolumn{1}{c|}{{\textbf{0.511}}} &
\multicolumn{1}{c|}{0.789} &
\multicolumn{1}{c|}{0.471} &
\multicolumn{1}{c|}{{\textbf{0.814}}} \\ \hline
financial\_phrase\_bank &
Yes &
fpb\_fiqa &
\multicolumn{1}{c|}{0.295} &
\multicolumn{1}{c|}{0.461} &
\multicolumn{1}{c|}{{\textbf{0.486}}} &
\multicolumn{1}{c|}{0.726} &
\multicolumn{1}{c|}{0.453} &
\multicolumn{1}{c|}{{\textbf{0.739}}} \\ \hline
financial\_phrase\_bank &
Yes &
sem\_eval &
\multicolumn{1}{c|}{0.305} &
\multicolumn{1}{c|}{0.432} &
\multicolumn{1}{c|}{{\textbf{0.488}}} &
\multicolumn{1}{c|}{0.752} &
\multicolumn{1}{c|}{0.452} &
\multicolumn{1}{c|}{{\textbf{0.814}}} \\ \hline
financial\_phrase\_bank &
No &
fiqa\_labeled\_df &
\multicolumn{1}{c|}{0.297} &
\multicolumn{1}{c|}{0.476} &
\multicolumn{1}{c|}{{\textbf{0.496}}} &
\multicolumn{1}{c|}{0.789} &
\multicolumn{1}{c|}{0.479} &
\multicolumn{1}{c|}{{\textbf{0.802}}} \\ \hline
financial\_phrase\_bank &
No &
fpb\_fiqa &
\multicolumn{1}{c|}{0.295} &
\multicolumn{1}{c|}{0.448} &
\multicolumn{1}{c|}{{\textbf{0.477}}} &
\multicolumn{1}{c|}{0.726} &
\multicolumn{1}{c|}{0.446} &
\multicolumn{1}{c|}{{\textbf{0.74}}} \\ \hline
financial\_phrase\_bank &
No &
sem\_eval &
\multicolumn{1}{c|}{0.305} &
\multicolumn{1}{c|}{0.429} &
\multicolumn{1}{c|}{{\textbf{0.488}}} &
\multicolumn{1}{c|}{0.752} &
\multicolumn{1}{c|}{0.452} &
\multicolumn{1}{c|}{{\textbf{0.806}}} \\ \hline

\end{tabular}
\label{Finbert_F1_Results}
\end{table*}

\begin{table*}[]
\caption{MCC obtained across different experiments using the XLex methodology based on the FinBERT transformer model. The highest values are represented in bold.}
\centering
\small
\setlength\tabcolsep{2pt}

\begin{tabular}{|c|c|c|ccc|ccc|}
\hline
\multicolumn{1}{|c|}{Source of the lexicon} &
  \multicolumn{1}{|c|}{Normalized} &
  \multicolumn{1}{|c|}{Evaluation set} &
  \multicolumn{3}{c|}{MCC on whole dataset} & \multicolumn{3}{c|}{\begin{tabular}[c]{@{}c@{}}MCC only on the part of the dataset for\\which the LM-based model has an answer
  \end{tabular}} \\ 
  \cline{4-9} 
 &
   &
   &
  \multicolumn{1}{c|}{LM} &
  \multicolumn{1}{c|}{XLex} &
  \multicolumn{1}{c|}{\begin{tabular}[c]{@{}c@{}}XLex+\\ LM\end{tabular}} &
  \multicolumn{1}{c|}{\begin{tabular}[c]{@{}c@{}}LM on\\ LM\end{tabular}} &
  \multicolumn{1}{c|}{\begin{tabular}[c]{@{}c@{}}XLex on\\ LM\end{tabular}} &
  \begin{tabular}[c]{@{}c@{}}(XLex+LM)\\ on LM\end{tabular} \\ \hline
nasdaq &
Yes &
financial\_phrase\_bank &
\multicolumn{1}{c|}{0.192} &
\multicolumn{1}{c|}{0.315} &
\multicolumn{1}{c|}{{\textbf{0.352}}} &
\multicolumn{1}{c|}{\textbf{0.489}} &
\multicolumn{1}{c|}{0.414} &
\multicolumn{1}{c|}{{\textbf{0.489}}} \\ \hline
nasdaq &
Yes &
fiqa\_labeled\_df &
\multicolumn{1}{c|}{0.216} &
\multicolumn{1}{c|}{0.497} &
\multicolumn{1}{c|}{{\textbf{0.525}}} &
\multicolumn{1}{c|}{0.594} &
\multicolumn{1}{c|}{0.604} &
\multicolumn{1}{c|}{{\textbf{0.673}}} \\ \hline
nasdaq &
Yes &
fpb\_fiqa &
\multicolumn{1}{c|}{0.209} &
\multicolumn{1}{c|}{0.375} &
\multicolumn{1}{c|}{{\textbf{0.415}}} &
\multicolumn{1}{c|}{0.519} &
\multicolumn{1}{c|}{0.451} &
\multicolumn{1}{c|}{{\textbf{0.526}}} \\ \hline
nasdaq &
Yes &
sem\_eval &
\multicolumn{1}{c|}{0.265} &
\multicolumn{1}{c|}{0.417} &
\multicolumn{1}{c|}{{\textbf{0.452}}} &
\multicolumn{1}{c|}{0.567} &
\multicolumn{1}{c|}{0.565} &
\multicolumn{1}{c|}{{\textbf{0.613}}} \\ \hline
nasdaq &
No &
financial\_phrase\_bank &
\multicolumn{1}{c|}{0.192} &
\multicolumn{1}{c|}{0.279} &
\multicolumn{1}{c|}{{\textbf{0.326}}} &
\multicolumn{1}{c|}{0.489} &
\multicolumn{1}{c|}{0.378} &
\multicolumn{1}{c|}{{\textbf{0.493}}} \\ \hline
nasdaq &
No &
fiqa\_labeled\_df &
\multicolumn{1}{c|}{0.216} &
\multicolumn{1}{c|}{0.503} &
\multicolumn{1}{c|}{{\textbf{0.516}}} &
\multicolumn{1}{c|}{0.594} &
\multicolumn{1}{c|}{0.632} &
\multicolumn{1}{c|}{{\textbf{0.648}}} \\ \hline
nasdaq &
No &
fpb\_fiqa &
\multicolumn{1}{c|}{0.209} &
\multicolumn{1}{c|}{0.334} &
\multicolumn{1}{c|}{{\textbf{0.381}}} &
\multicolumn{1}{c|}{0.519} &
\multicolumn{1}{c|}{0.425} &
\multicolumn{1}{c|}{{\textbf{0.526}}} \\ \hline
nasdaq &
No &
sem\_eval &
\multicolumn{1}{c|}{0.265} &
\multicolumn{1}{c|}{0.415} &
\multicolumn{1}{c|}{{\textbf{0.465}}} &
\multicolumn{1}{c|}{0.567} &
\multicolumn{1}{c|}{0.546} &
\multicolumn{1}{c|}{{\textbf{0.624}}} \\ \hline
fiqa\_fpb\_sentfin\_neutral &
Yes &
financial\_phrase\_bank &
\multicolumn{1}{c|}{0.192} &
\multicolumn{1}{c|}{0.454} &
\multicolumn{1}{c|}{{\textbf{0.494}}} &
\multicolumn{1}{c|}{0.489} &
\multicolumn{1}{c|}{0.418} &
\multicolumn{1}{c|}{{\textbf{0.5}}} \\ \hline
fiqa\_fpb\_sentfin\_neutral &
Yes &
fiqa\_labeled\_df &
\multicolumn{1}{c|}{0.216} &
\multicolumn{1}{c|}{0.565} &
\multicolumn{1}{c|}{{\textbf{0.608}}} &
\multicolumn{1}{c|}{0.594} &
\multicolumn{1}{c|}{0.53} &
\multicolumn{1}{c|}{{\textbf{0.648}}} \\ \hline
fiqa\_fpb\_sentfin\_neutral &
Yes &
fpb\_fiqa &
\multicolumn{1}{c|}{0.209} &
\multicolumn{1}{c|}{0.394} &
\multicolumn{1}{c|}{{\textbf{0.47}}} &
\multicolumn{1}{c|}{0.519} &
\multicolumn{1}{c|}{0.386} &
\multicolumn{1}{c|}{{\textbf{0.54}}} \\ \hline
fiqa\_fpb\_sentfin\_neutral &
Yes &
sem\_eval &
\multicolumn{1}{c|}{0.265} &
\multicolumn{1}{c|}{0.39} &
\multicolumn{1}{c|}{{\textbf{0.511}}} &
\multicolumn{1}{c|}{0.567} &
\multicolumn{1}{c|}{0.43} &
\multicolumn{1}{c|}{{\textbf{0.636}}} \\ \hline
fiqa\_fpb\_sentfin\_neutral &
No &
financial\_phrase\_bank &
\multicolumn{1}{c|}{0.192} &
\multicolumn{1}{c|}{0.441} &
\multicolumn{1}{c|}{{\textbf{0.478}}} &
\multicolumn{1}{c|}{0.489} &
\multicolumn{1}{c|}{0.424} &
\multicolumn{1}{c|}{{\textbf{0.5}}} \\ \hline
fiqa\_fpb\_sentfin\_neutral &
No &
fiqa\_labeled\_df &
\multicolumn{1}{c|}{0.216} &
\multicolumn{1}{c|}{0.557} &
\multicolumn{1}{c|}{{\textbf{0.592}}} &
\multicolumn{1}{c|}{0.594} &
\multicolumn{1}{c|}{0.53} &
\multicolumn{1}{c|}{{\textbf{0.624}}} \\ \hline
fiqa\_fpb\_sentfin\_neutral &
No &
fpb\_fiqa &
\multicolumn{1}{c|}{0.209} &
\multicolumn{1}{c|}{0.398} &
\multicolumn{1}{c|}{{\textbf{0.472}}} &
\multicolumn{1}{c|}{0.519} &
\multicolumn{1}{c|}{0.392} &
\multicolumn{1}{c|}{{\textbf{0.543}}} \\ \hline
fiqa\_fpb\_sentfin\_neutral &
No &
sem\_eval &
\multicolumn{1}{c|}{0.265} &
\multicolumn{1}{c|}{0.407} &
\multicolumn{1}{c|}{{\textbf{0.517}}} &
\multicolumn{1}{c|}{0.567} &
\multicolumn{1}{c|}{0.451} &
\multicolumn{1}{c|}{{\textbf{0.648}}} \\ \hline
financial\_phrase\_bank &
Yes &
fiqa\_labeled\_df &
\multicolumn{1}{c|}{0.216} &
\multicolumn{1}{c|}{0.456} &
\multicolumn{1}{c|}{{\textbf{0.531}}} &
\multicolumn{1}{c|}{0.594} &
\multicolumn{1}{c|}{0.465} &
\multicolumn{1}{c|}{{\textbf{0.637}}} \\ \hline
financial\_phrase\_bank &
Yes &
fpb\_fiqa &
\multicolumn{1}{c|}{0.209} &
\multicolumn{1}{c|}{0.382} &
\multicolumn{1}{c|}{{\textbf{0.468}}} &
\multicolumn{1}{c|}{0.519} &
\multicolumn{1}{c|}{0.361} &
\multicolumn{1}{c|}{{\textbf{0.524}}} \\ \hline
financial\_phrase\_bank &
Yes &
sem\_eval &
\multicolumn{1}{c|}{0.265} &
\multicolumn{1}{c|}{0.297} &
\multicolumn{1}{c|}{{\textbf{0.483}}} &
\multicolumn{1}{c|}{0.567} &
\multicolumn{1}{c|}{0.371} &
\multicolumn{1}{c|}{{\textbf{0.653}}} \\ \hline
financial\_phrase\_bank &
No &
fiqa\_labeled\_df &
\multicolumn{1}{c|}{0.216} &
\multicolumn{1}{c|}{0.429} &
\multicolumn{1}{c|}{{\textbf{0.489}}} &
\multicolumn{1}{c|}{0.594} &
\multicolumn{1}{c|}{0.5} &
\multicolumn{1}{c|}{{\textbf{0.615}}} \\ \hline
financial\_phrase\_bank &
No &
fpb\_fiqa &
\multicolumn{1}{c|}{0.209} &
\multicolumn{1}{c|}{0.346} &
\multicolumn{1}{c|}{{\textbf{0.443}}} &
\multicolumn{1}{c|}{0.519} &
\multicolumn{1}{c|}{0.341} &
\multicolumn{1}{c|}{{\textbf{0.523}}} \\ \hline
financial\_phrase\_bank &
No &
sem\_eval &
\multicolumn{1}{c|}{0.265} &
\multicolumn{1}{c|}{0.288} &
\multicolumn{1}{c|}{{\textbf{0.483}}} &
\multicolumn{1}{c|}{0.567} &
\multicolumn{1}{c|}{0.388} &
\multicolumn{1}{c|}{{\textbf{0.648}}} \\ \hline

\end{tabular}
\label{Finbert_MCC_Results}
\end{table*}

\clearpage
\twocolumn

\bibliographystyle{IEEEtran}
\bibliography{biblio.bib}

\begin{IEEEbiography}
[{\includegraphics[width=1in,height=1.25in,clip,keepaspectratio]{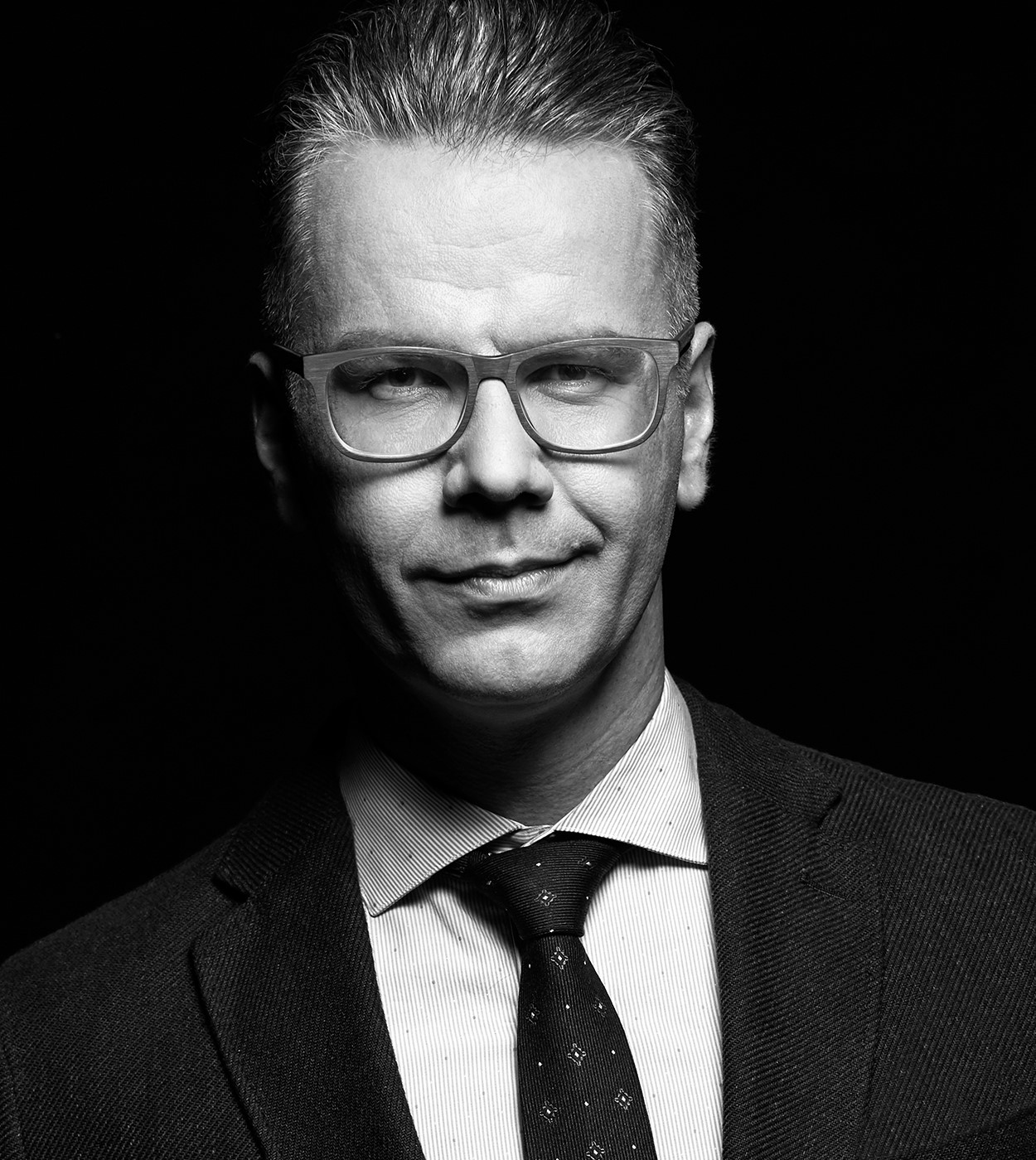}}]{Maryan Rizinski} received the B.S. and M.S. degrees in electrical engineering and information technologies from University Ss. Cyril and Methodius in Skopje, where he is a Ph.D. candidate in computer science. He is currently an engineering manager at Bosch, with over ten years of industry experience leading globally-distributed software engineering teams. His expertise spans multiple aspects of the software project lifecycle management, from planning, requirement gathering, and analysis, estimations to driving delivery, rollout, and troubleshooting for international customers. Throughout his professional career, he has managed the implementation of Internet of Things (IoT) and fiber-optics infrastructure projects and has been mentoring and consulting startup IT companies. He is also a lecturer of computer science at Boston University’s Metropolitan College, where he is teaching and facilitating networking and data science classes. His doctoral research focuses on novel approaches for using machine learning (ML) and natural language processing (NLP) in the financial industry and other related areas. His research aims to enable more accurate decision-making and address fundamental problems of improving the explainability of deep-learning models and addressing ML-related ethical challenges in finance applications. His past research interests focused on computer networking, wireless communications, and new Internet and IoT architectures.
\end{IEEEbiography}

\begin{IEEEbiography}
[{\includegraphics[width=1in,height=1.25in,clip,keepaspectratio]{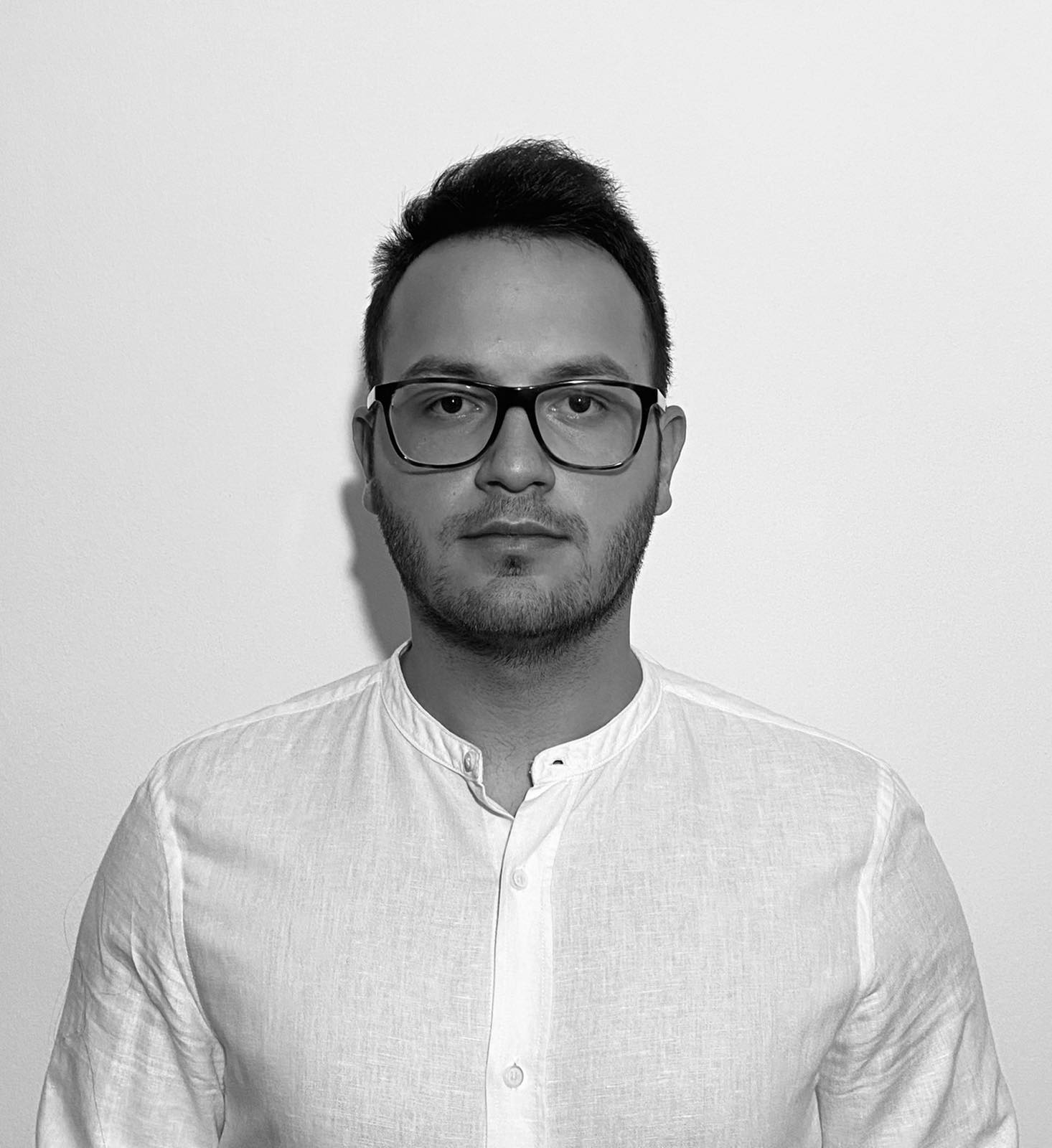}}]{Hristijan Peshov} received the Bachelor of Science degree in software engineering and information systems from the Faculty of Computer Science and Engineering, Saints Cyril and Methodius University in Skopje, in 2022. He also works as a Software Engineer. His research interests include data science, machine learning, explainable AI, natural language processing, and network analysis.
\end{IEEEbiography}

\begin{IEEEbiography}[{\includegraphics[width=1in,height=1.25in,clip,keepaspectratio]{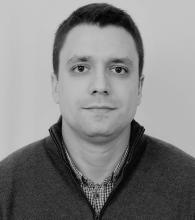}}]{Kostadin Mishev} received the bachelor’s degree in informatics and computer engineering and the master’s degree in computer networks and e-technologies degree from Saints Cyril and Methodius University, Skopje, in 2013 and 2016, respectively, where he is currently pursuing the Ph.D. degree. He is also a Teaching and a Research Assistant with the Faculty of Computer Science and Engineering, Saints Cyril and Methodius University. His research interests include Data Science, Natural Language Processing, Semantic Web, Web technologies, and Computer Networks.
\end{IEEEbiography}

\begin{IEEEbiography}[{\includegraphics[width=1in,height=1.25in,clip,keepaspectratio]{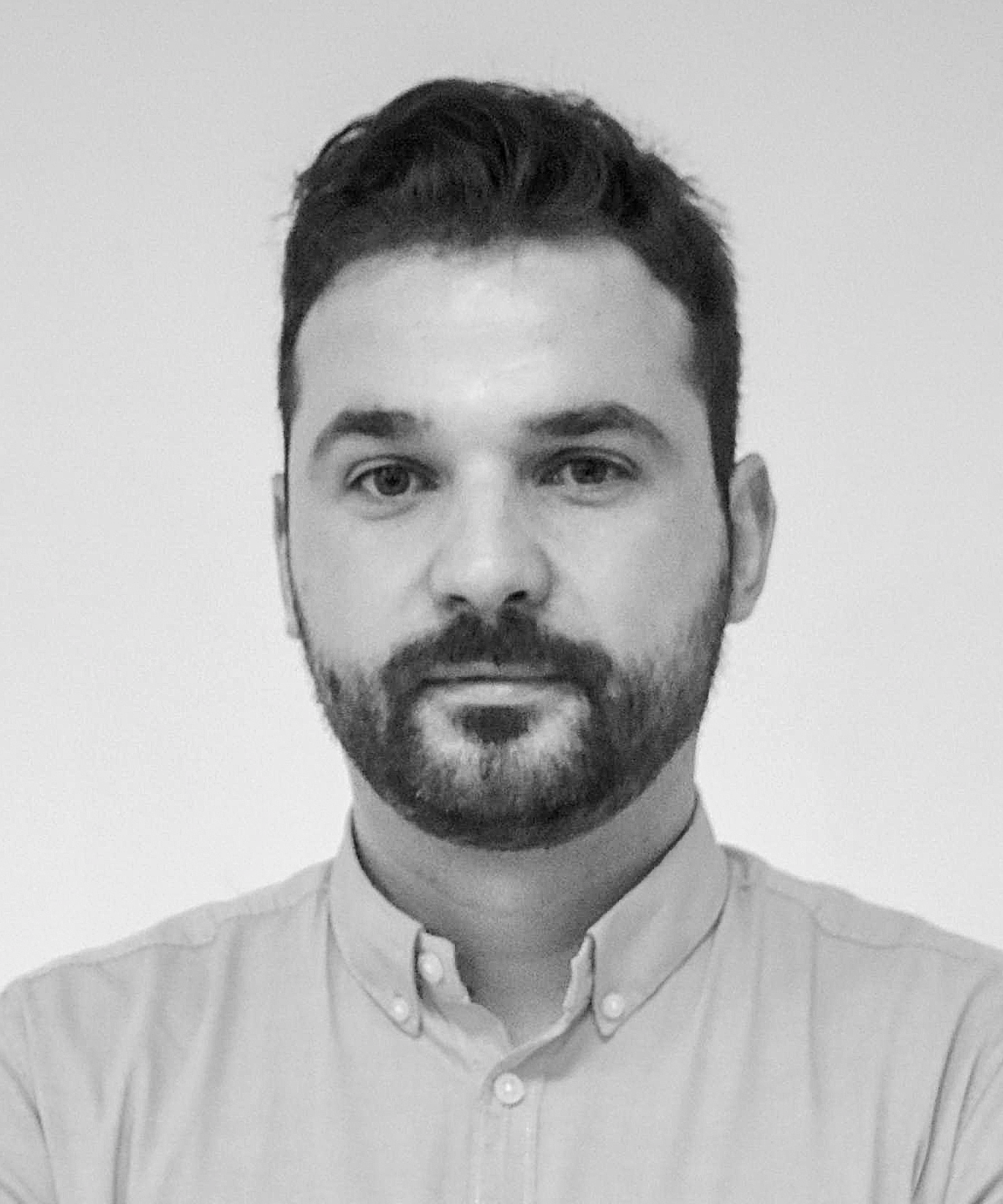}}]{Milos Jovanovik} is an Associate Professor at the Faculty of Computer Science and Engineering, at the Ss. Cyril and Methodius University in Skopje. He's also a Senior R\&D Knowledge Graphs Engineer at OpenLink Software, London, UK. He obtained his Ph.D. in 2016 at the Ss. Cyril and Methodius University in Skopje, in the field of Computer Science and Engineering, with a doctoral thesis in the domain of Linked Data. He has published over 50 scientific papers and has participated in over 30 research projects on international and national levels. His main research interests include Knowledge Graphs, Linked Data, Open Data, and Data Science.
\end{IEEEbiography}

\begin{IEEEbiography}[{\includegraphics[width=1in,height=1.25in,clip,keepaspectratio]{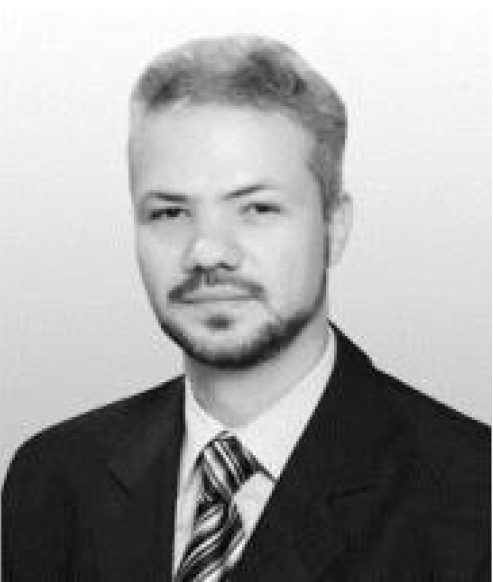}}]
{Dimitar Trajanov} (Member, IEEE) received a Ph.D. degree in computer science. He is a Full professor at the Faculty of Computer Science and Engineering - ss. Cyril and Methodius University – Skopje and Visiting Research Professor at Boston University. From March 2011 until September 2015, he was the founding Dean of the Faculty of Computer Science and Engineering, and in his tenure, the faculty became the largest technical Faculty in Macedonia. Dimitar Trajanov is the leader of the Regional Social Innovation Hub, established in 2013 as a cooperation between UNDP and the Faculty of Computer Science and Engineering. Dimitar Trajanov is the author of more than 200 journal and conference papers and seven books. He has been involved in more than 70 research and industry projects, of which in more than 40 projects as a project leader. His research interests include Data Science, Machine Learning, NLP, FinTech, Semantic Web, e-commerce,  Technology for Development, ESG, and Climate Change.

\end{IEEEbiography}

\EOD

\end{document}